\documentclass[10pt,twocolumn,letterpaper]{article}

\usepackage{iccv}
\usepackage{times}
\usepackage{epsfig}
\usepackage{graphicx}
\usepackage{amsmath}
\usepackage{amssymb}
\usepackage{subfigure}

\graphicspath{{./graphics/}}
\usepackage{tabularx}
\usepackage{booktabs}
\usepackage{caption}
\usepackage{tikz}
\usepackage{pgfplots}
\usepackage{pgfplotstable}
\usepackage{standalone}
\usepackage{bm}
\usepackage[normalem]{ulem}
\newlength{\itemwidth}

\newcolumntype{Y}{>{\centering\arraybackslash}X}

\usetikzlibrary{calc}
\usetikzlibrary{tikzmark}

\pgfplotsset{compat=newest}
\definecolor{999933}{RGB}{153, 153, 51}
\definecolor{BEBADA}{RGB}{190,186,218}
\definecolor{FB8072}{RGB}{251,128,114}
\definecolor{80B1D3}{RGB}{128,177,211}
\definecolor{FDB463}{RGB}{253,180,98}
\definecolor{8DD3C7}{RGB}{141,211,199}
\pgfplotscreateplotcyclelist{custompalette}{
    {999933!50!black,fill=999933},
    {BEBADA!50!black,fill=BEBADA},
    {FB8072!50!black,fill=FB8072},
    {80B1D3!50!black,fill=80B1D3},
    {FDB463!50!black,fill=FDB463},
    {8DD3C7!50!black,fill=8DD3C7}
}

\usepackage[pagebackref=true,breaklinks=true,letterpaper=true,colorlinks,bookmarks=false]{hyperref}

\iccvfinalcopy

\begin{document}

\title{Video Frame Interpolation via Adaptive Separable Convolution}

\author{Simon Niklaus\\
Portland State University\\
{\tt\small sniklaus@pdx.edu}
\and
Long Mai\\
Portland State University\\
{\tt\small mtlong@cs.pdx.edu}
\and
Feng Liu\\
Portland State University\\
{\tt\small fliu@cs.pdx.edu}
}

\maketitle

\begin{abstract}

    Standard video frame interpolation methods first estimate optical flow between input frames and then synthesize an intermediate frame guided by motion. Recent approaches merge these two steps into a single convolution process by convolving input frames with spatially adaptive kernels that account for motion and re-sampling simultaneously. These methods require large kernels to handle large motion, which limits the number of pixels whose kernels can be estimated at once due to the large memory demand. To address this problem, this paper formulates frame interpolation as local separable convolution over input frames using pairs of 1D kernels. Compared to regular 2D kernels, the 1D kernels require significantly fewer parameters to be estimated. Our method develops a deep fully convolutional neural network that takes two input frames and estimates pairs of 1D kernels for all pixels simultaneously. Since our method is able to estimate kernels and synthesizes the whole video frame at once, it allows for the incorporation of perceptual loss to train the neural network to produce visually pleasing frames. This deep neural network is trained end-to-end using widely available video data without any human annotation. Both qualitative and quantitative experiments show that our method provides a practical solution to high-quality video frame interpolation.

\end{abstract}

\vspace{-0.15in}
\section{Introduction}
\label{sec:intro}
\begin{figure}\centering
    \setlength{\tabcolsep}{0.0cm}
    \setlength{\itemwidth}{4.1cm}

    \begin{tabularx}{\columnwidth}{c @{\hspace{0.1cm}} c}
            \includegraphics[width=\itemwidth]{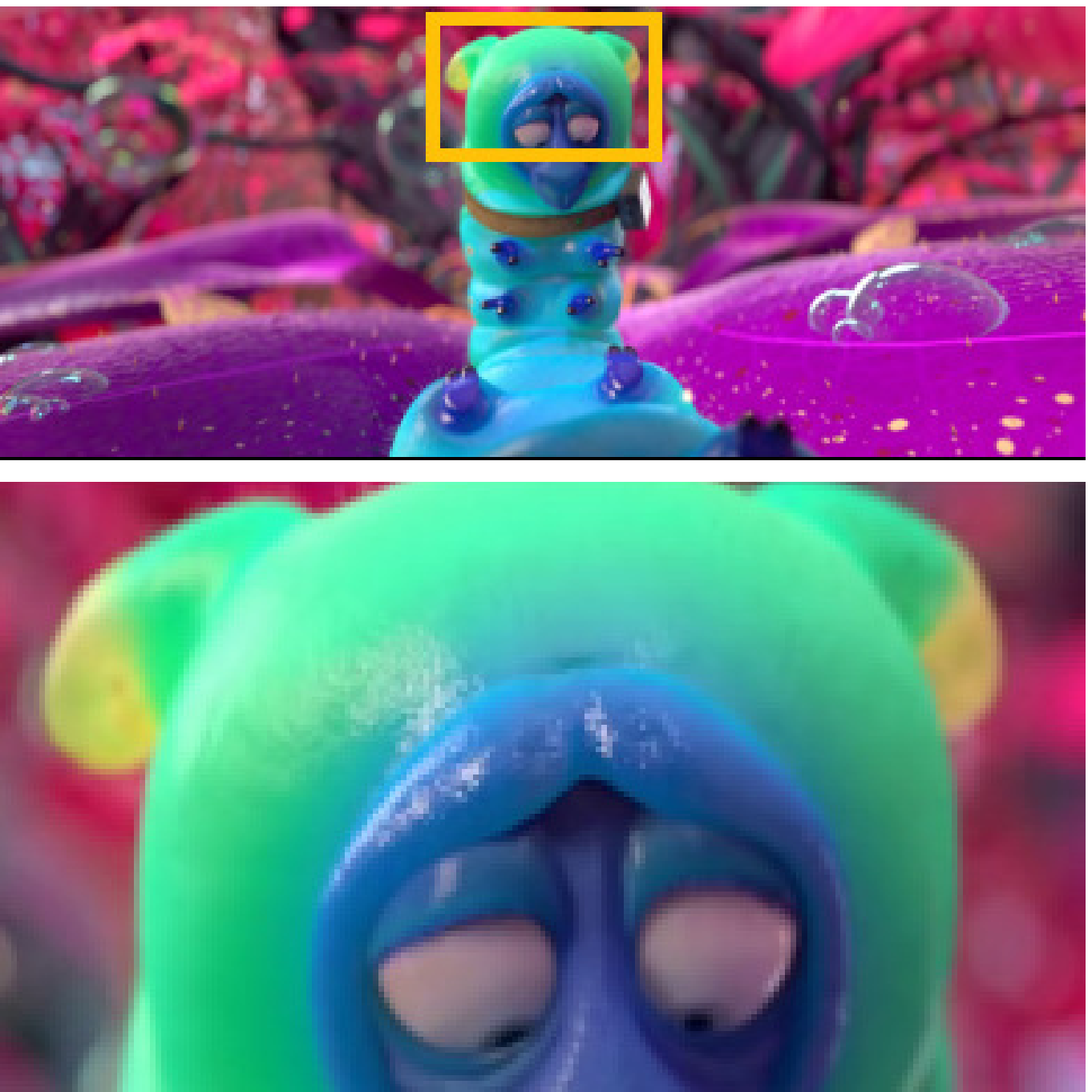}
        &
            \includegraphics[width=\itemwidth]{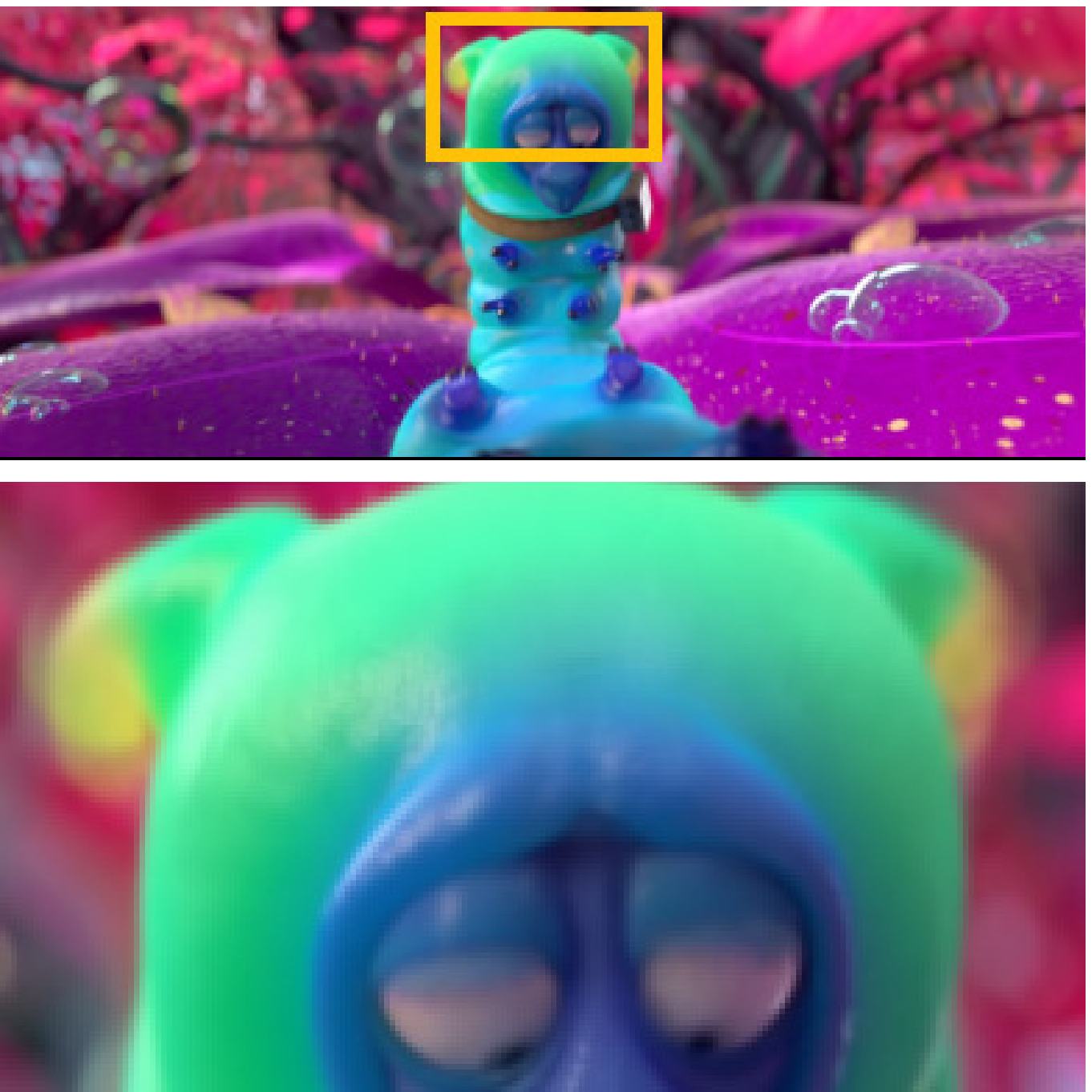}
        \vspace{-0.1cm} \\
            \footnotesize (a) Ground truth
        &
            \footnotesize (b) Niklaus~\etal~\cite{Niklaus_CVPR_2017}
        \\
    \end{tabularx}
    \begin{tabularx}{\columnwidth}{c @{\hspace{0.1cm}} c}
            \includegraphics[width=\itemwidth]{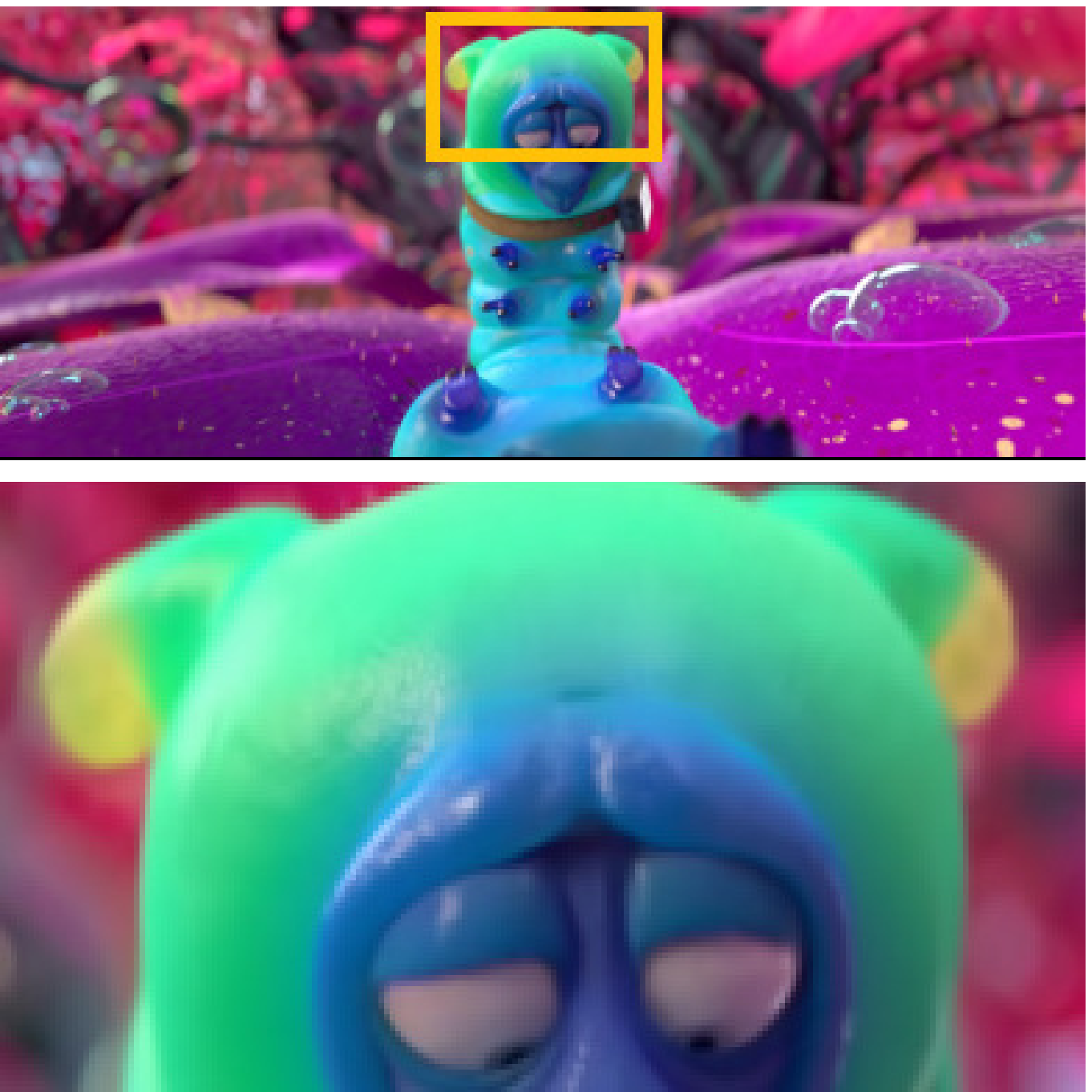}
        &
            \includegraphics[width=\itemwidth]{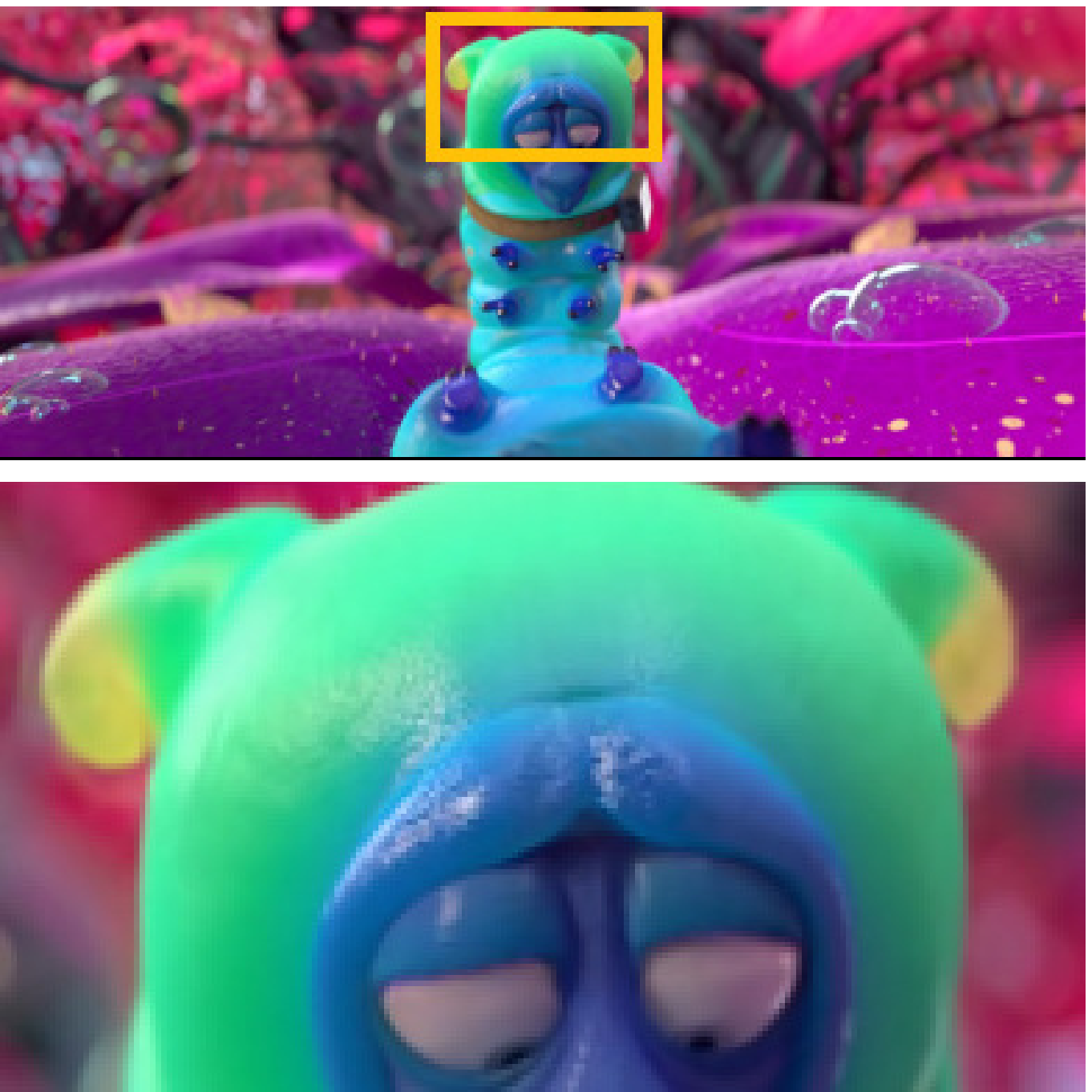}
        \vspace{-0.1cm} \\
            \footnotesize (c) Ours - $\mathcal{L}_1$
        &
            \footnotesize (d) Ours - $\mathcal{L}_F$
        \\
    \end{tabularx}\vspace{-0.1in}
    \caption{Video frame interpolation. Compared to the recent convolution approach that utilizes 2D kernels~\cite{Niklaus_CVPR_2017} (b), our separable convolution methods, especially the one with perceptual loss (d), incorporate 1D kernels that allow for full-frame interpolation and produce higher-quality results.}\vspace{-0.2in}
    \label{fig:teaser}
\end{figure}

Traditional video frame interpolation methods estimate optical flow between input frames and synthesizing intermediate frames guided by optical flow~\cite{Baker_OTHER_2011}. However, their performance largely depends on the quality of optical flow, which is challenging to estimate accurately in regions with occlusion, blur, and abrupt brightness change. {\let\thefootnote\relax\footnote{\url{http://graphics.cs.pdx.edu/project/sepconv}}}

Based on the observation that the ultimate goal of frame interpolation is to produce high-quality video frames and optical flow estimation is only an intermediate step, recent methods formulate frame interpolation~\cite{Niklaus_CVPR_2017} or extrapolation~\cite{Finn_NIPS_2016, Jia_NIPS_2016, Xue_NIPS_2016} as a convolution process. Specifically, they estimate spatially-adaptive convolution kernels for each output pixel and convolve the kernels with the input frames to generate a new frame. The convolution kernels jointly account for the two separate steps of motion estimation and re-sampling involved in traditional frame interpolation methods. In order to handle large motion, large kernels are required. For example, Niklaus~\etal employ a neural network to output two $41 \times 41$ kernels for each output pixel~\cite{Niklaus_CVPR_2017}. To generate the kernels for all pixels in a 1080p video frame, the output kernels alone will require 26 GB of memory. The memory demand increases quadratically with the kernel size and thus limits the maximal motion to be handled. Given this limitation, Niklaus~\etal trained a neural network to output the kernels pixel by pixel.

This paper presents a spatially-adaptive separable convolution approach for video frame interpolation. Our work is inspired by the success of using separable filters to approximate full 2D filters for other computer vision tasks, like image structure extraction~\cite{Rigamonti_CVPR_2013}. For frame synthesis, two 2D convolution kernels are required to generate an output pixel. Our approach approximates each of these with a pair of 1D kernels, one horizontal and one vertical. In this way, an $n \times n$ convolution kernel can be encoded using only $2n$ variables. This allows our method to employ a fully convolutional neural network that takes two video frames as input and produces the separable kernels for all output pixels at once. For a 1080p video frame, using separable kernels that approximate $41 \times 41$ ones only requires 1.27 GB instead of 26 GB of memory. Since our method is able to generate the full-frame output, we can incorporate perceptual loss functions~\cite{Dosovitskiy_NIPS_2016, Johnson_ECCV_2016, Ledig_CORR_2016, Sajjadi_CORR_2016, Zhu_ECCV_2016} to further improve the visual quality of the interpolation results, as shown in Figure~\ref{fig:teaser}.

Our deep neural network is fully convolutional and can be trained end-to-end using widely available video data without any difficult-to-obtain meta data like optical flow. Our experiments show that our method is able to compare favorably to representative state-of-the-art interpolation methods both qualitatively and quantitatively on representative challenging scenarios and provides a practical solution to high-quality video frame interpolation.

\section{Related Work}
\label{sec:related}
Video frame interpolation is a classic topic in computer vision and video processing. Common frame interpolation approaches estimate dense motion, typically optical flow, between two input frames and then interpolate one or more intermediate frames guided by the motion~\cite{Baker_OTHER_2011, Werlberger_OTHER_2011, Yu_OTHER_2013}. The performance of these methods often depends on optical flow and special care, such as flow interpolation, is necessary to handle problems with optical flow~\cite{Baker_OTHER_2011}. Generic image-based rendering algorithms can also be used to improve frame synthesis results~\cite{Mahajan_TOG_2009, Zitnick_TOG_2004}. Different from optical flow based methods, Meyer~\etal developed a phase-based interpolation method that represents motion in the phase shift of individual pixels and generates intermediate frames by per-pixel phase modification~\cite{Meyer_CVPR_2015}. This phase-based method often produces impressive interpolation results; however, it sometimes cannot preserve high-frequency details in videos with large temporal changes.

Our research is inspired by the success of applying deep learning to optical flow estimation~\cite{Bailer_ICCV_2015, Dosovitskiy_ICCV_2015, Gadot_CVPR_2015, Guney_ACCV_2016, Teney_CORR_2016, Tran_CVPR_2015, Weinzaepfel_ICCV_2013}, artistic style transfer~\cite{Gatys_CVPR_2016, Johnson_ECCV_2016, Li_ECCV_2016}, and image enhancement~\cite{Burger_CVPR_2012, Dong_ICCV_2015, Dong_PAMI_2016, Jiansun_CVPR_2015, Svoboda_CORR_2016, Xie_NIPS_2012, Xu_NIPS_2014, Zhang_ECCV_2016, Zhu_ECCV_2016}. Our work belongs to the category of research that employs deep neural networks for view synthesis. Some of these methods render unseen views from input images for objects like faces and chairs, instead of complex real-world scenes~\cite{Dosovitskiy_CVPR_2015, Kulkarni_NIPS_2015, Tatarchenko_ECCV_2016, Yang_NIPS_2015}. Flynn~\etal developed a method that generates a novel image by projecting input images onto multiple depth planes and combining these depth planes to create the novel view~\cite{Flynn_CVPR_2016}. Kalantari~\etal proposed a view expansion method for light field imaging that uses two sequential convolutional neural networks to model the disparity and color estimation steps of view interpolation and trained these two networks simultaneously~\cite{Kalantari_TOG_2016}. Xie~\etal developed a neural network that synthesizes an extra view from a monocular video to convert it to a stereo video~\cite{Xie_ECCV_2016}.

Recently, Zhou~\etal developed an method that employs a convolutional neural network to estimate appearance flow and then uses this estimation to warp input pixels to create a novel view~\cite{Zhou_ECCV_2016}. Their method can warp individual input frames and blend them together to produce a frame between the input ones. The deep voxel flow approach, a concurrent work to our paper, developed a deep neural network to output dense voxel flows optimized frame interpolation results~\cite{Liu_CORR_2017}. Long~\etal also developed a convolutional neural network to interpolate a frame in between two input ones; however, their method generates the interpolated frame as an intermediate step to estimate optical flow~\cite{Long_ECCV_2016}.

Our method is most relevant to the recent frame interpolation~\cite{Niklaus_CVPR_2017} or extrapolation~\cite{Finn_NIPS_2016, Jia_NIPS_2016, Xue_NIPS_2016} methods that combine motion estimation and frame synthesis into a single convolution step. These methods estimate spatially-varying kernels for each output pixel and convolve them with input frames to generate a new frame. Since these convolution methods require large kernels to handle large motion, they cannot synthesize all the pixels for a high-resolution video simultaneously, limited by the available memory. For example, the method from Niklaus~\etal interpolates frames pixel by pixel. Although they employed a shift-and-stitch strategy to generate multiple pixels in each pass, the number of pixels that can be synthesized simultaneously is still limited~\cite{Niklaus_CVPR_2017}. Other methods only generate a relatively low-resolution image. Our work extends these algorithms by estimating separable 1D kernels to approximate 2D kernels which significantly reduces the required amount of memory. Consequently, our method can interpolate a 1080p frame in one pass. Moreover, our method also supports the incorporation of perceptual loss functions, which need to be constrained on a continuous image region, to improve the visual quality of the interpolated frames.

\section{Video Frame Interpolation}
\label{sec:method}
\begin{figure*}\centering
    \hspace*{-0.3cm}\includegraphics[]{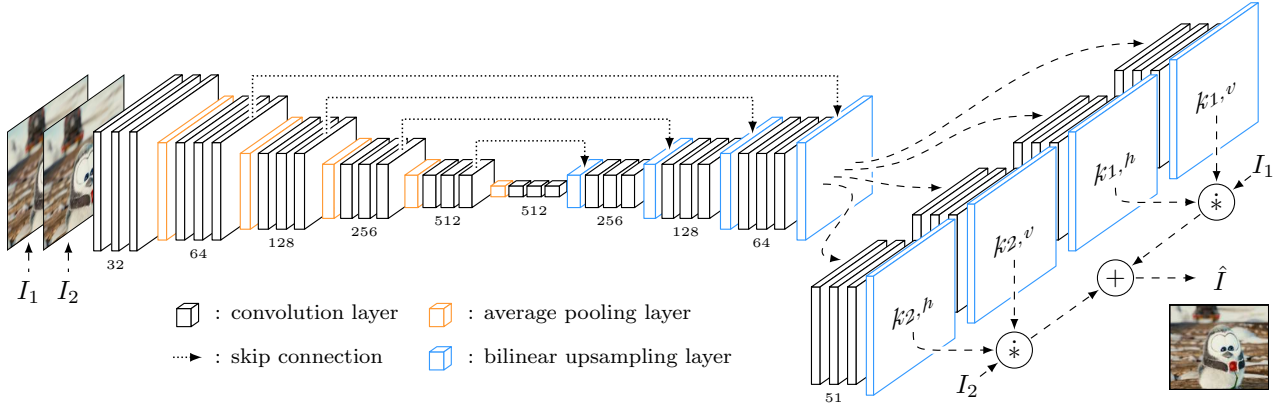}\vspace{-0.2cm}
	\caption{An overview of our neural network architecture. Given input frames $I_1$ and $I_2$, an encoder-decoder network extracts features that are given to four sub-networks that each estimate one of the four 1D kernels for each output pixel in a dense pixel-wise manner. The estimated pixel-dependent kernels are then convolved with the input frames to produce the interpolated frame $\hat{I}$. Note that $\dot{\ast}$ denotes a local convolution.}\vspace{-0.2in}
	\label{fig:architecture}
\end{figure*}

To make this paper self-complete, we first briefly describe the recent adaptive convolution approach to video frame interpolation~\cite{Niklaus_CVPR_2017} and define notations. We then describe how we develop our separable convolution-based frame interpolation method. 

Our goal is to interpolate a frame $\hat{I}$ temporally in the middle of the two input video frames $I_1$ and $I_2$. For each output pixel $\hat{I}(x, y)$, the convolution-based interpolation method estimates a pair of 2D convolution kernels $K_1(x, y)$ and $K_2(x, y)$ and uses them to convolve with $I_1$ and $I_2$ to compute the color of the output pixel as follows.
\begin{equation}
    \hat{I}(x, y) = K_1(x, y) \ast P_1(x, y) + K_2(x, y) \ast P_2(x, y)
\end{equation}
where $P_1(x, y)$ and $P_2(x, y)$ are the patches centered at $(x, y)$ in $I_1$ and $I_2$. The pixel-dependent kernels $K_1$ and $K_2$ capture both motion and re-sampling information required for interpolation. To capture large motion, large-size kernels are required. Niklaus~\etal used $41\times 41$ kernels~\cite{Niklaus_CVPR_2017} and it is difficult to estimate them at once for all the pixels of a high-resolution frame simultaneously, due to the large amount of parameters and the limited memory. Their method thus estimates each individual pair of kernels pixel by pixel using a deep convolutional neural network.

Our method addresses this problem by estimating a pair of 1D kernels that approximate a 2D kernel. That is, we estimate $\langle k_{1,v}, k_{1,h} \rangle$ and $\langle k_{2,v}, k_{2,h} \rangle$ to approximate $K_1$ as $k_{1,v} \ast k_{1,h}$ and $K_2$ as $k_{2,v} \ast k_{2,h}$. Thus, our method reduces the number of kernel parameters from $n^2$ to $2n$ for each kernel. This enables the synthesis of a high-resolution video frame in one pass and the incorporation of perceptual loss to further improve the visual quality of the interpolation results, as detailed in the following subsections. 

\subsection{Separable kernel estimation}

We design a fully convolutional neural network that given input frames $I_1$ and $I_2$, estimates two pairs of 1D kernels $\langle k_{1,v}, k_{1,h} \rangle$ and $\langle k_{2,v}, k_{2,h} \rangle$ for each pixel in the output frame $\hat{I}$, as illustrated in Figure~\ref{fig:architecture}. We treat each color channel equally and apply the same 1D kernels to each of the RGB channels to synthesize the output pixel. Note that applying the estimated kernels to the input images is a local convolution and we implement it as a network layer of our neural network similar to a position-varying dynamics convolution layer in recent work~\cite{Finn_NIPS_2016, Jia_NIPS_2016, Xue_NIPS_2016}; therefore our neural network is end-to-end trainable. 

Our neural network consists of a contracting component to extract features and an expanding part that incorporates upsampling layers to perform the dense prediction. We furthermore use skip connections~\cite{Bishop_BOOK_2006, Long_CVPR_2015} to let the expanding layers incorporate features from the contracting part of the neural network, as shown in Figure~\ref{fig:architecture}. To estimate four sets of 1D kernels, we direct the information flow in the last expansion layer into four sub-networks, with each sub-network estimating one of the kernels. We could have modeled this jointly with a combined representation of the four kernels as well; however, we noticed a faster convergence during training when using four sub-networks.

\begin{figure}\centering
    \setlength{\tabcolsep}{0.0cm}
    \setlength{\itemwidth}{2.05cm}

    \begin{tabularx}{\columnwidth}{c @{\hspace{0.05cm}} c @{\hspace{0.05cm}} c @{\hspace{0.05cm}} c}
            \includegraphics[width=\itemwidth]{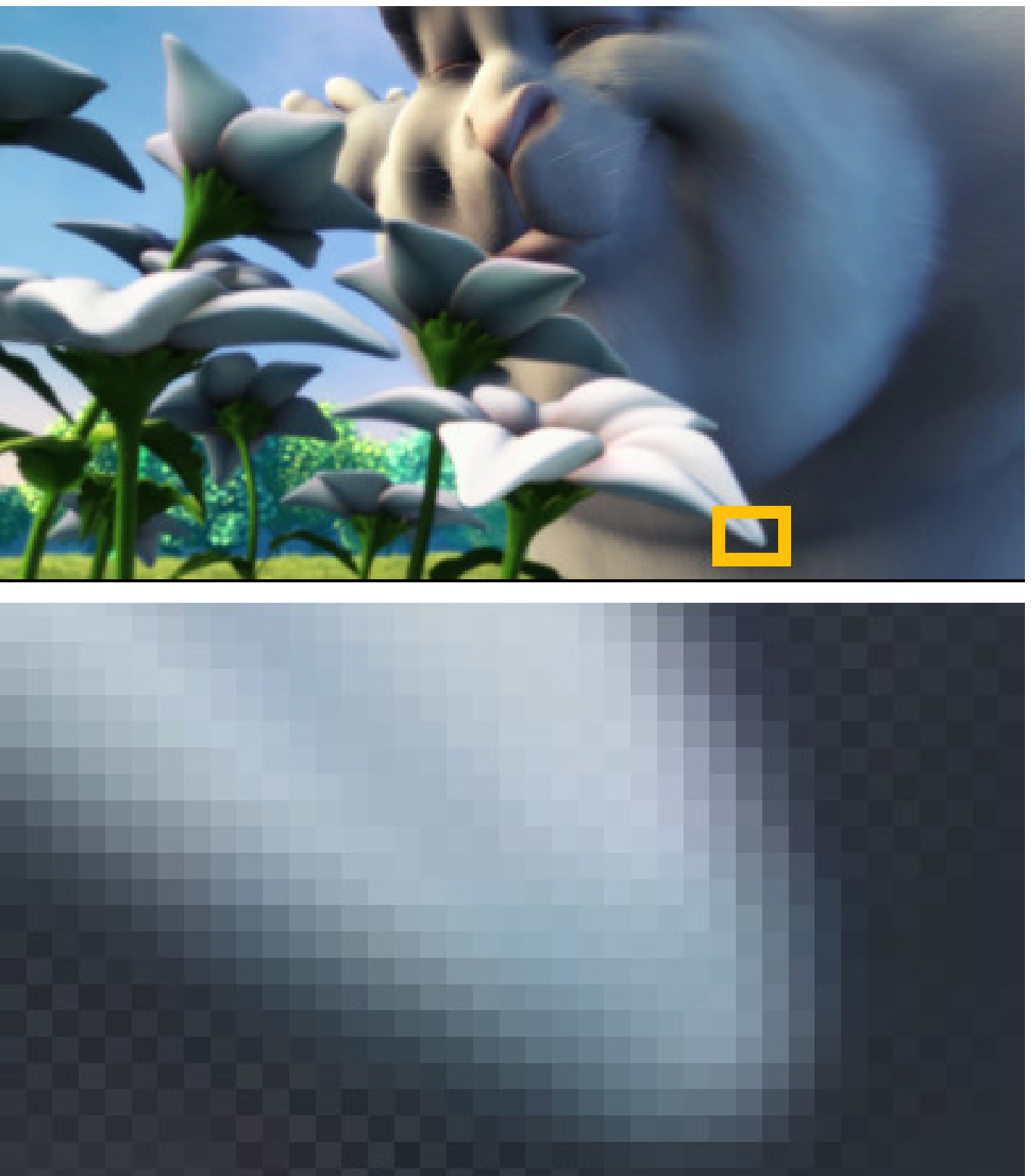}
        &
            \includegraphics[width=\itemwidth]{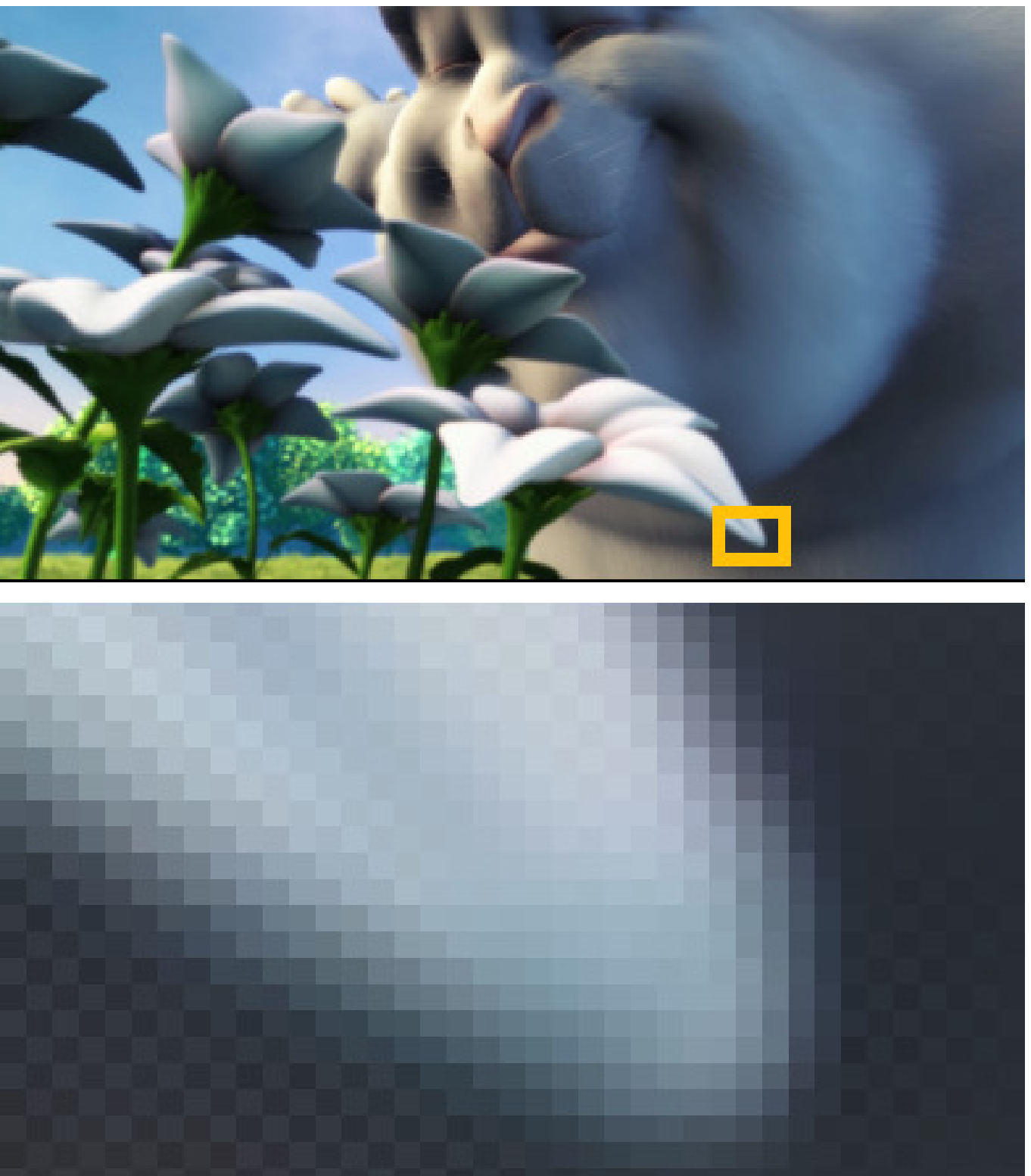}
        &
            \includegraphics[width=\itemwidth]{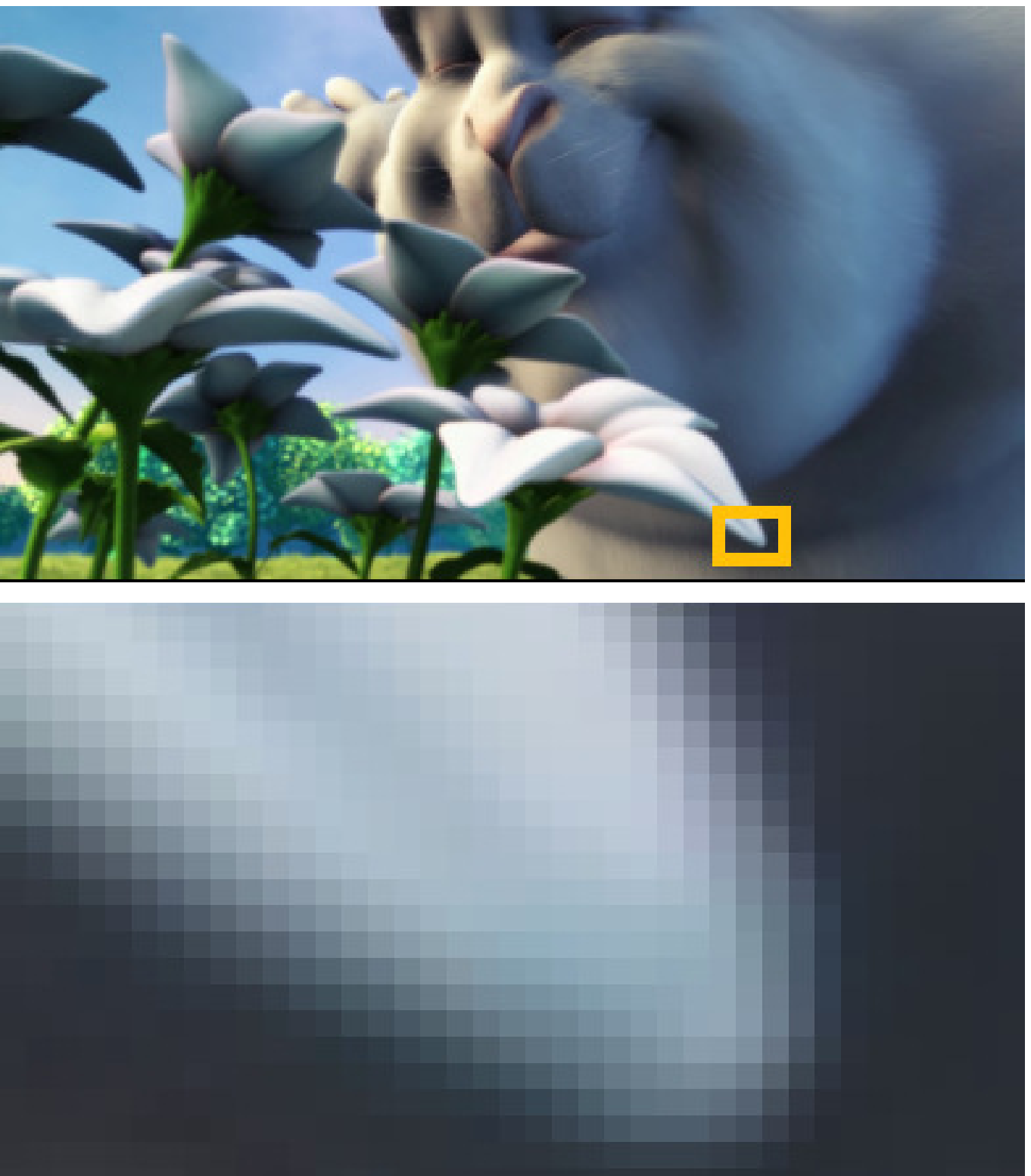}
        &
            \includegraphics[width=\itemwidth]{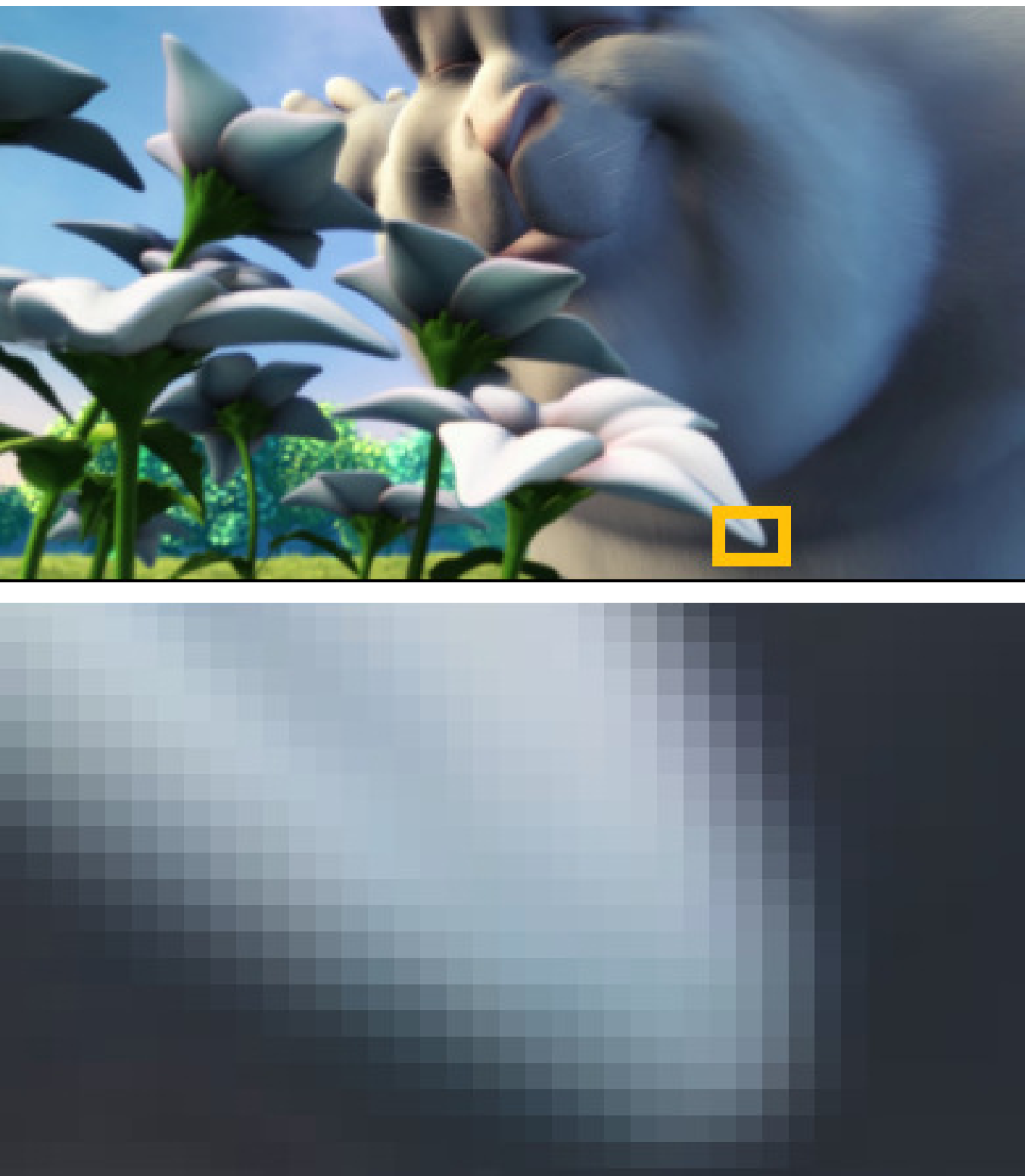}
        \vspace{-0.1cm} \\
            \footnotesize Transposed
        &
            \footnotesize Sub-pixel
        &
            \footnotesize Nearest
        &
            \footnotesize Bilinear
        \\
    \end{tabularx}\vspace{-0.2cm}
    \caption{Checkerboard artifacts.}\vspace{-0.2in}
    \label{fig:checkerboard}
\end{figure} 

We found stacks of $3 \times 3$ convolution layers together with Rectified Linear Units to be effective. Like Zhao~\etal~\cite{Zhao_CORR_2016}, we noticed that the average pooling performs well in the context of pixel-wise predictions and used it in our network accordingly. The upsampling layers in the expanding part can be implemented in various ways, such as transposed convolution, sub-pixel convolution, nearest-neighbor, and bilinear interpolation~\cite{Dong_PAMI_2016, Shi_CVPR_2016, Zeiler_ICCV_2011}. Odena~\etal reported that checkerboard artifacts can occur for image generation tasks if the upsampling layers are not well selected~\cite{Odena_OTHER_2016}. Interestingly, while our method generates images by first estimating convolution kernels, these artifacts can still occur, as shown in Figure~\ref{fig:checkerboard}. We followed the suggestions of Odena~\etal and handled these artifacts by using bilinear interpolation to perform the upsampling in the decoder of our network.

\vspace{0.05in}
\noindent\textbf{Loss function.}
We consider two types of loss functions that measure the difference between an interpolated frame $\hat{I}$ and its corresponding ground truth $I_{gt}$. The first one is $\ell_1$ norm based on per-pixel color difference, as defined below.
\begin{equation}
    \mathcal{L}_1 = \left\| \hat{I} - I_{gt} \right\| _1
\end{equation}
Alternatively the $\ell_2$ norm can be used; however, we found it often leads to blurry results, as also reported in other image generation tasks~\cite{Goroshin_NIPS_2015, Long_ECCV_2016, Mathieu_ICLR_2016, Ranzato_CORR_2014, Srivastava_ICML_2015}.

The second type of loss functions that this work explores is perceptual loss, which has often been found effective in producing visually pleasing images~\cite{Dosovitskiy_NIPS_2016, Johnson_ECCV_2016, Ledig_CORR_2016, Sajjadi_CORR_2016, Zhu_ECCV_2016}. Perceptual loss is usually based on high-level features of images and is defined as follows.
\begin{equation}
    \mathcal{L}_F = \left\| \phi(\hat{I}) - \phi(I_{gt}) \right\| _2^2
\end{equation}
where $\phi$ extracts features from an image. We tried various loss functions based on different feature extractors, such as SSIM loss~\cite{Ridgeway_CORR_2015} and feature reconstruction loss~\cite{Johnson_ECCV_2016}. We empirically found that the feature reconstruction loss based on the \verb|relu4_4| layer of the VGG-19 network~\cite{Simonyan_CORR_2014} produces good results for our frame interpolation task. 

\subsection{Training}

We initialized our neural network parameters using a convolution aware initialization method~\cite{Aghajanyan_CORR_2017} and trained it using AdaMax~\cite{Kingma_CORR_2014} with $\beta_1 = 0.9$, $\beta_2 = 0.999$, a learning rate of 0.001 and a mini-batch size of 16 samples. We chose a small mini-batch size since we experienced a degradation in the quality of the trained model as described by Keskar~\etal~\cite{Keskar_CORR_2016} when using more samples per mini-batch. We used patches of size $128 \times 128$ instead of training on entire frames. This allows us to avoid patches that contain no useful information and leads to diverse mini-batches, which as described by Bansal~\etal~\cite{Bansal_CORR_2017} improves training.

\vspace{0.05in}
\noindent\textbf{Training dataset.}
We extracted training samples from widely available videos, where each training sample consists of three consecutive frames with the middle frame serving as the ground truth. Since the video quality has a great influence on the quality of the trained model, we acquired video material from selected YouTube channels such as ``Tom Scott'', ``Casey Neistat'', ``Linus Tech Tips'', and ``Austin Evans'', whose videos consistently have a high-quality. Note that we downloaded these videos with a resolution of $1920 \times 1080$ but scale them to $1280 \times 720$ in order to reduce the influence of video compression.

Following Niklaus~\etal~\cite{Niklaus_CVPR_2017}, we did not use samples that span across video shot boundaries and discarded samples with a lack of texture. To increase the diversity of our training dataset, we avoided samples that are temporally close to each other. Instead of using the full frames, we randomly cropped $150 \times 150$ patches and selected those with sufficiently large motion. To compute the motion in each sample, we estimated the mean optical flow between the first and the last patch using SimpleFlow~\cite{Tao_OTHER_2012}.

We composed our dataset from the extracted samples by randomly selecting $250,000$ of them without replacement. The random selection was guided by the estimated mean optical flow, making sure that samples with a large flow magnitude were more likely to be included. Overall, 10\% of the pixels in the resulting training dataset have a flow magnitude of at least $17$ pixels and 5\% of them have a magnitude of at least $23$ pixels. The largest motion is $39$ pixels.

\vspace{0.05in}
\noindent\textbf{Data augmentation.}
We performed data augmentation on the fly during training. While each sample in the training dataset is of size $150 \times 150$ pixels, we used patches with a size of $128 \times 128$ pixels for training. This makes it possible to perform data augmentation by random cropping, preventing the network from learning spatial priors that potentially exist in the training dataset. We augmented the motion of each sample by shifting the cropped windows in the first and last frames while leaving the cropped window of the ground truth unchanged. By doing this systematically and shifting the cropped windows of the first and last frames in opposite directions, the ground truth will still be sound. We found that performing shifts by up to $6$ pixels works well, which augments the flow magnitude by approximately $8.5$ pixels. We also randomly flipped the cropped patches horizontally or vertically and randomly swap their temporal order, which makes the motion within the training dataset symmetric and prevents the network from being biased.

\subsection{Implementation details}

Below we discuss implementation details with respect to speed, boundary handling, and hyper-parameter selection.

\vspace{0.05in}
\noindent\textbf{Computational efficiency.}
We used Torch~\cite{Collobert_OTHER_2011} to implement our convolutional neural network. To achieve a high computational efficiency and allow our network to directly render the interpolated frame, we wrote our own layer in CUDA that applies the estimated 1D kernels. If applicable, we used implementations based on cuDNN~\cite{Chetlur_CORR_2014} for the other layers of the network to further improve the speed. With a Nvidia Titan X (Pascal), our system is able to interpolate a $1280 \times 720$ frame in $0.5$ seconds as well as a $1920 \times 1080$ frame in $0.9$ seconds. Training our network takes about $20$ hours using four Nvidia Titan X (Pascal).

\vspace{0.05in}
\noindent\textbf{Boundary handling.}
Due to the utilized convolution-based interpolation formulation, the input needs to be padded such that boundary pixels can be processed. We tried zero padding, reflective padding, and padding by repetition. Empirically, we found padding by repetition to work well and used it accordingly. Note that boundary handling is not needed during training, where an output with a reduced size is still acceptable.

\vspace{0.05in}
\noindent\textbf{Hyper-parameter selection.}
We used a validation dataset in order to select reasonable hyper-parameters for our network architecture as well as for the training. This validation dataset is disjoint from the training dataset but has been created in a similar manner.

Besides common parameters such as the learning rate, our model comes with a crucial domain-specific hyper-parameter, which is the size of the 1D kernels for interpolation. We found kernels of size $51$ pixels to work well, which we attribute to the largest flow magnitude in the dataset, $39$ pixels, together with $8.5$-pixels of extra motion from augmentation. While increasing the kernel size is desirable to handle larger motion, restricted by the flow in our dataset, we did not observe improvements with larger kernels. 

Another important hyper-parameter for our method is the number of pooling layers. Pooling layers have a great influence on the receptive field~\cite{Luo_NIPS_2016} of a convolutional neural network, which in our context is related to the aperture problem in motion estimation. A larger number of pooling layers increases the receptive field to potentially handle large motion; on the other hand, the largest flow magnitude in the training dataset provides an upper bound for the number of useful pooling layers. Empirically, we found using five pooling layers produces good interpolation results.

\section{Experiments}
\label{sec:exp}
We compare our method with representative state-of-the-art methods and evaluate them both qualitatively and quantitatively. For the optical flow based methods, we selected MDP-Flow2~\cite{Xu_PAMI_2012}, which currently achieves the lowest interpolation error in the Middlebury benchmark and DeepFlow2~\cite{Weinzaepfel_ICCV_2013}, which is the neural network based approach with the lowest interpolation error~\cite{Baker_OTHER_2011}. We follow recent frame interpolation papers~\cite{Liu_CORR_2017, Meyer_CVPR_2015} and use the algorithm from the Middlebury benchmark~\cite{Baker_OTHER_2011} to synthesize frames from the estimated optical flow. We also compare our method with the phase-based frame interpolation method from Meyer~\etal~\cite{Meyer_CVPR_2015} as well as the AdaConv method based on adaptive convolution from Niklaus~\etal~\cite{Niklaus_CVPR_2017} as alternatives to optical flow based methods. For all these methods, we use the code or trained models from the original papers. Please refer to our video for more results.

\begin{figure}\centering
    \setlength{\tabcolsep}{0.0cm}
    \setlength{\itemwidth}{2.75cm}

    \begin{tabularx}{\textwidth}{c @{\hspace{0.05cm}} c @{\hspace{0.05cm}} c}
            \includegraphics[width=\itemwidth]{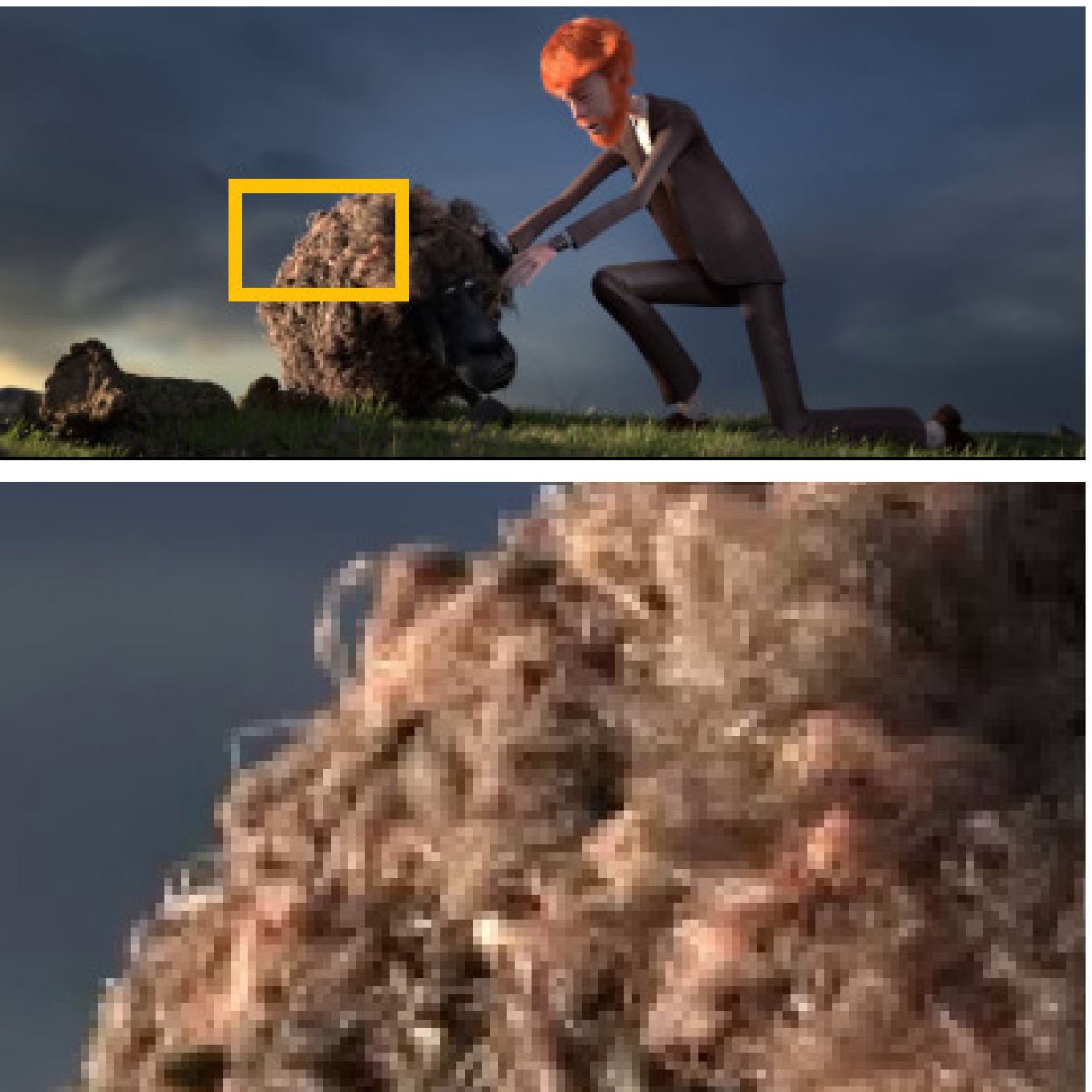}
        &
            \includegraphics[width=\itemwidth]{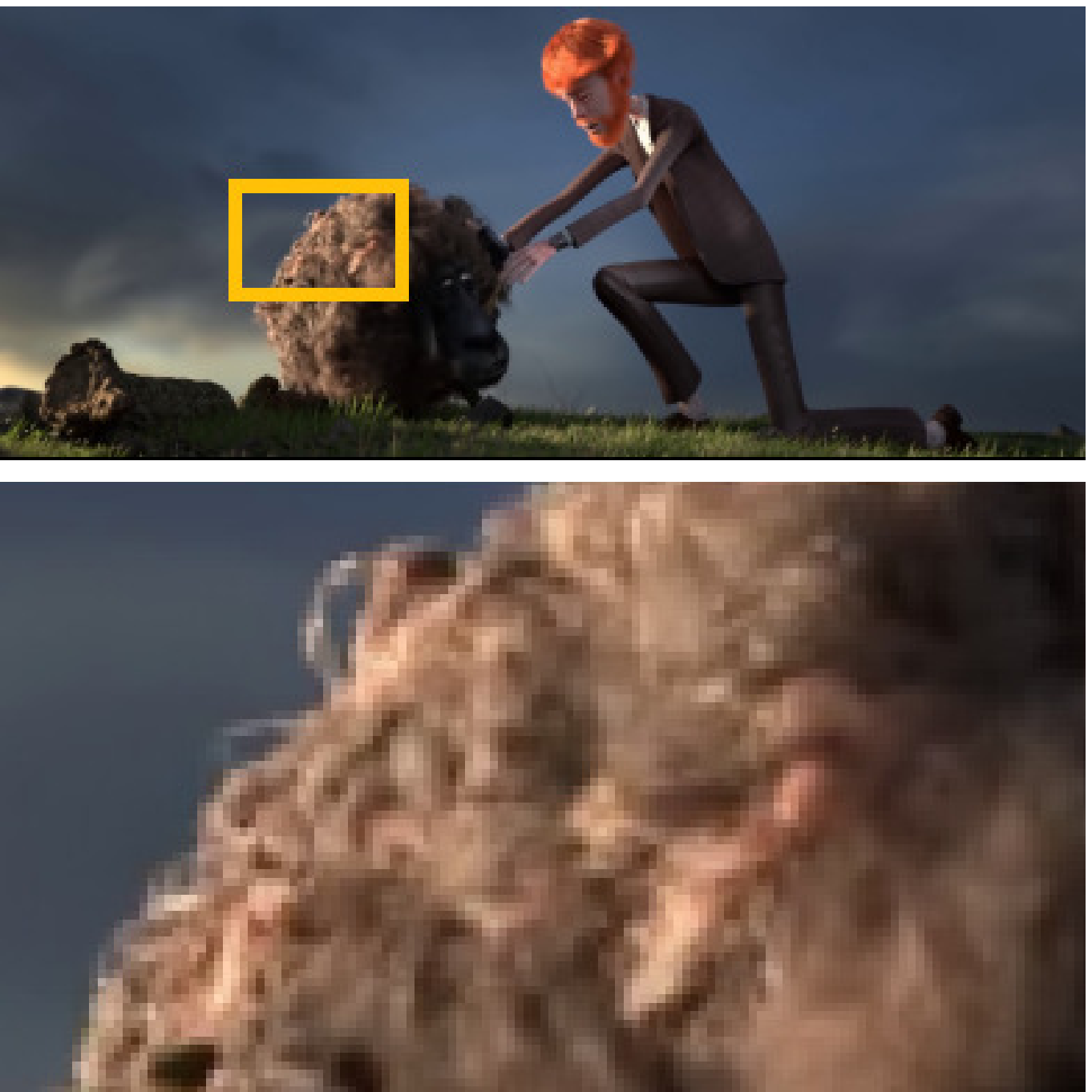}
        &
            \includegraphics[width=\itemwidth]{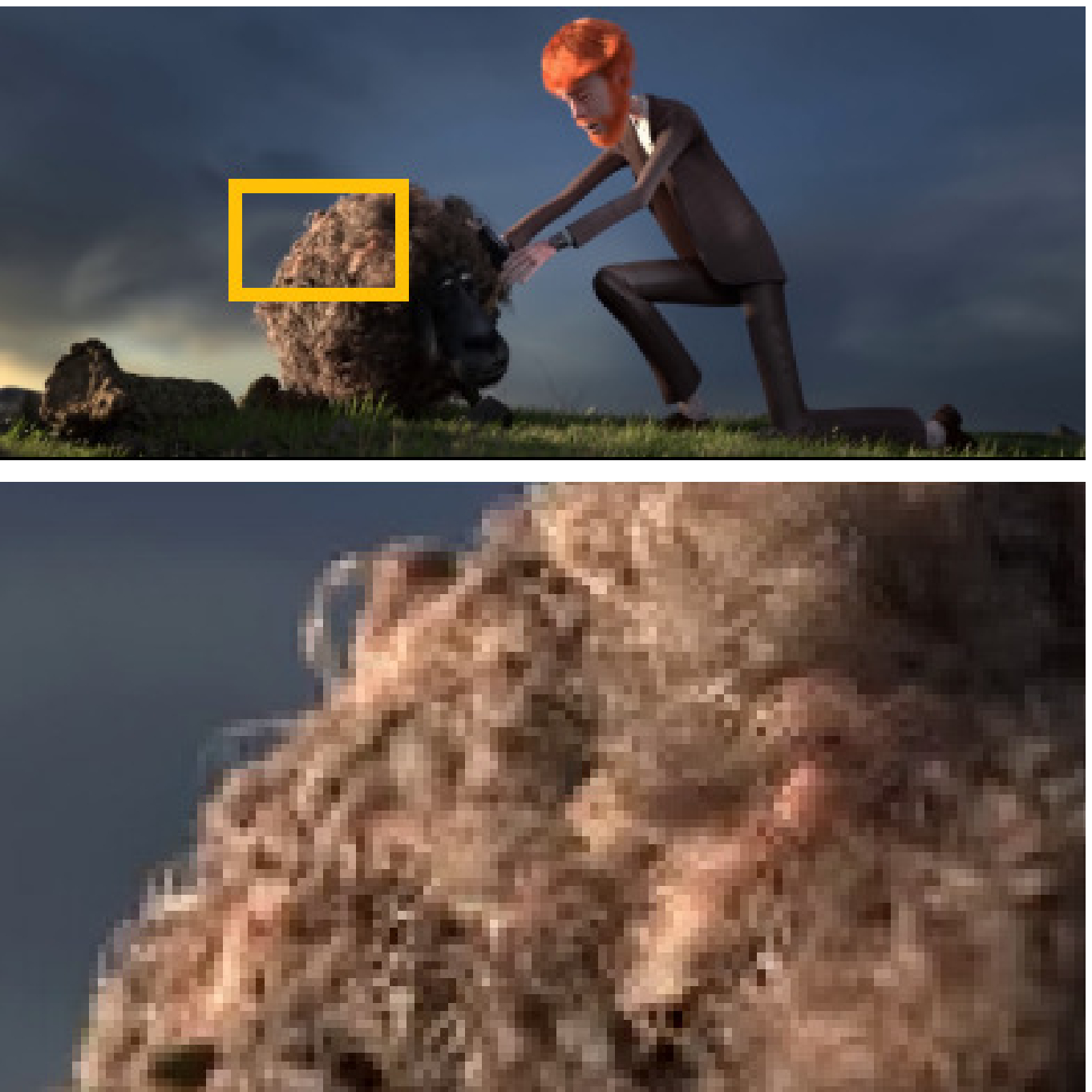}
        \vspace{-0.1cm} \\
    \end{tabularx}
    \begin{tabularx}{\textwidth}{c @{\hspace{0.05cm}} c @{\hspace{0.05cm}} c}
            \includegraphics[width=\itemwidth]{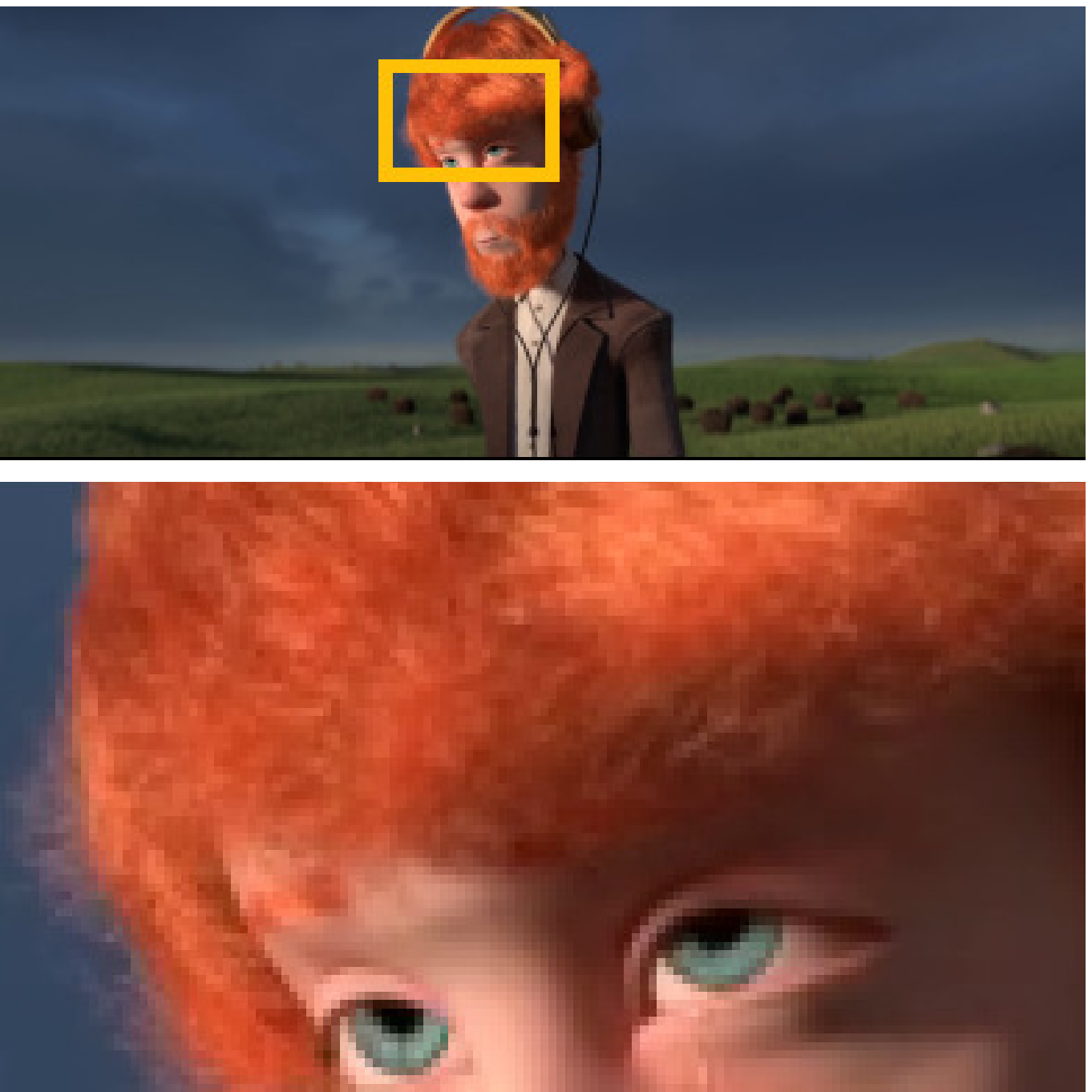}
        &
            \includegraphics[width=\itemwidth]{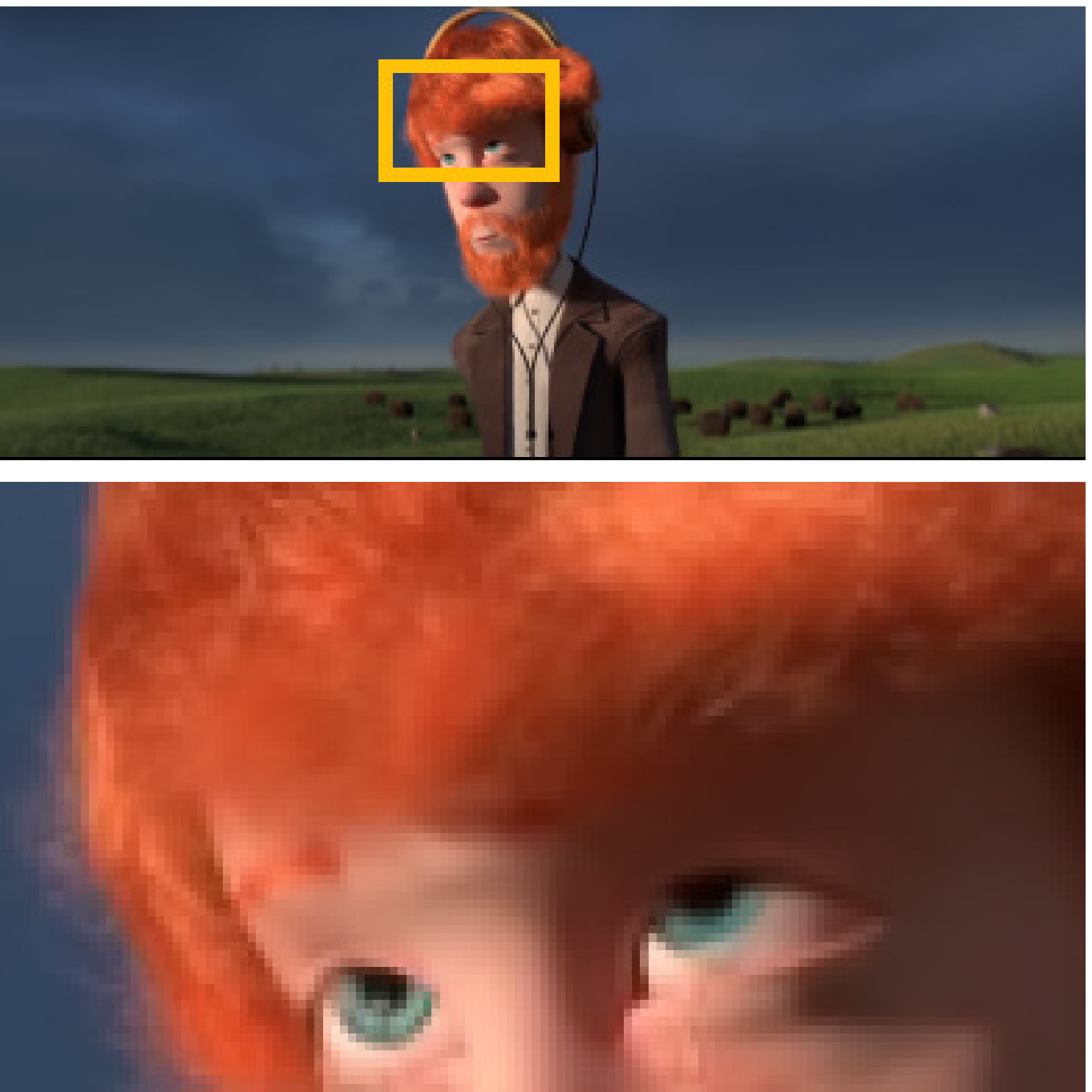}
        &
            \includegraphics[width=\itemwidth]{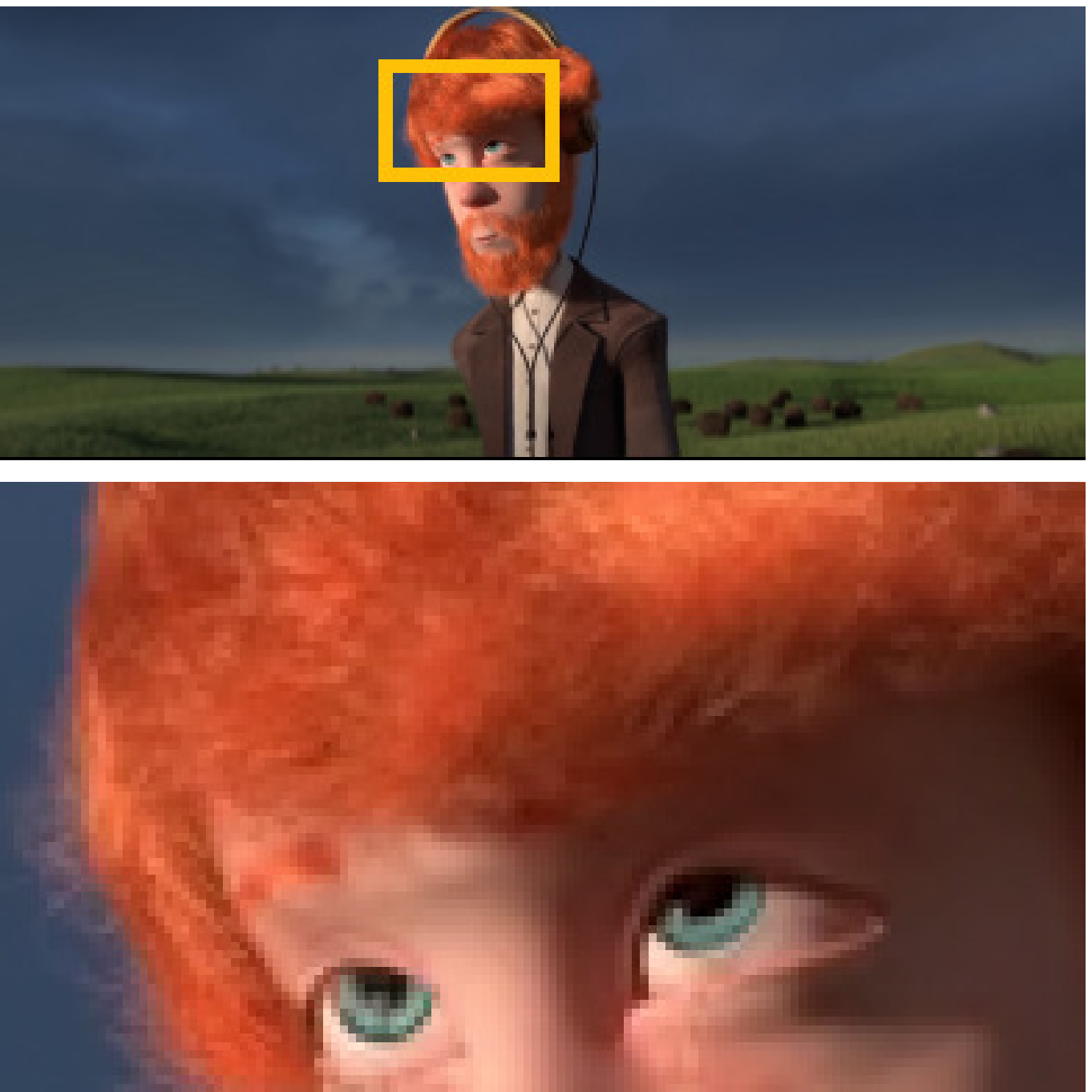}
        \vspace{-0.1cm} \\
            \footnotesize Input frame 1
        &
            \footnotesize Ours - $\mathcal{L}_1$
        &
            \footnotesize Ours - $\mathcal{L}_F$
        \\
    \end{tabularx}\vspace{-0.3cm}
    \caption{The effect of loss functions.}\vspace{-0.2in}
    \label{fig:loss}
\end{figure}

\subsection{Loss functions}

Our method incorporates two types of loss functions: $\mathcal{L}_1$ loss and feature reconstruction loss $\mathcal{L}_F$. To examine their effect, we trained two versions of our neural network. For the first one, we only used $\mathcal{L}_1$ loss and refer to this network as ``$\mathcal{L}_1$" for simplicity in this paper. For the second one, we used both $\mathcal{L}_1$ loss and $\mathcal{L}_F$ loss and refer to this network as ``$\mathcal{L}_F$" for simplicity. We tried different training schemes, including using linear combinations of $\mathcal{L}_1$ and $\mathcal{L}_F$ with different weights, and first training the network with $\mathcal{L}_1$ loss and then fine tuning it using $\mathcal{L}_F$ loss. We found that the latter leads to the best visual quality and used this scheme accordingly. As shown in Figure~\ref{fig:loss}, incorporating $\mathcal{L}_F$ loss leads to sharper images with more high frequency details. This is in line with the findings in recent work on image generation and super resolution~\cite{Dosovitskiy_NIPS_2016, Johnson_ECCV_2016, Ledig_CORR_2016, Sajjadi_CORR_2016, Zhu_ECCV_2016}.

\begin{figure*}\centering
    \setlength{\tabcolsep}{0.0cm}
    \setlength{\itemwidth}{2.45cm}

    \begin{tabularx}{\textwidth}{c @{\hspace{0.05cm}} c @{\hspace{0.05cm}} c @{\hspace{0.05cm}} c @{\hspace{0.05cm}} c @{\hspace{0.05cm}} c @{\hspace{0.05cm}} c}
            \includegraphics[width=\itemwidth]{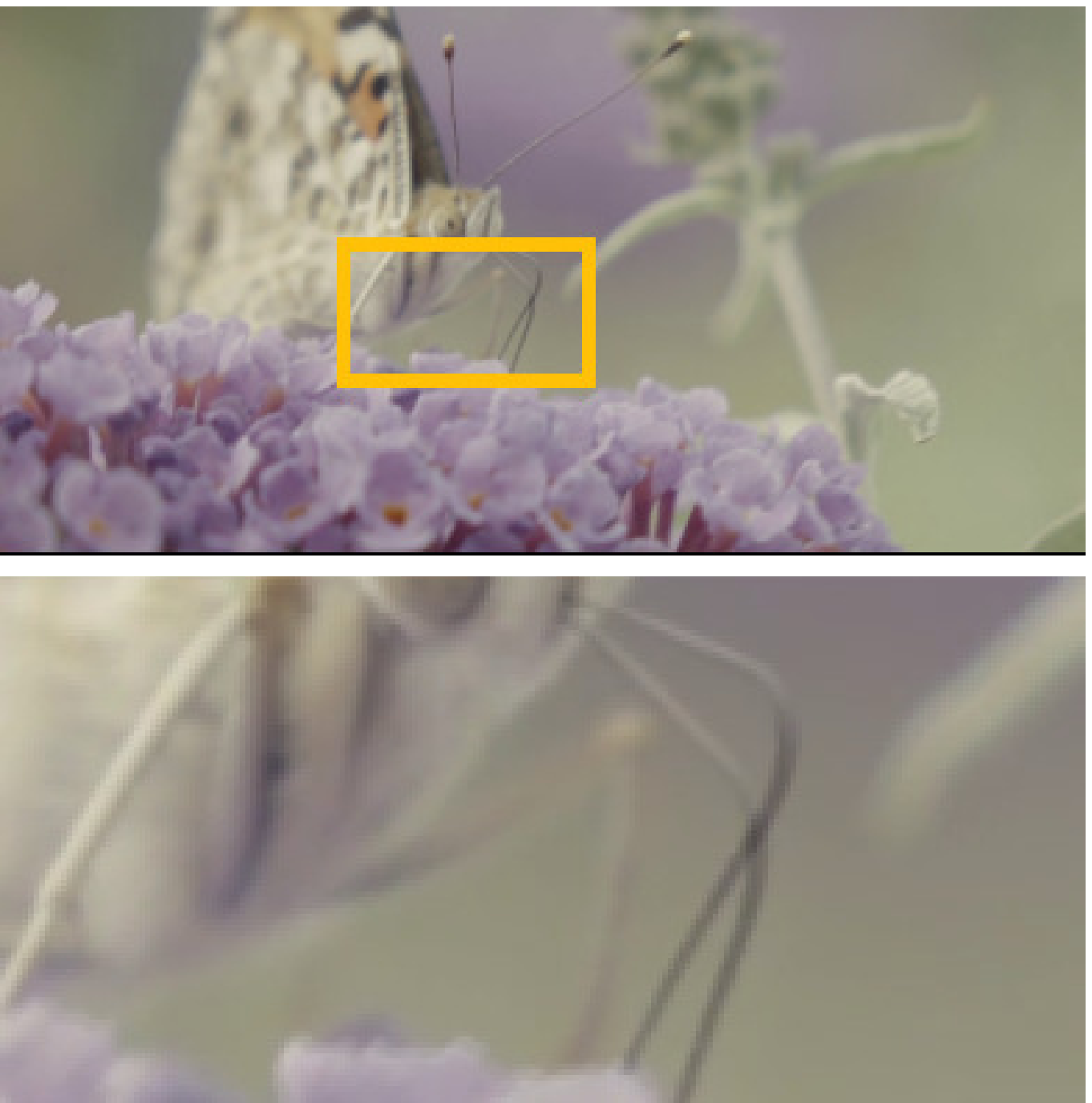}
        &
            \includegraphics[width=\itemwidth]{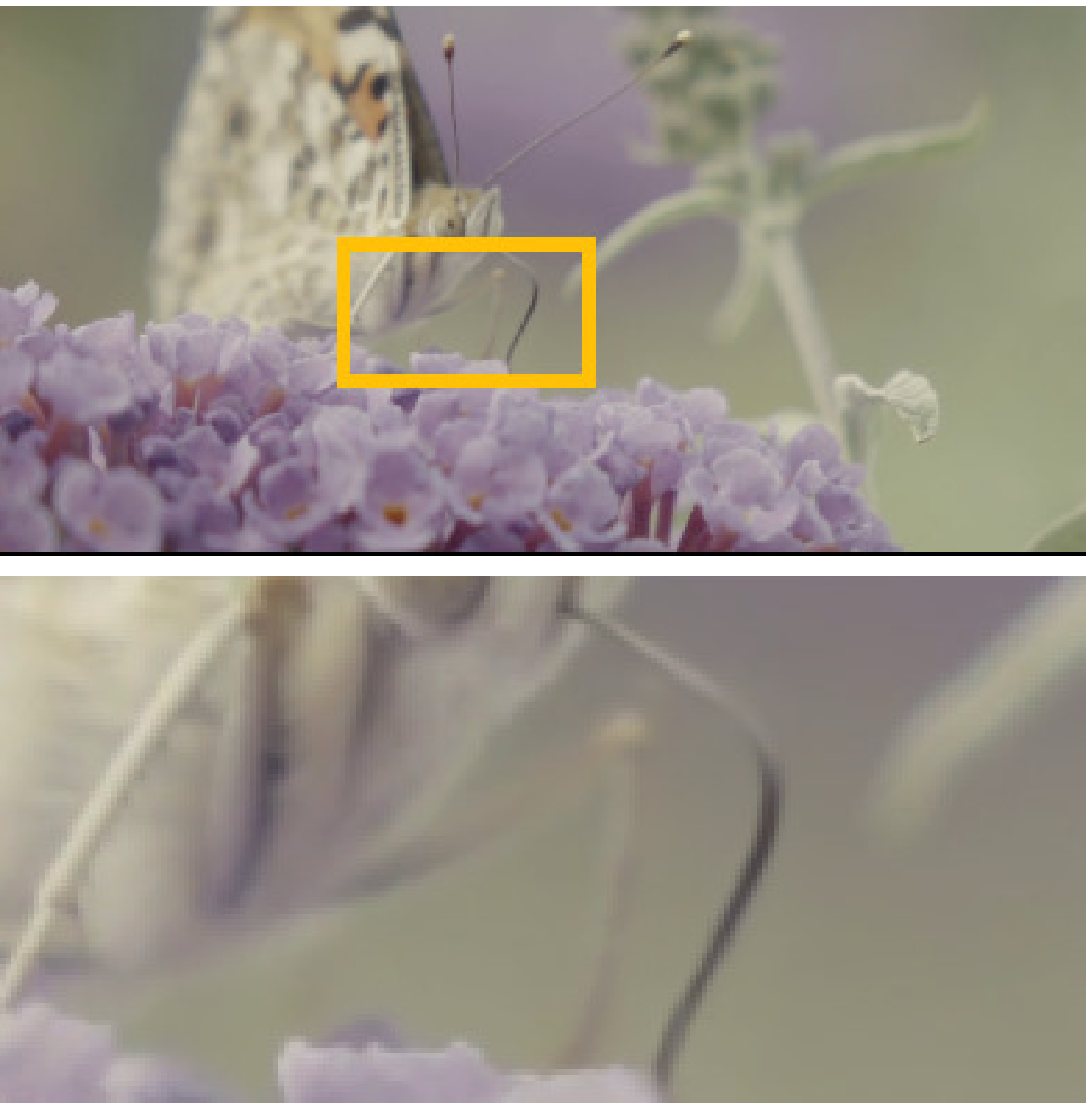}
        &
            \includegraphics[width=\itemwidth]{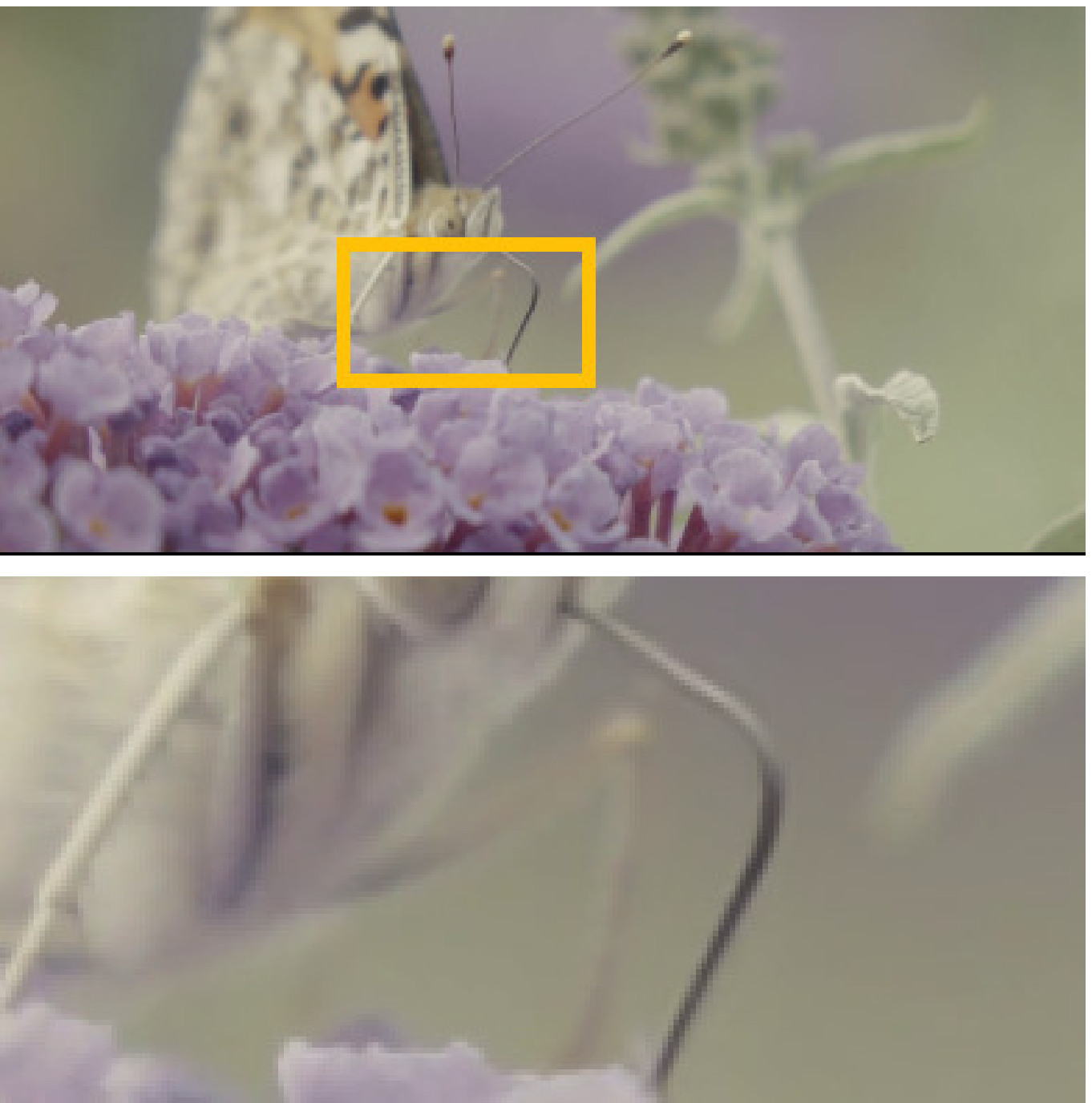}
        &
            \includegraphics[width=\itemwidth]{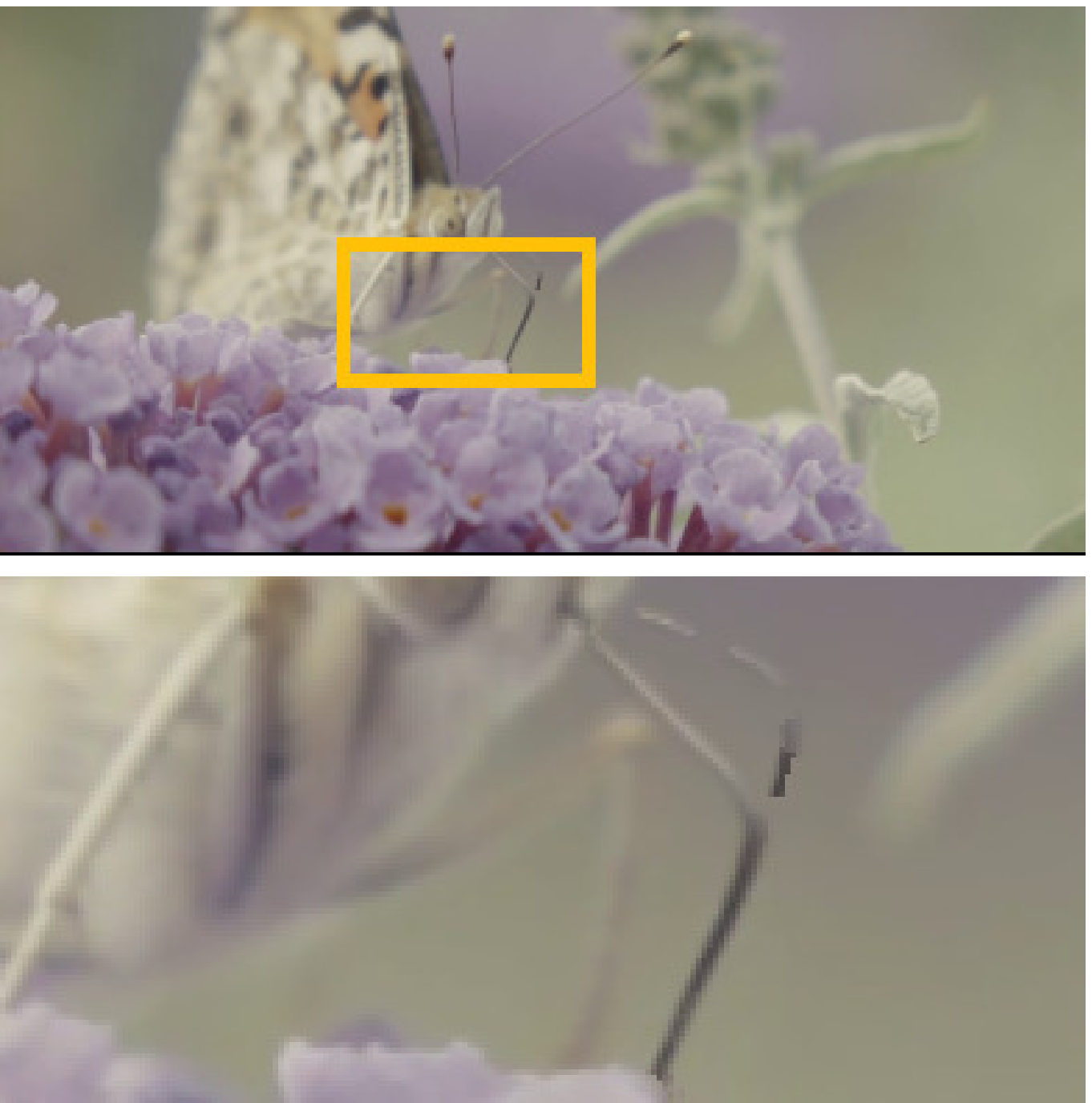}
        &
            \includegraphics[width=\itemwidth]{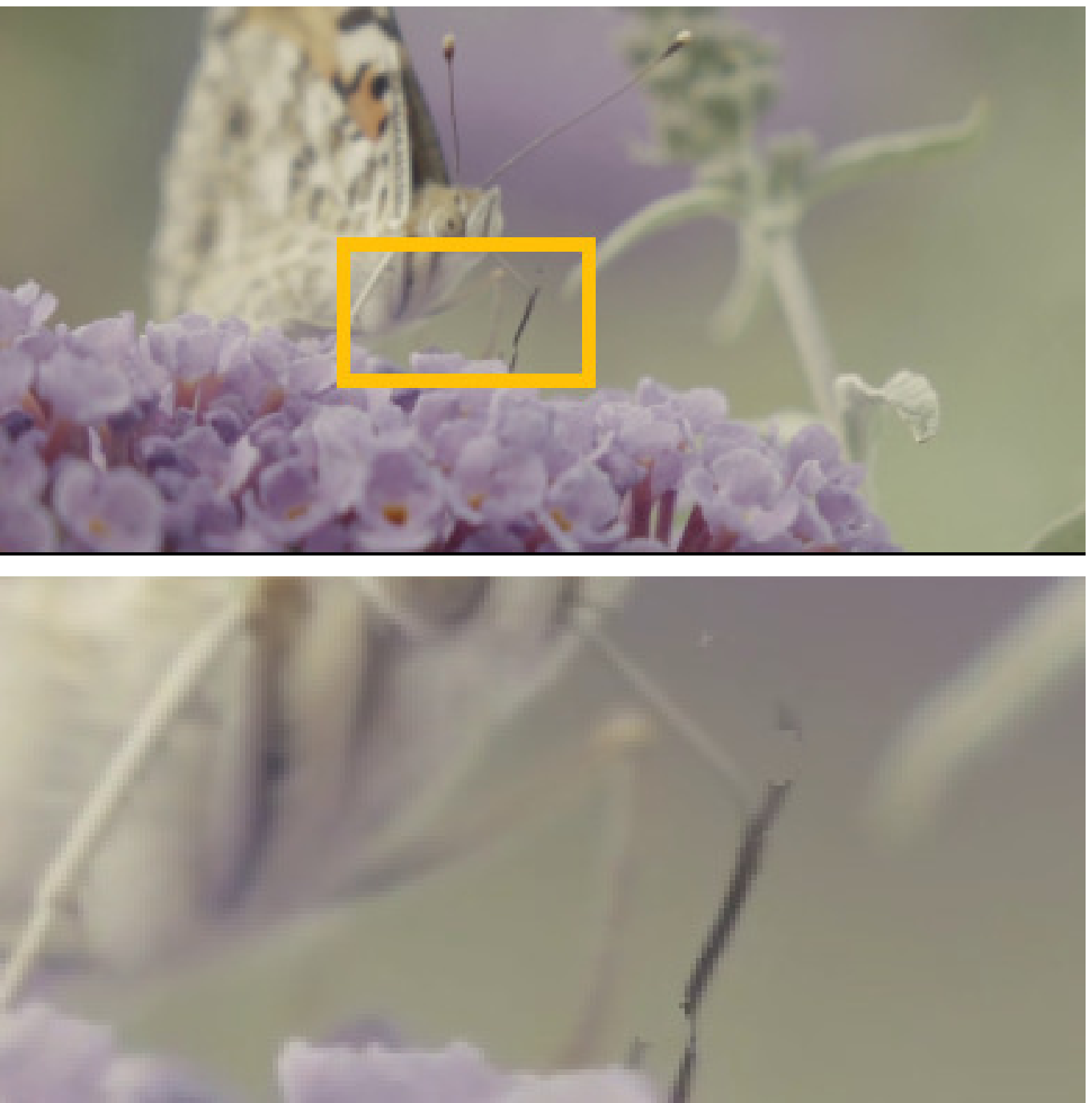}
        &
            \includegraphics[width=\itemwidth]{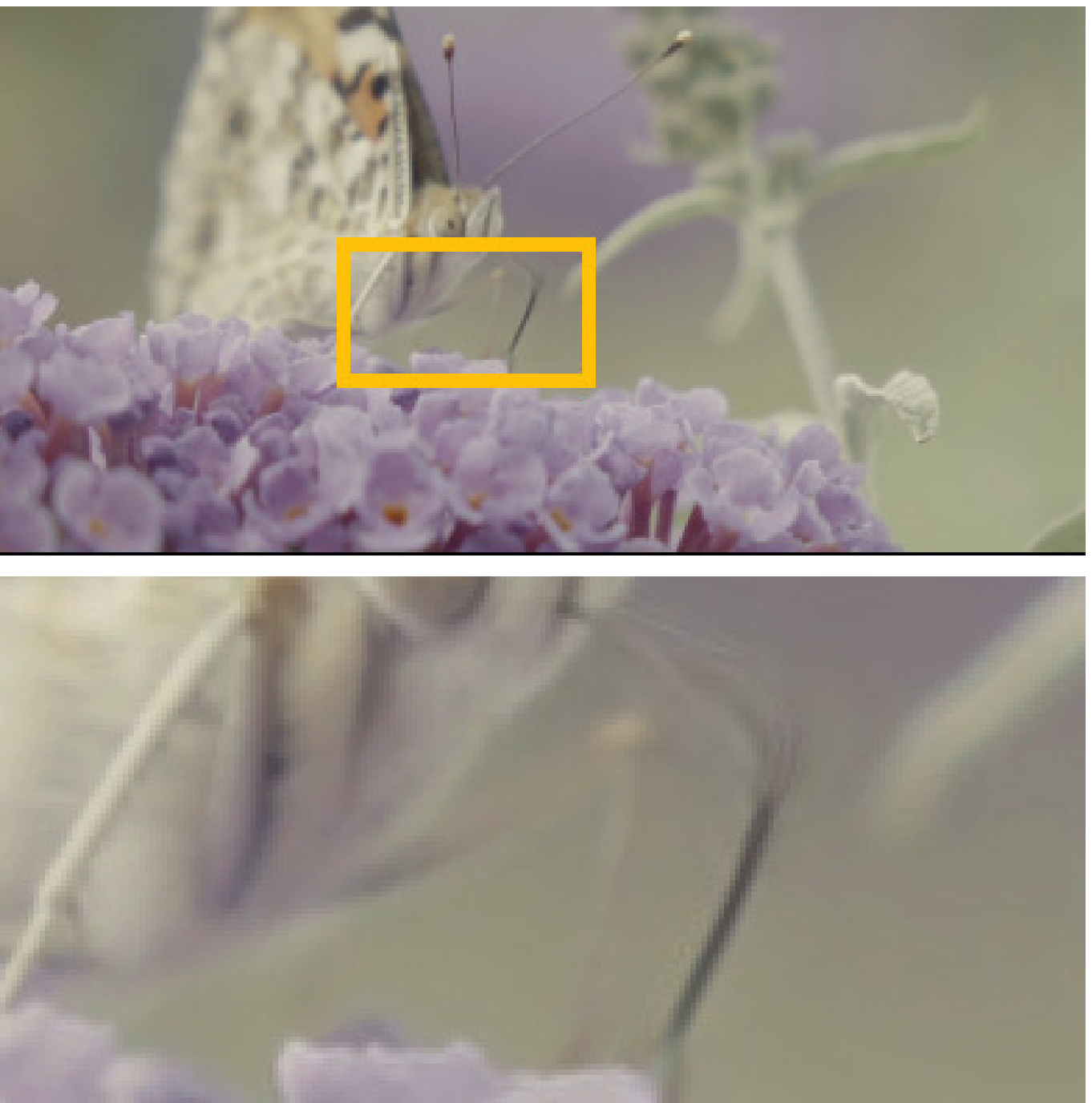}
        &
            \includegraphics[width=\itemwidth]{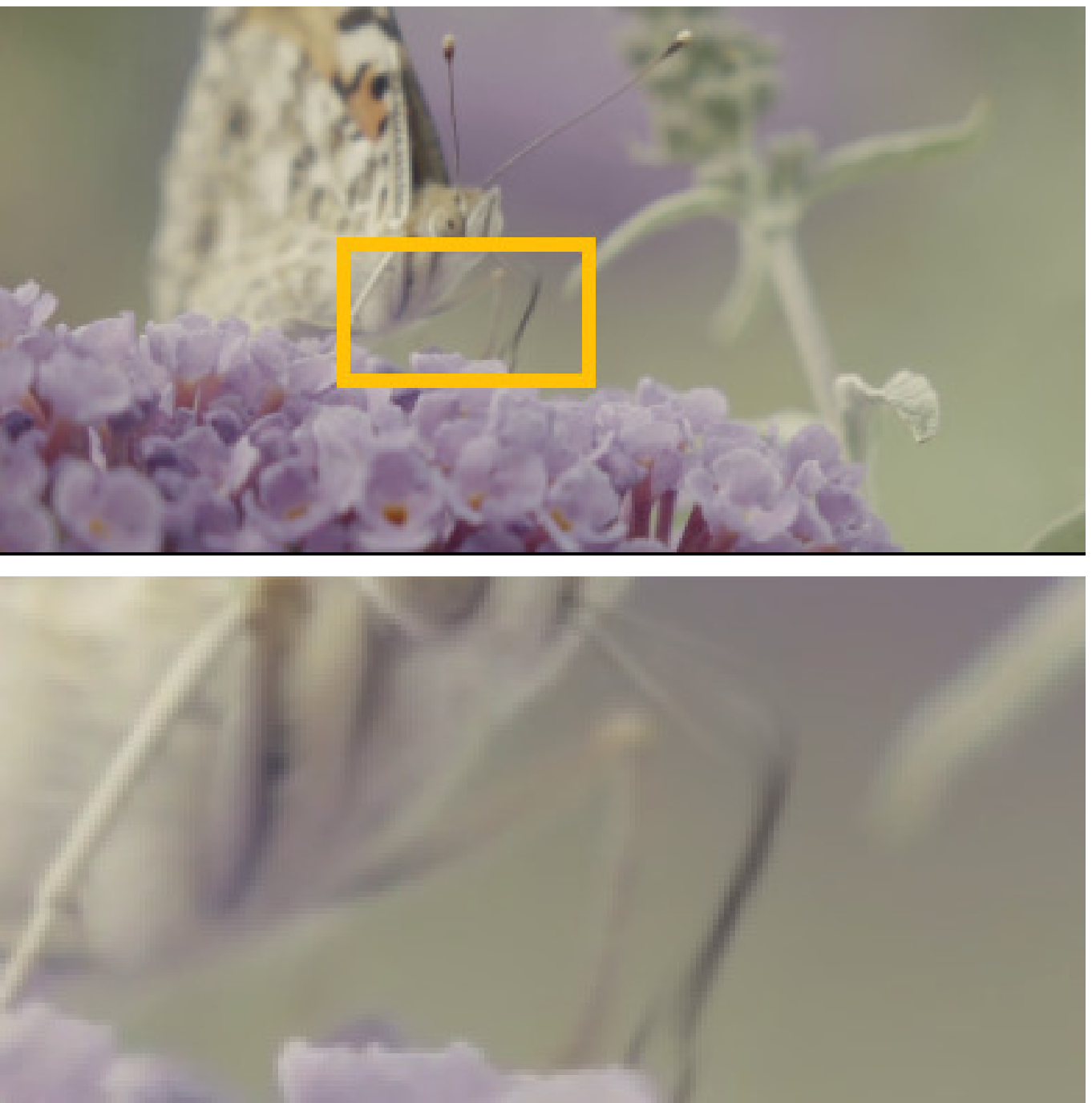}
        \vspace{-0.1cm} \\
    \end{tabularx}
    \begin{tabularx}{\textwidth}{c @{\hspace{0.05cm}} c @{\hspace{0.05cm}} c @{\hspace{0.05cm}} c @{\hspace{0.05cm}} c @{\hspace{0.05cm}} c @{\hspace{0.05cm}} c}
            \includegraphics[width=\itemwidth]{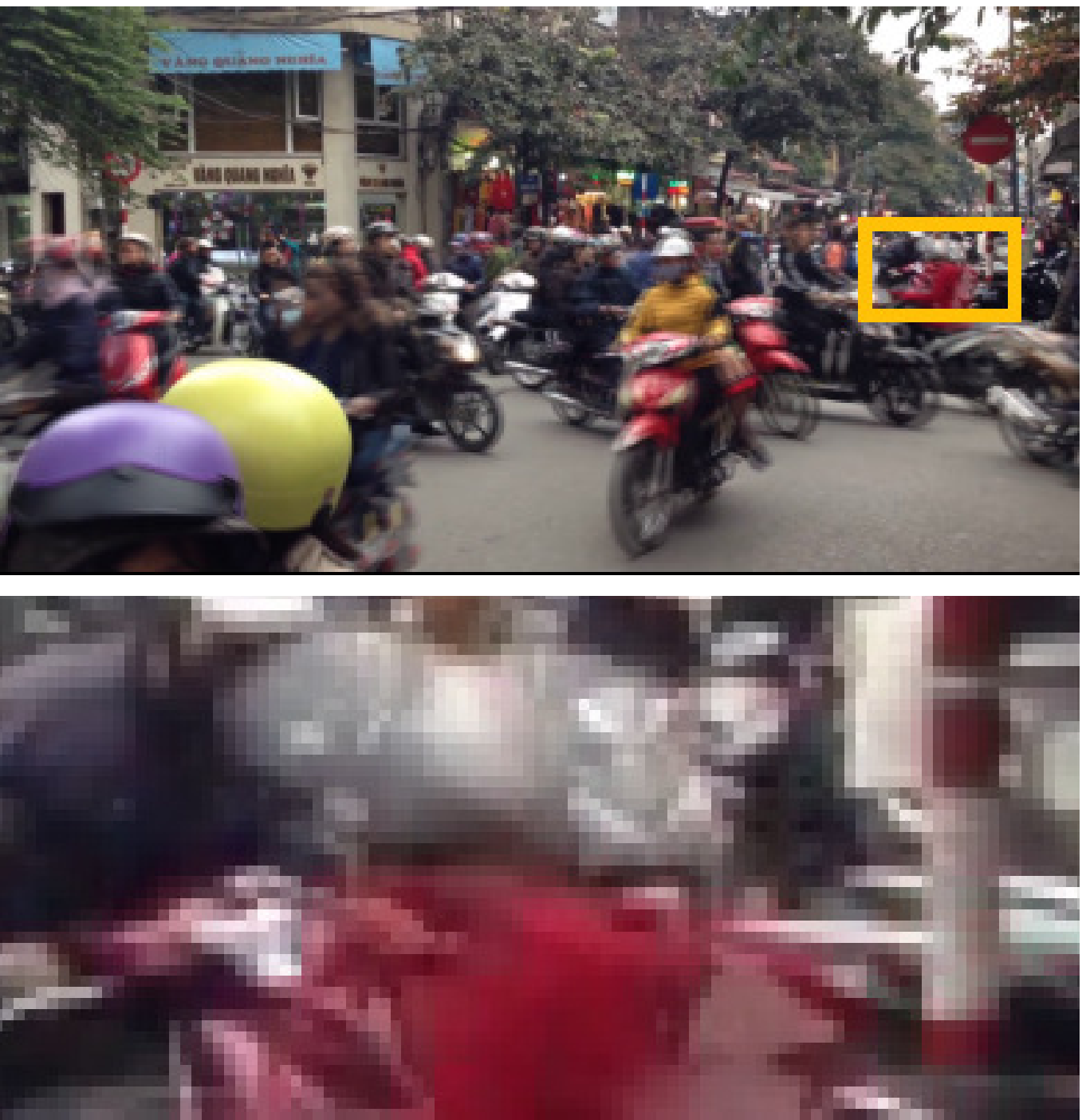}
        &
            \includegraphics[width=\itemwidth]{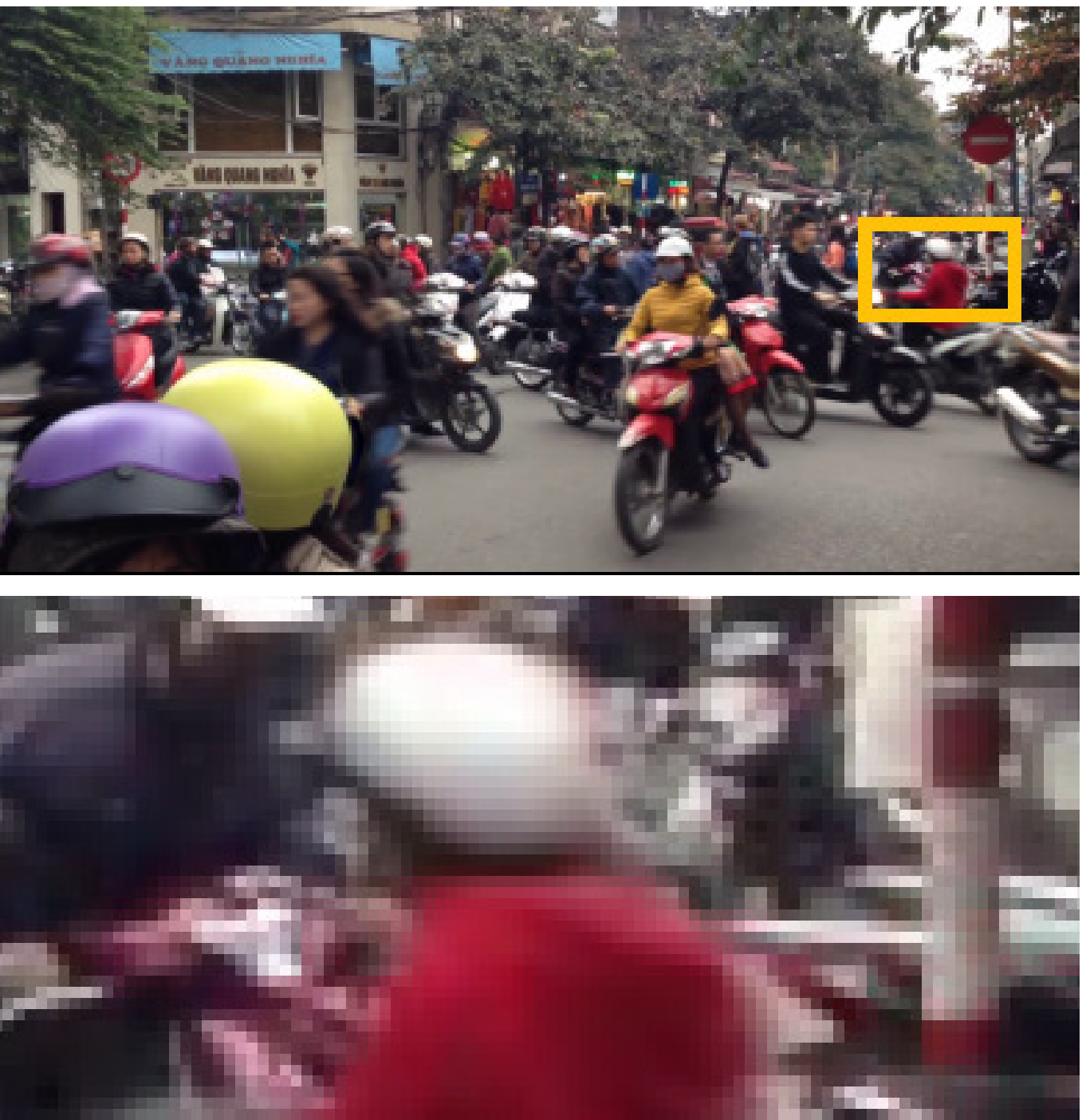}
        &
            \includegraphics[width=\itemwidth]{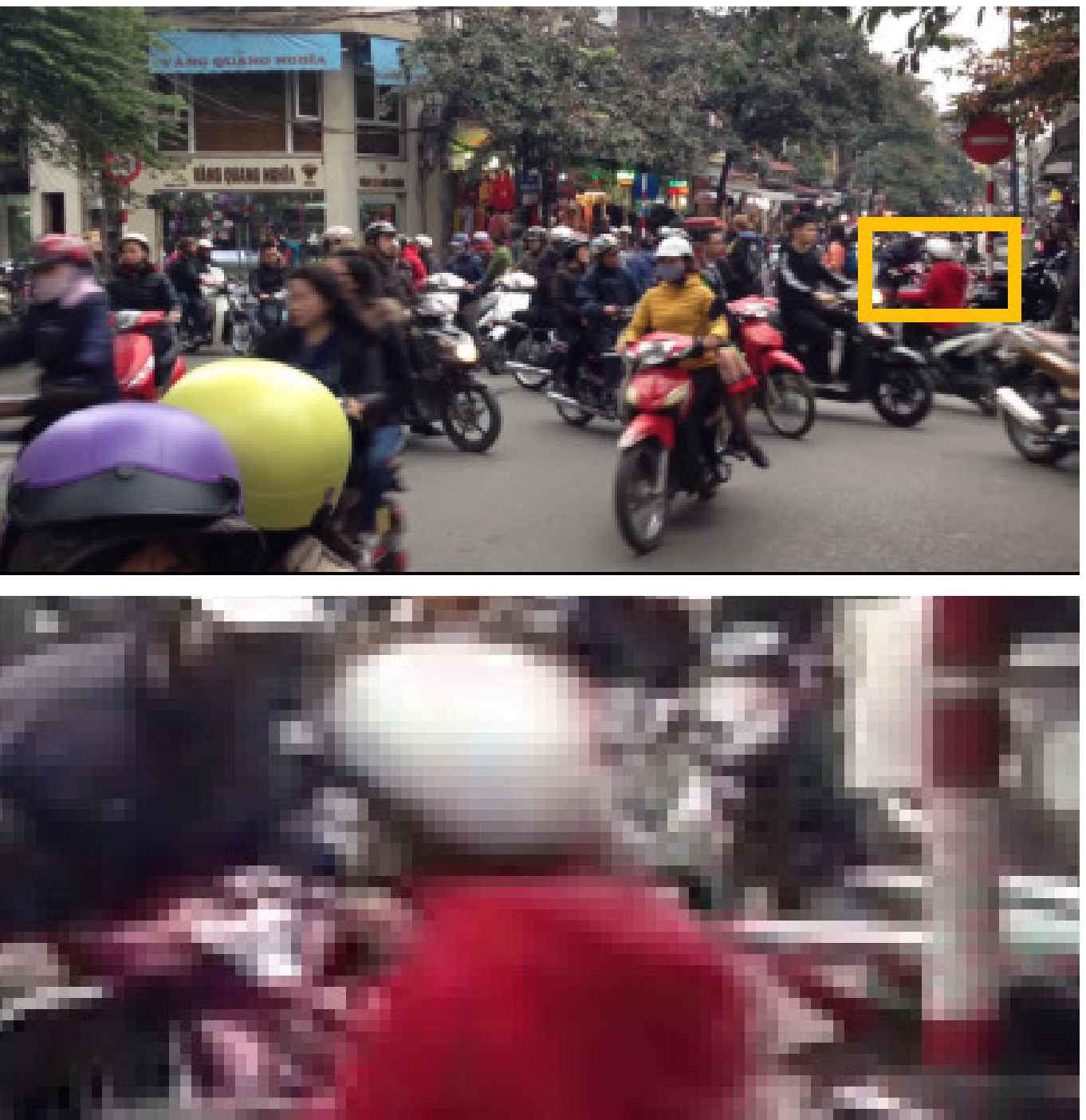}
        &
            \includegraphics[width=\itemwidth]{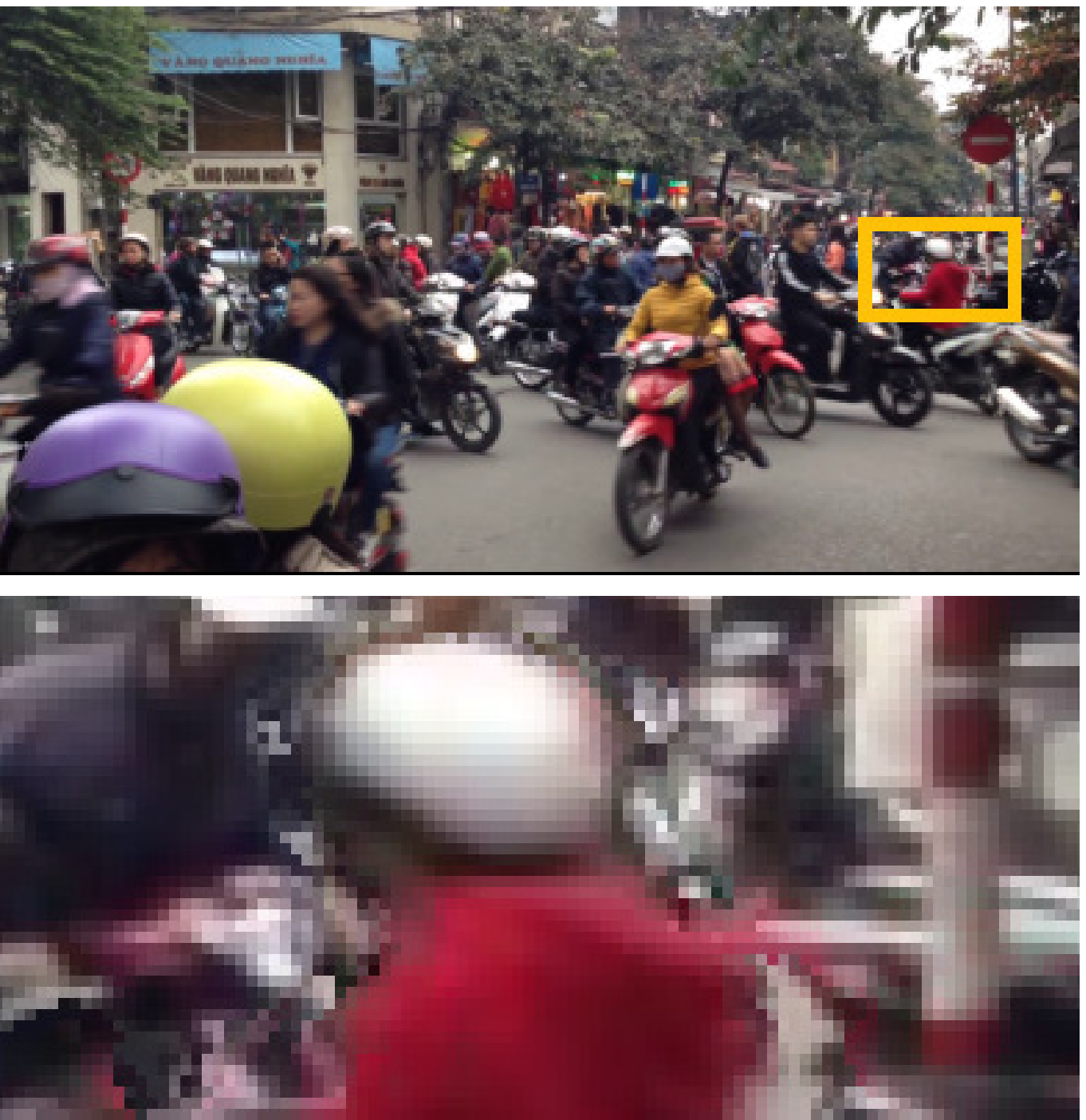}
        &
            \includegraphics[width=\itemwidth]{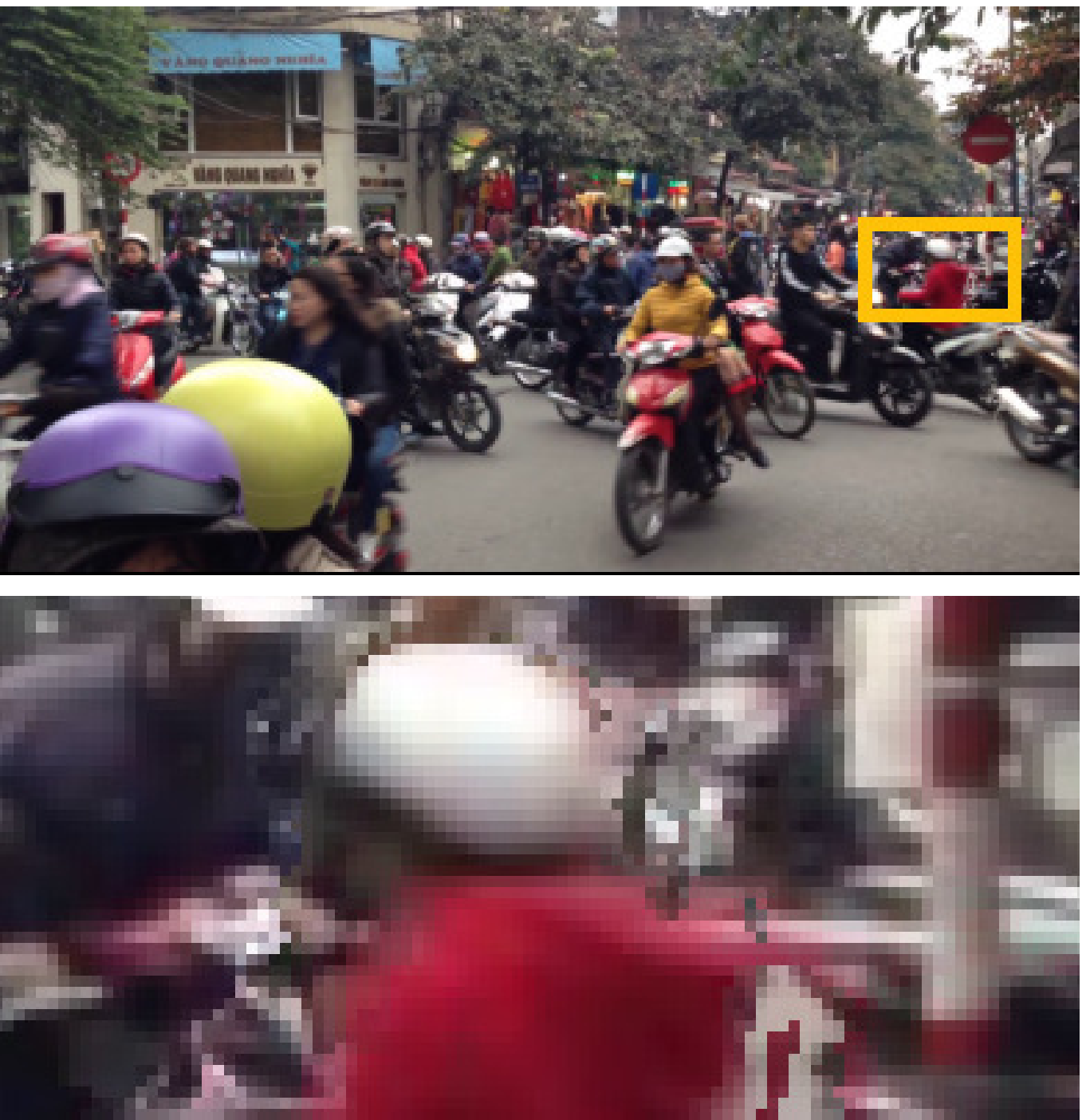}
        &
            \includegraphics[width=\itemwidth]{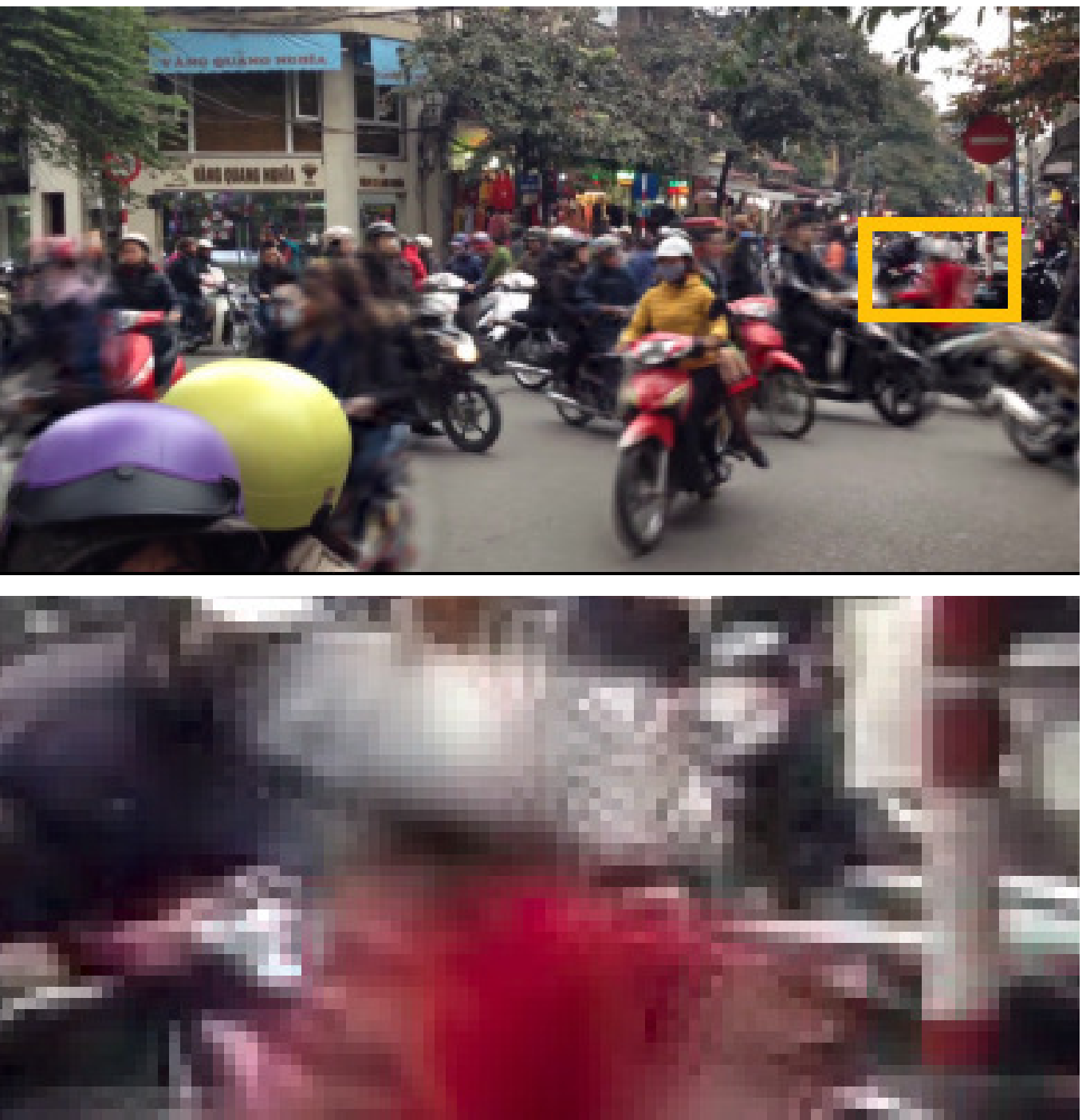}
        &
            \includegraphics[width=\itemwidth]{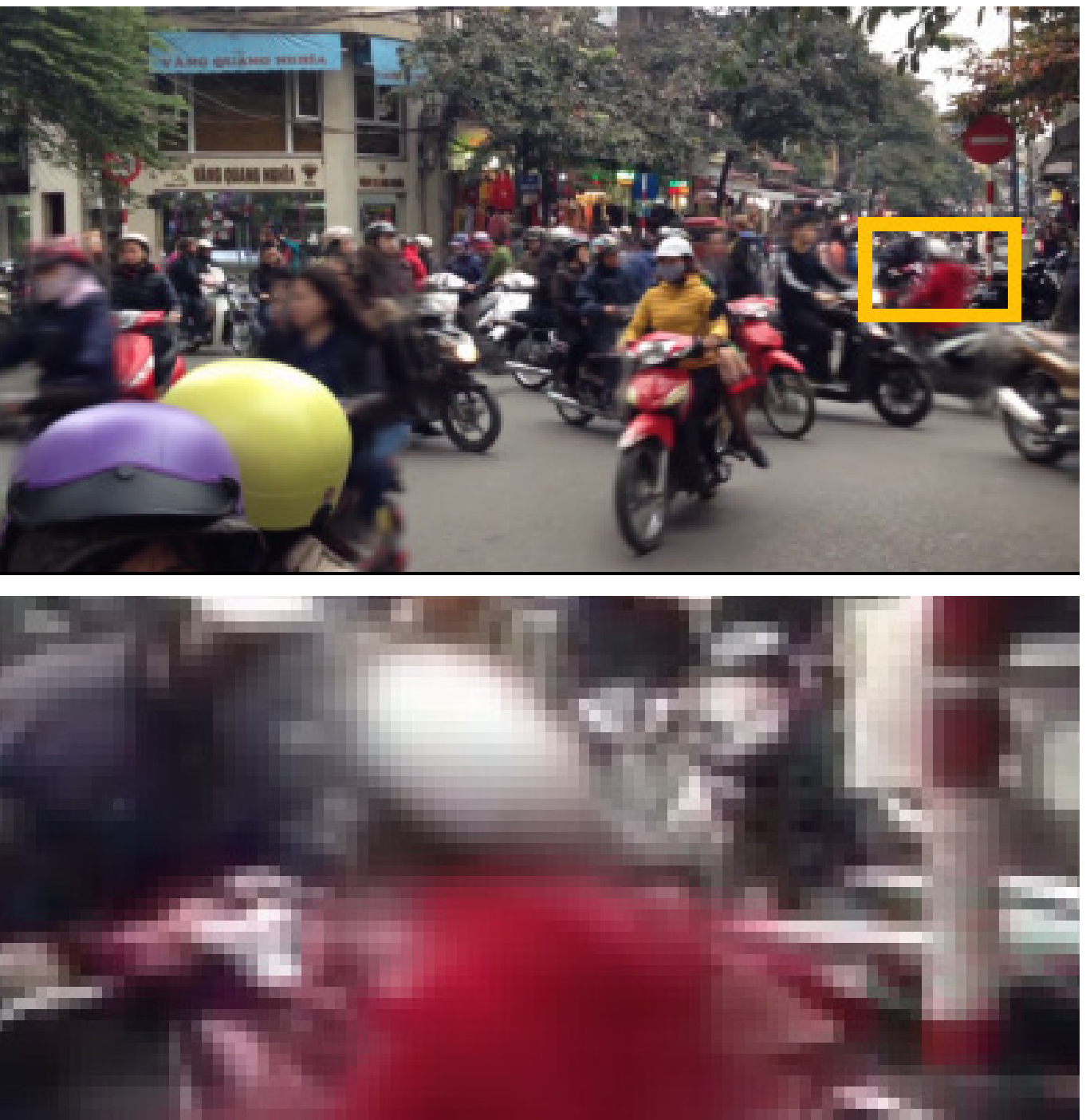}
        \vspace{-0.1cm} \\
            \footnotesize Overlayed input
        &
            \footnotesize Ours - $\mathcal{L}_1$
        &
            \footnotesize Ours - $\mathcal{L}_F$
        &
            \footnotesize MDP-Flow2
        &
            \footnotesize DeepFlow2
        &
            \footnotesize Meyer~\etal
        &
            \footnotesize AdaConv
        \\
    \end{tabularx}
    \begin{tabularx}{\textwidth}{c @{\hspace{0.05cm}} c @{\hspace{0.05cm}} c @{\hspace{0.05cm}} c @{\hspace{0.05cm}} c @{\hspace{0.05cm}} c @{\hspace{0.05cm}} c}
            \includegraphics[width=\itemwidth]{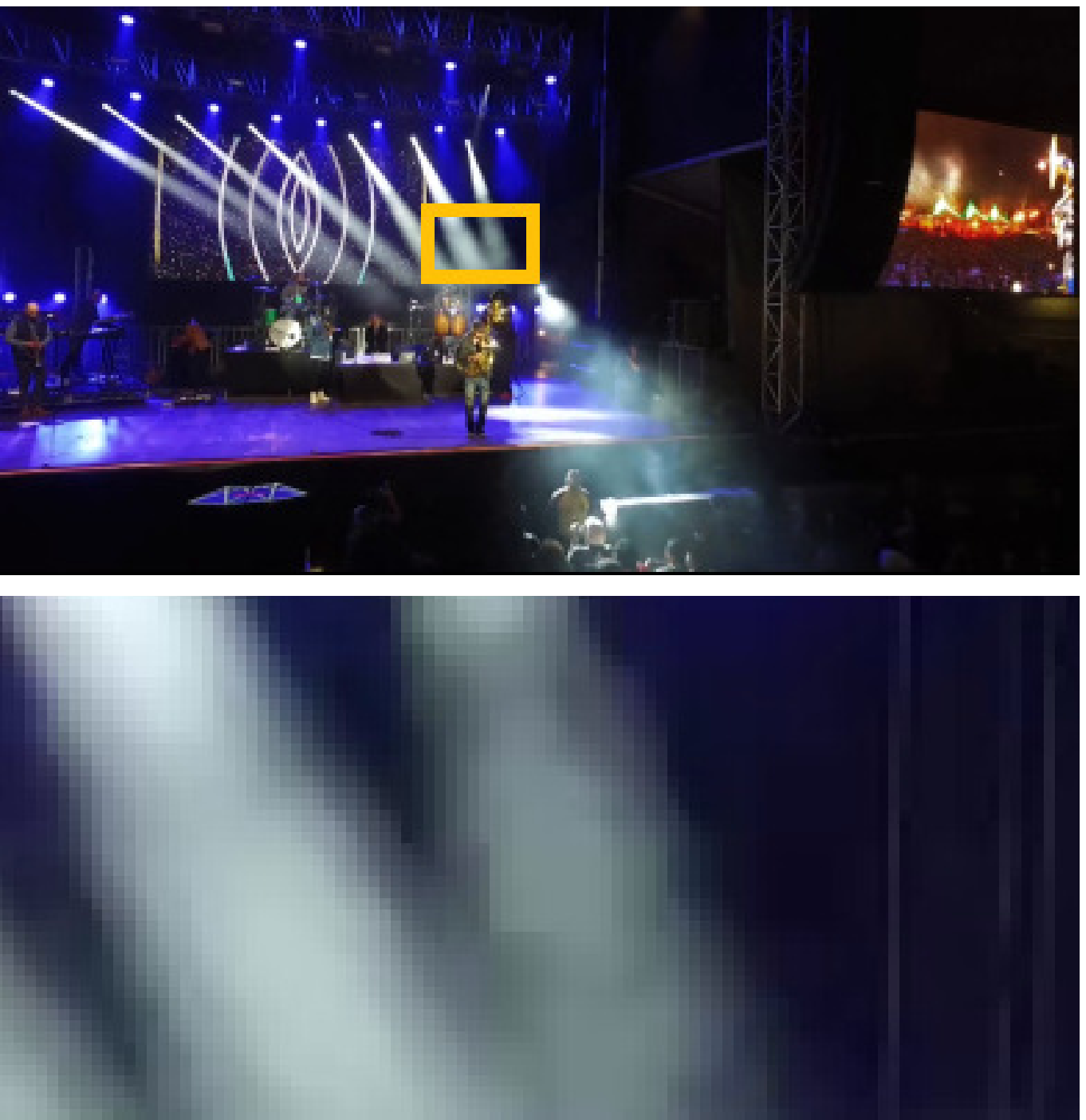}
        &
            \includegraphics[width=\itemwidth]{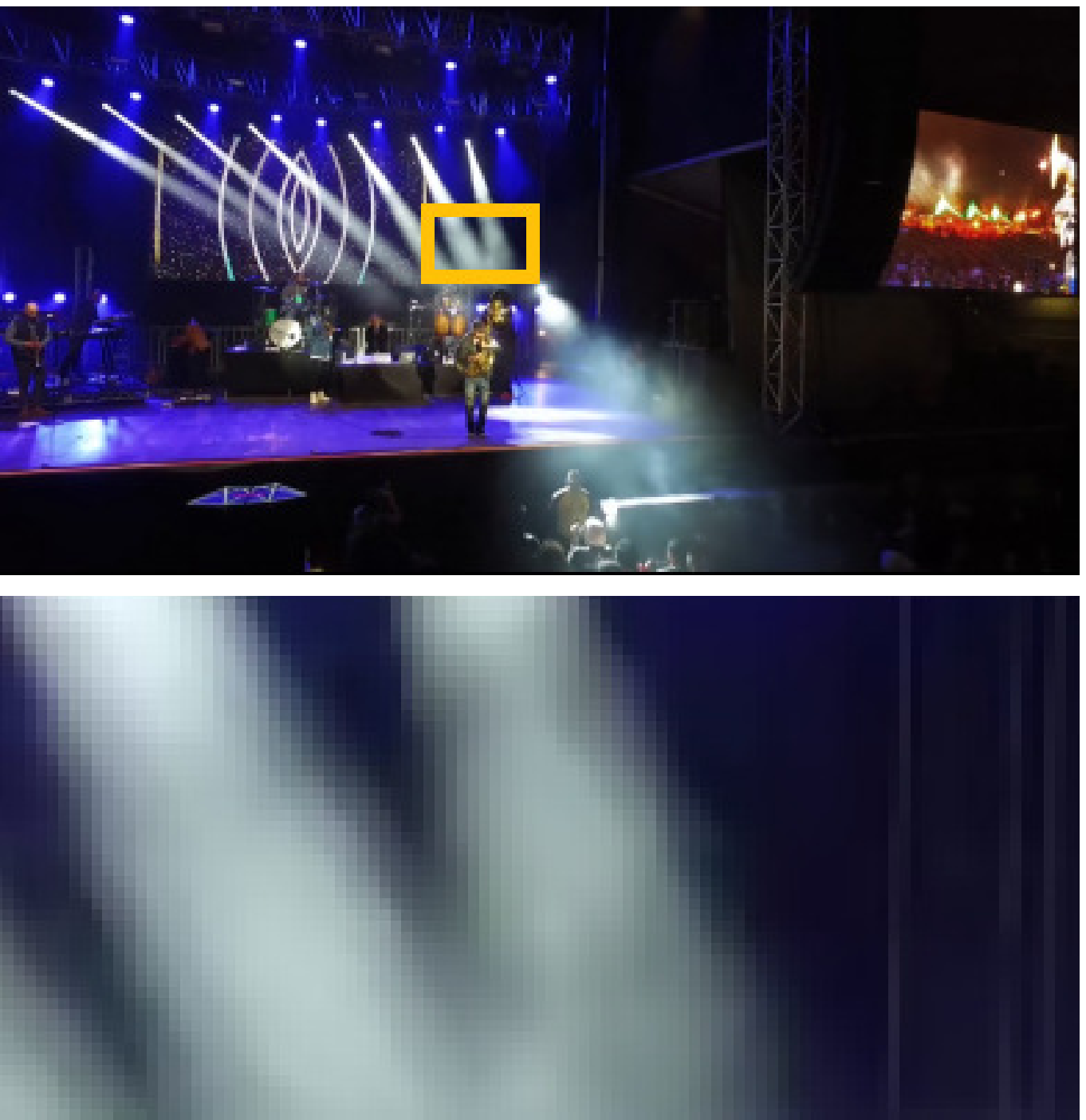}
        &
            \includegraphics[width=\itemwidth]{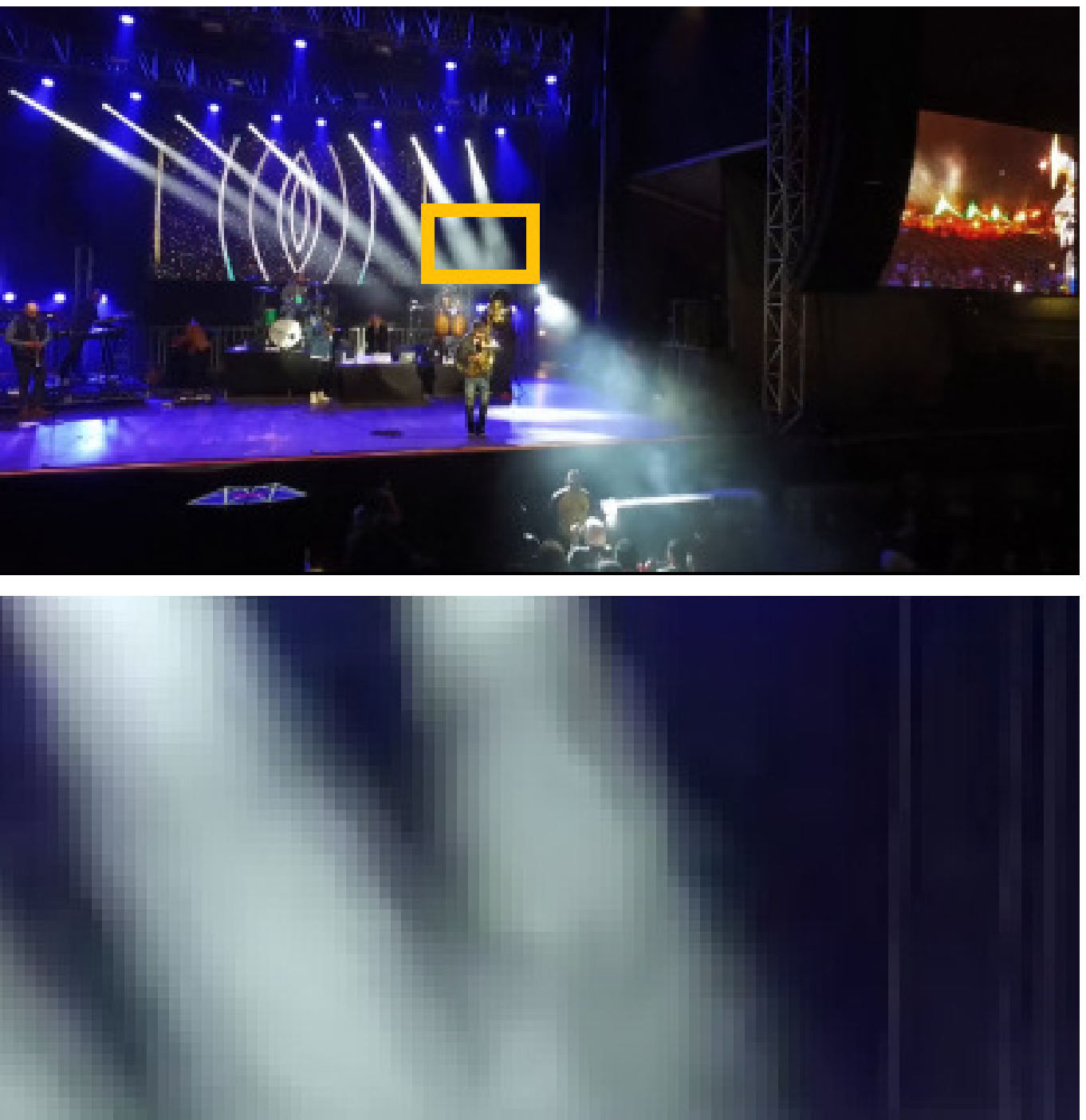}
        &
            \includegraphics[width=\itemwidth]{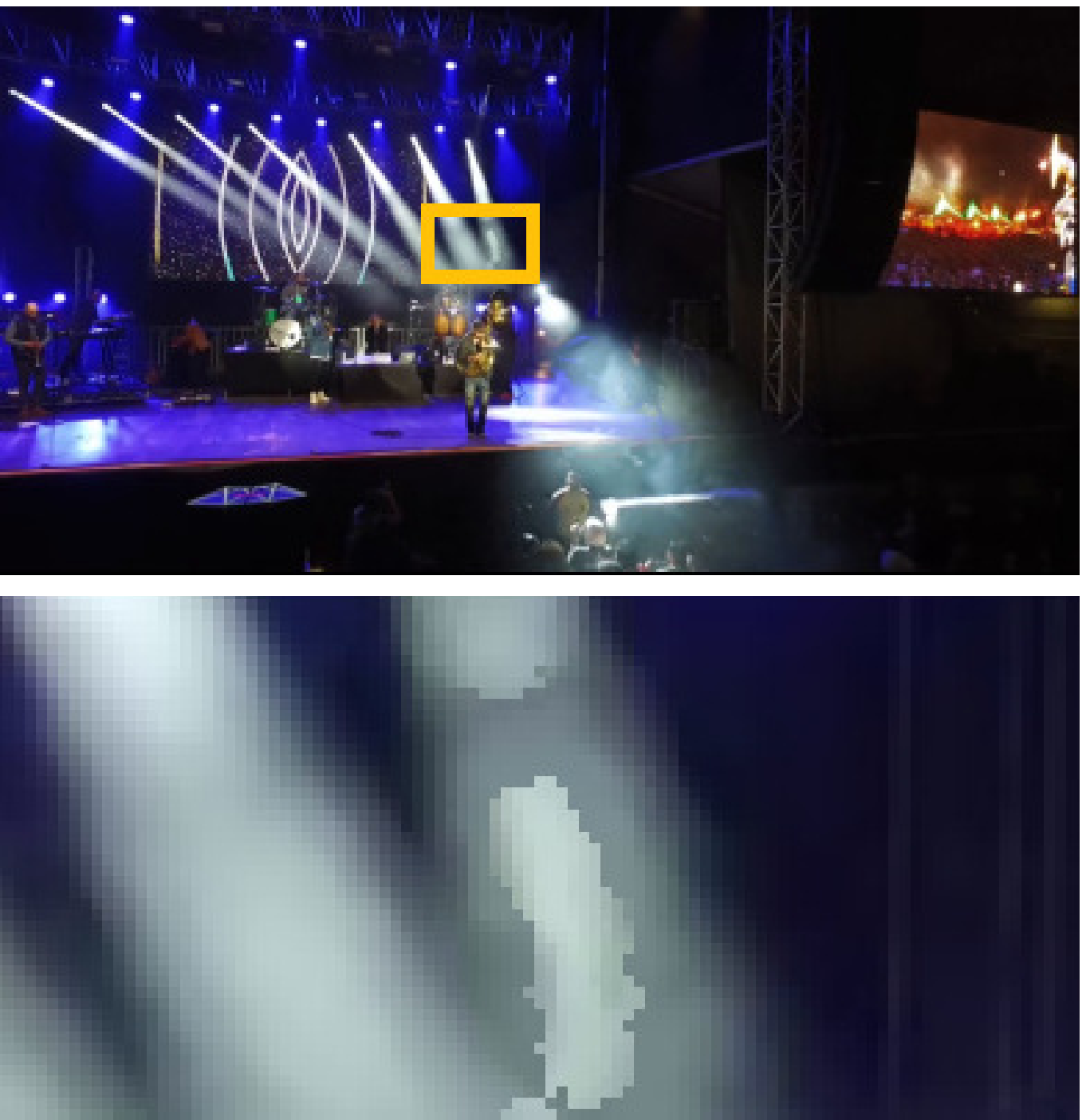}
        &
            \includegraphics[width=\itemwidth]{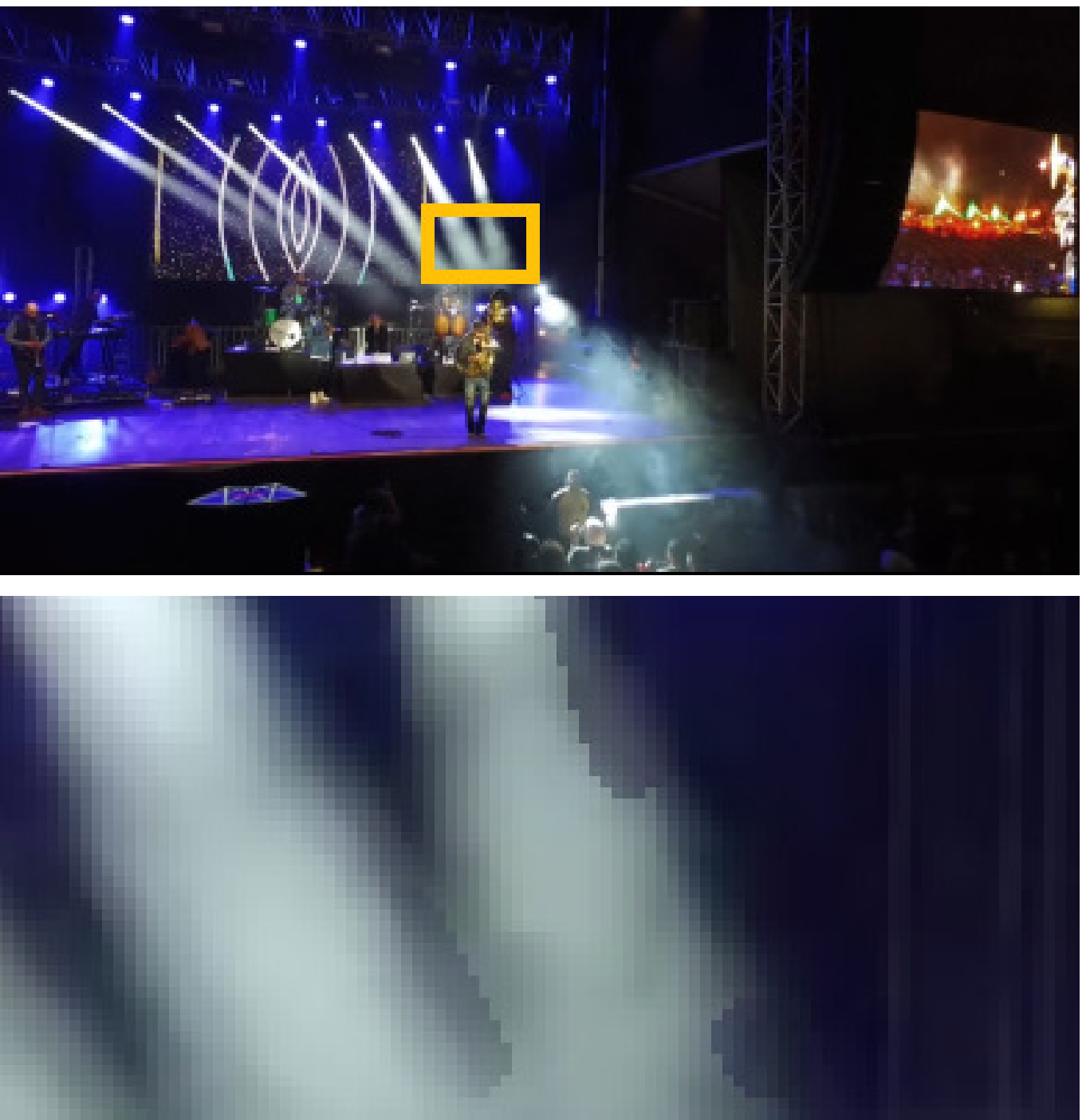}
        &
            \includegraphics[width=\itemwidth]{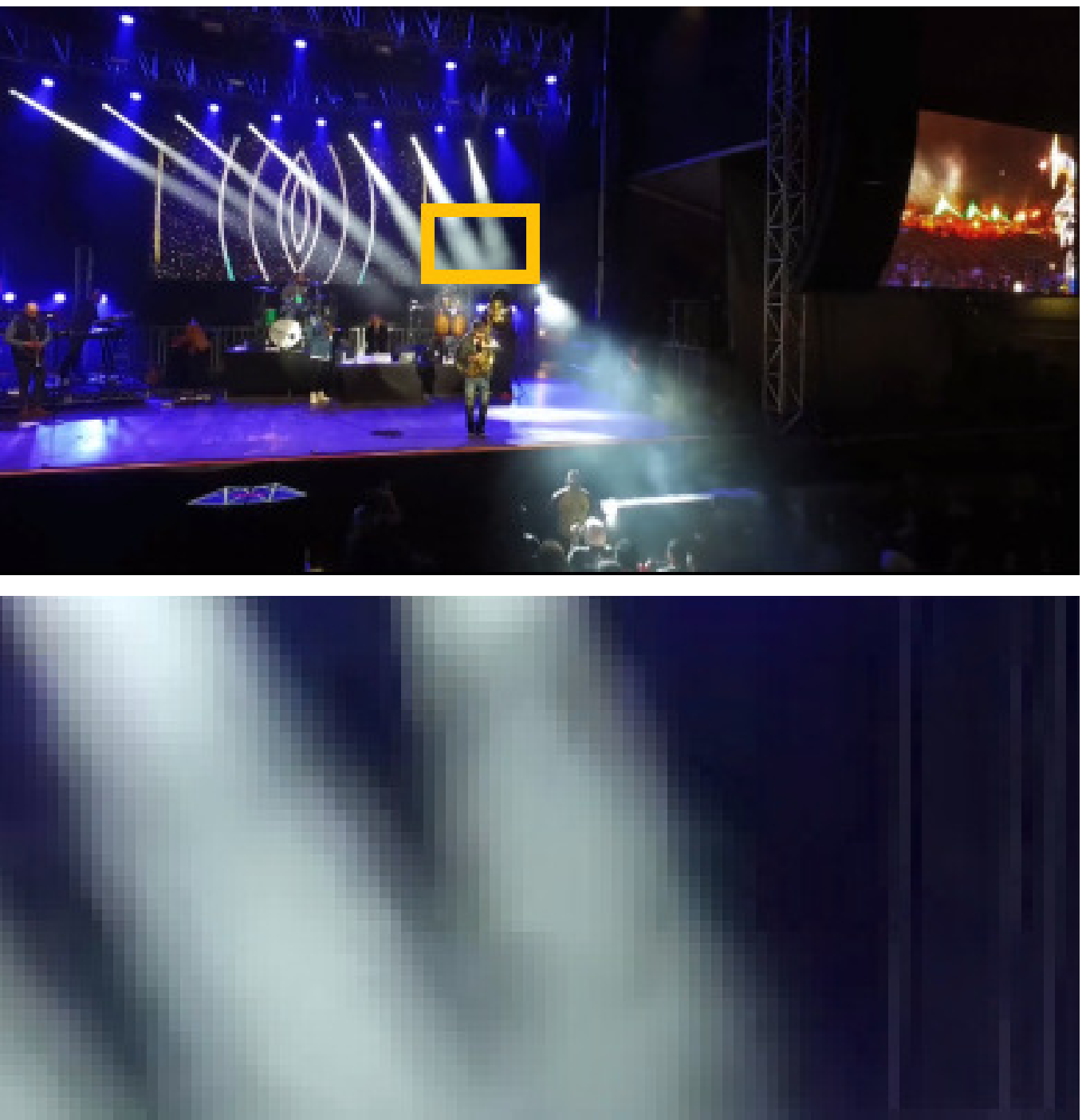}
        &
            \includegraphics[width=\itemwidth]{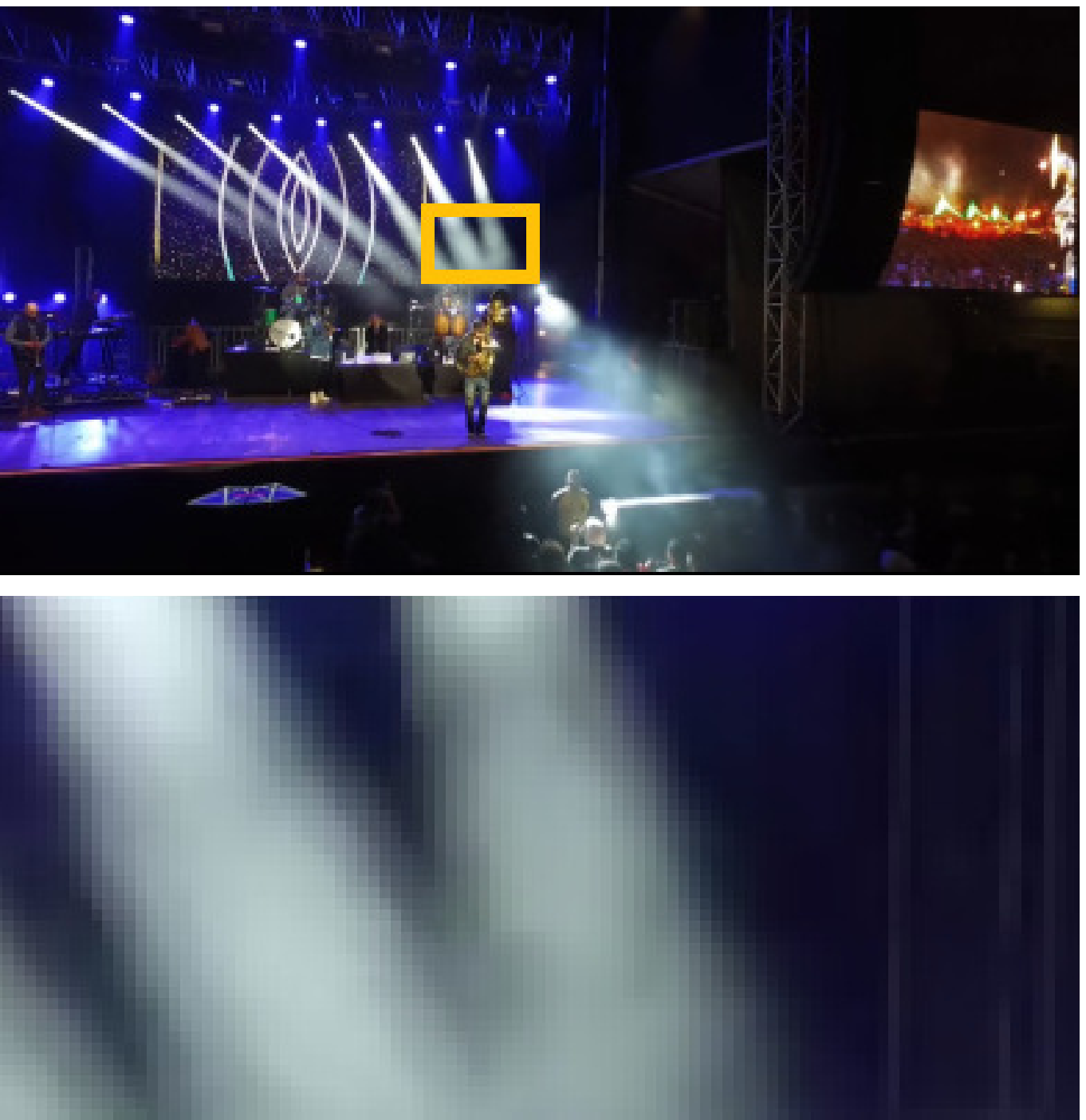}
        \vspace{-0.1cm} \\
    \end{tabularx}
    \begin{tabularx}{\textwidth}{c @{\hspace{0.05cm}} c @{\hspace{0.05cm}} c @{\hspace{0.05cm}} c @{\hspace{0.05cm}} c @{\hspace{0.05cm}} c @{\hspace{0.05cm}} c}
            \includegraphics[width=\itemwidth]{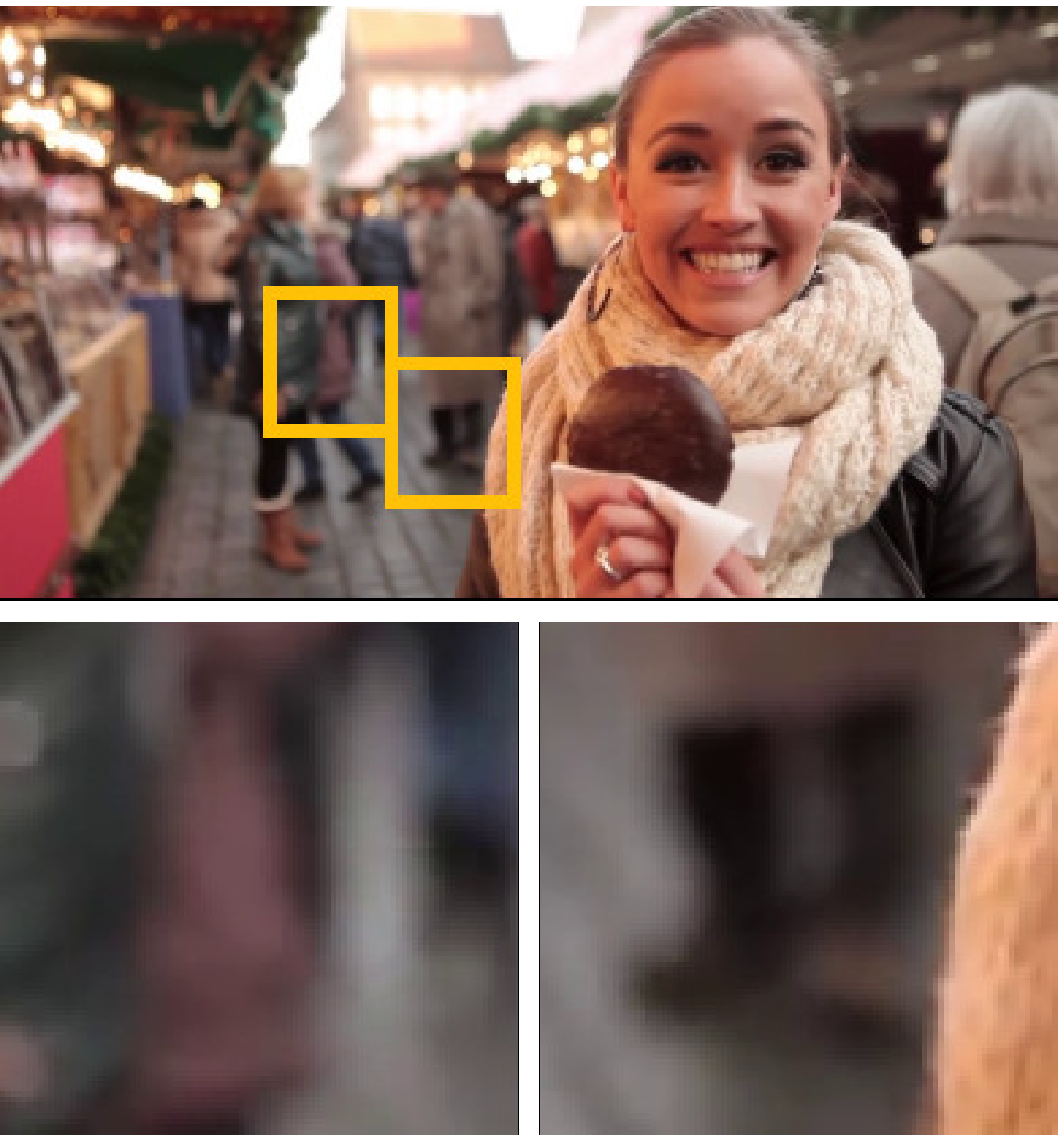}
        &
            \includegraphics[width=\itemwidth]{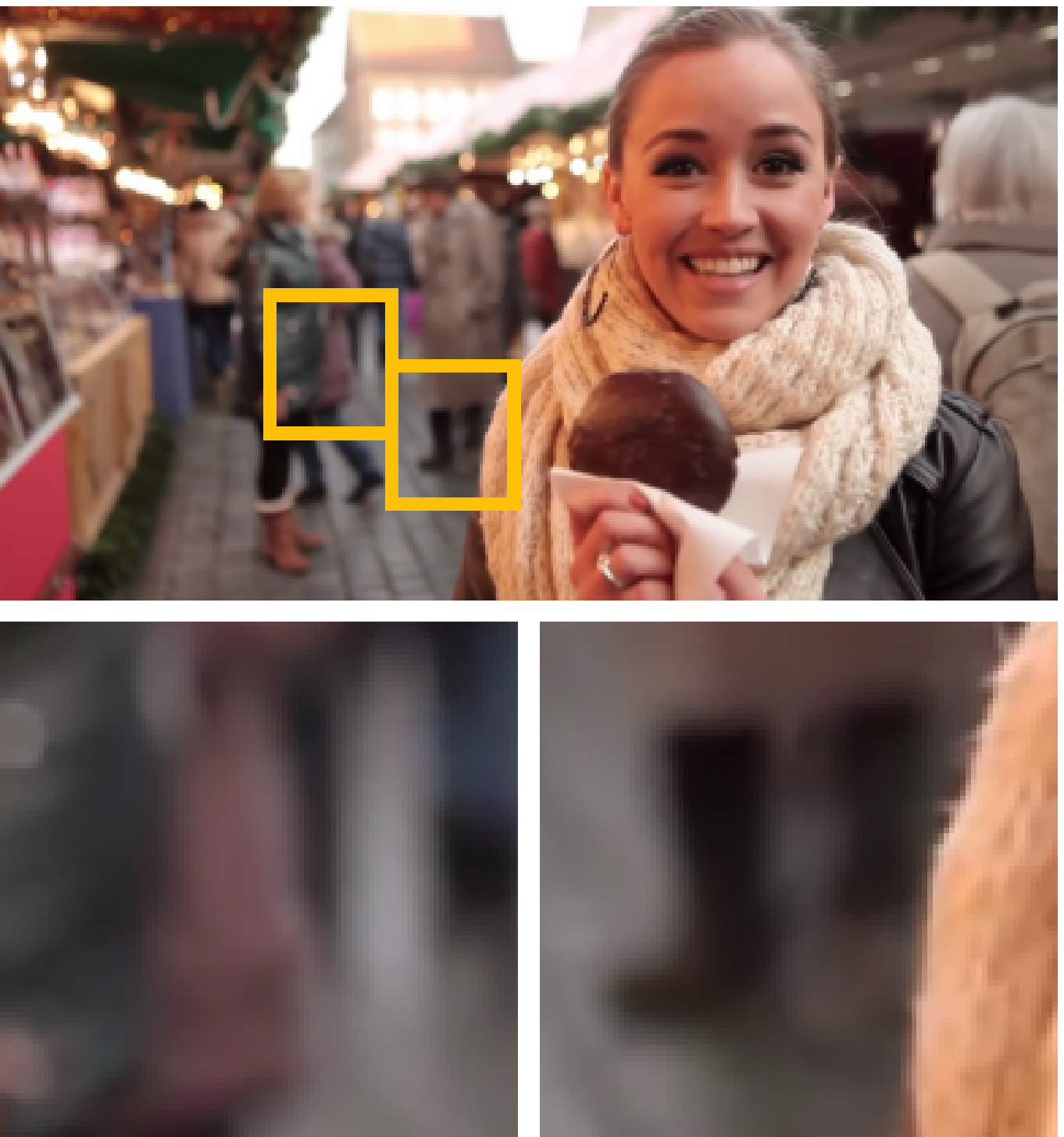}
        &
            \includegraphics[width=\itemwidth]{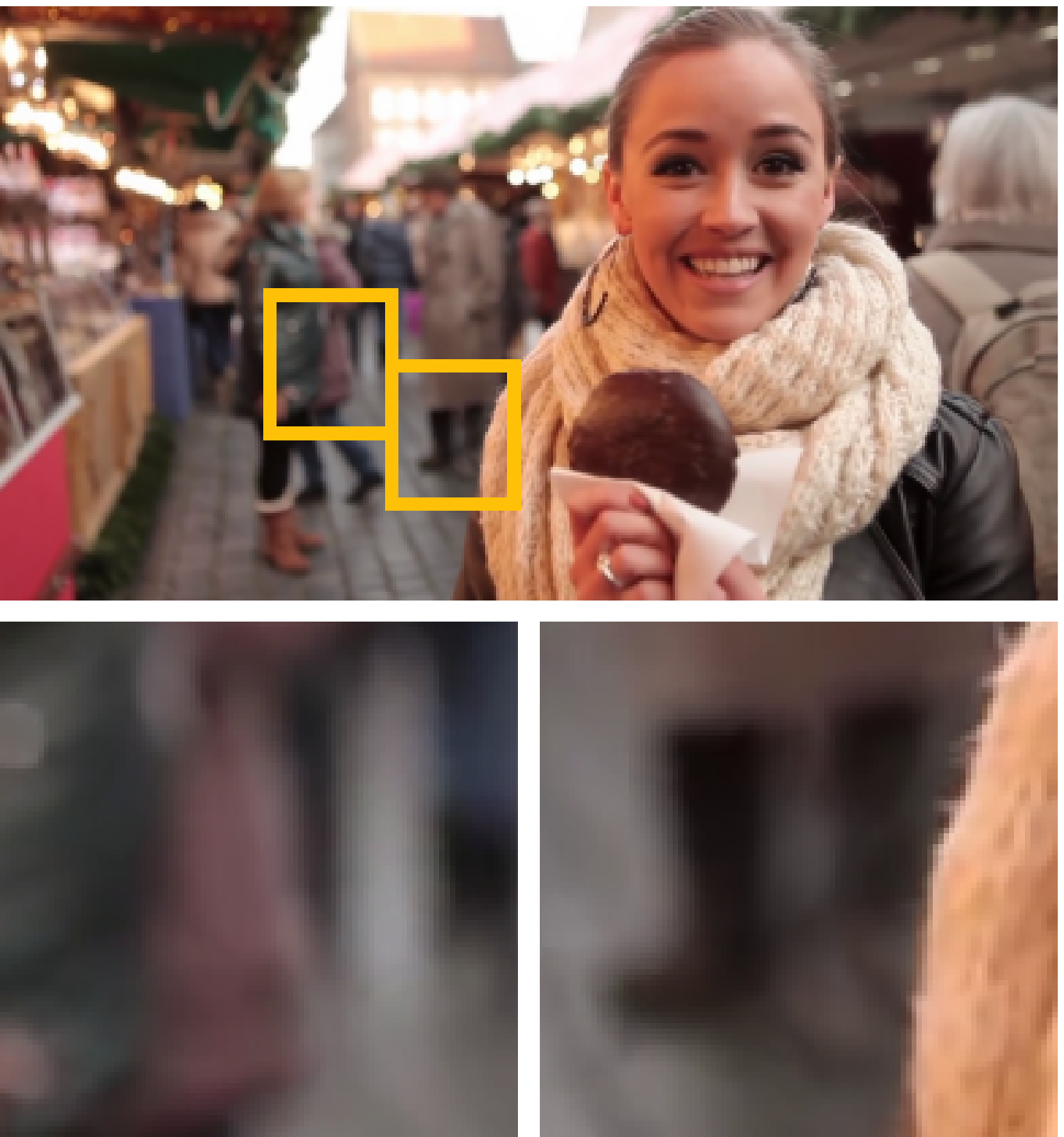}
        &
            \includegraphics[width=\itemwidth]{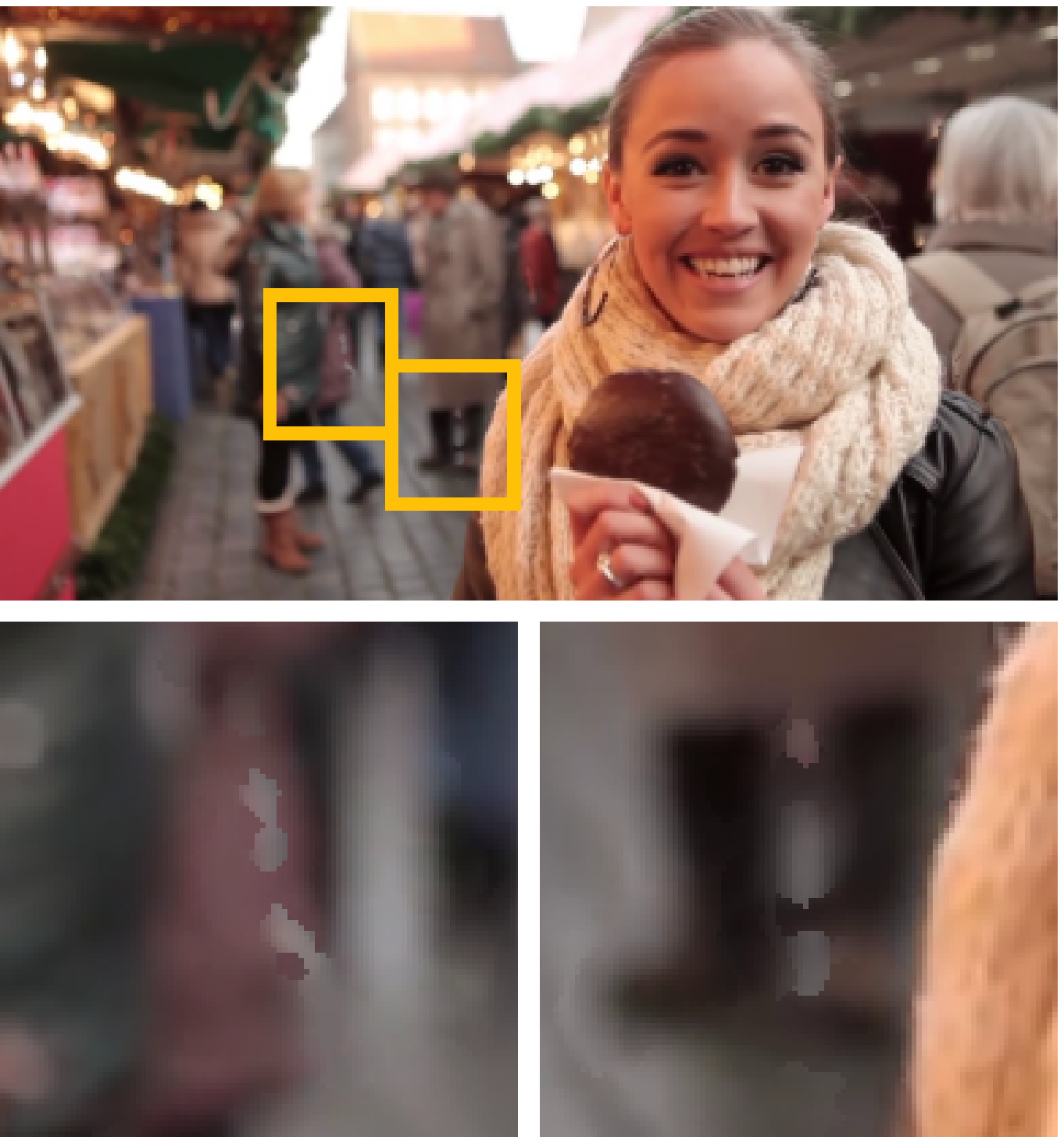}
        &
            \includegraphics[width=\itemwidth]{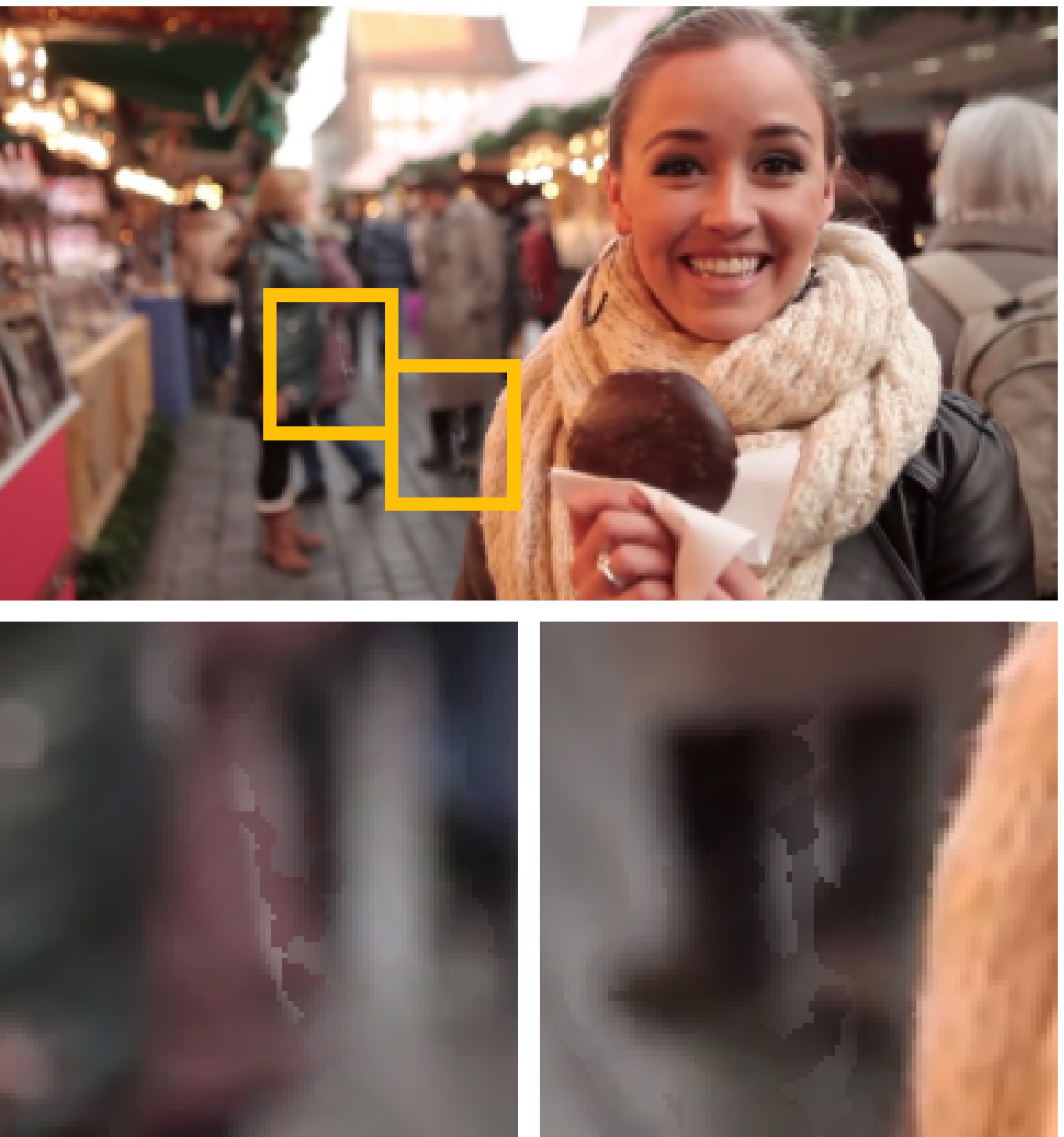}
        &
            \includegraphics[width=\itemwidth]{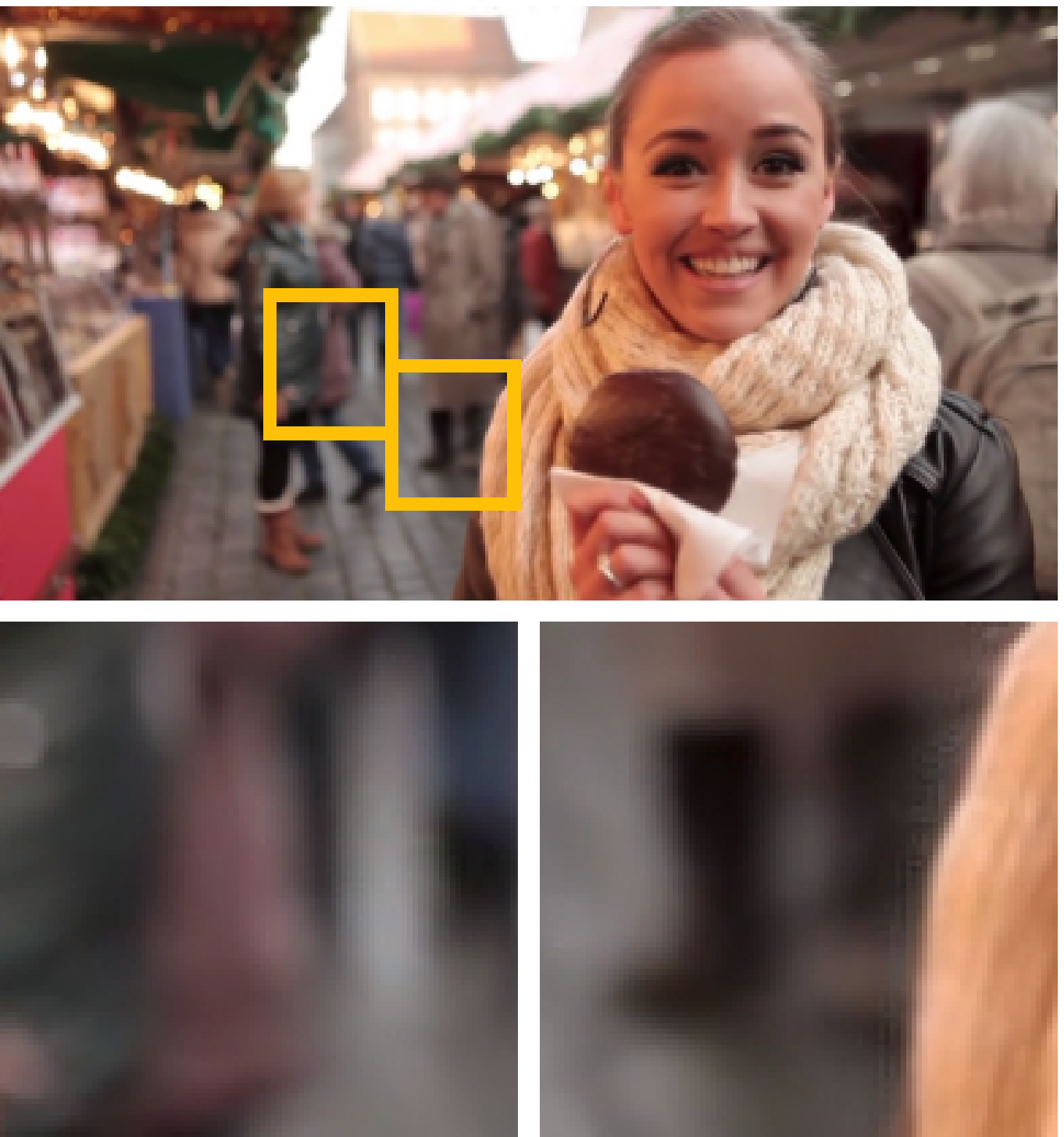}
        &
            \includegraphics[width=\itemwidth]{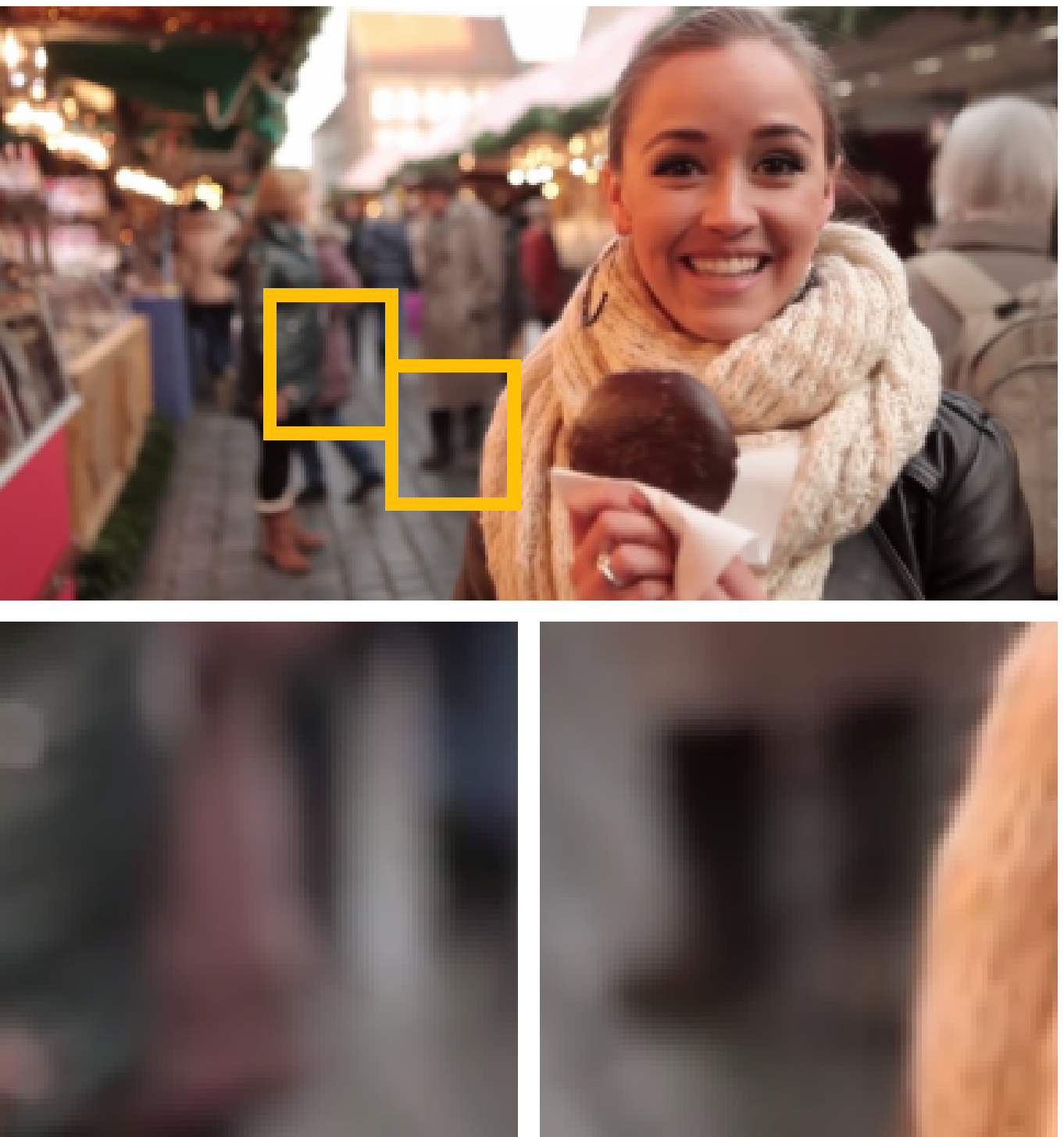}
        \vspace{-0.1cm} \\
            \footnotesize Input frame 1
        &
            \footnotesize Ours - $\mathcal{L}_1$
        &
            \footnotesize Ours - $\mathcal{L}_F$
        &
            \footnotesize MDP-Flow2
        &
            \footnotesize DeepFlow2
        &
            \footnotesize Meyer~\etal
        &
            \footnotesize AdaConv
        \\
    \end{tabularx}\vspace{-0.3cm}
    \caption{Visual comparison among frame interpolation methods.}\vspace{-0.5cm}
    \label{fig:examples}
\end{figure*}

\subsection{Visual comparison}

We examine how our separable convolution approach handles challenging cases of video frame interpolation.

The top row in Figure~\ref{fig:examples} shows an example where the delicate butterfly leg makes it difficult to estimate optical flow accurately, causing artifacts in the flow-based results. Since the leg motion is also large, the phase-based approach cannot handle it well either and produces ghosting artifacts. The result from AdaConv appears blurry. Both our results are sharp and free from ghosting artifacts.

The second row shows an example of a busy street. As people are moving in opposing directions, there is significant occlusion. Both our methods handle occlusion better than the others. We attribute this to the convolution approach and the use of 1D kernels with fewer parameters.

In the third row, we show an example of a stage where the rightmost spotlight is being turned on. This violates the brightness constancy assumption of optical flow methods, leading to visible artifacts in the frame interpolation results. The last row shows an example with shallow depth of field, which is common in professional videos. The blurry background makes flow estimation difficult and compromises the flow-based frame interpolation results. For these examples, the other methods, including ours, work well.

\begin{figure}\centering
    \setlength{\tabcolsep}{0.0cm}
    
    \includegraphics[width=\columnwidth]{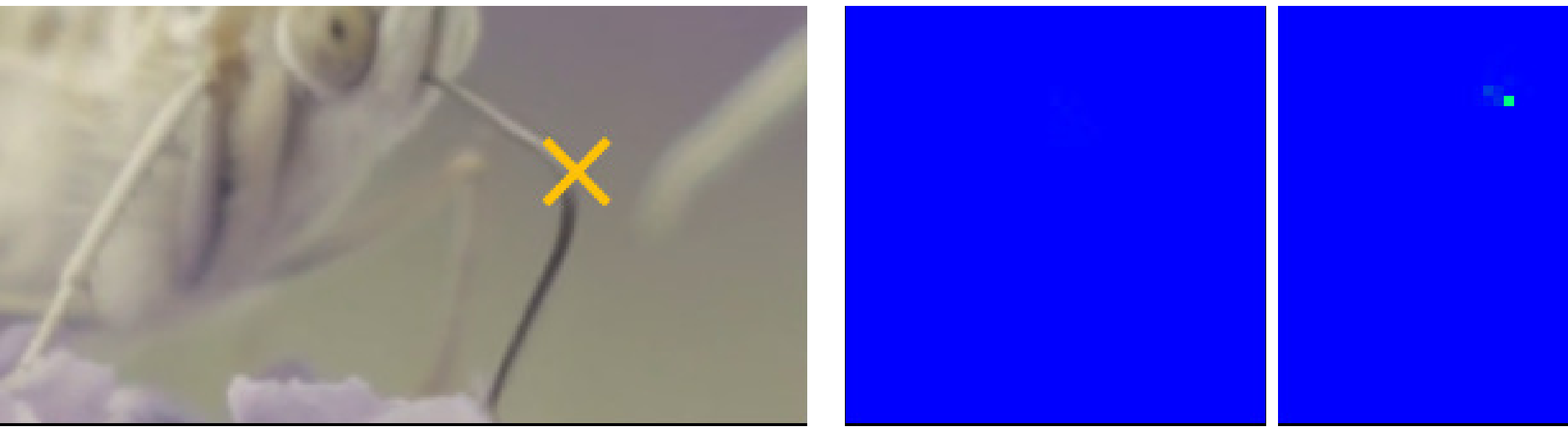}
    \includegraphics[width=\columnwidth]{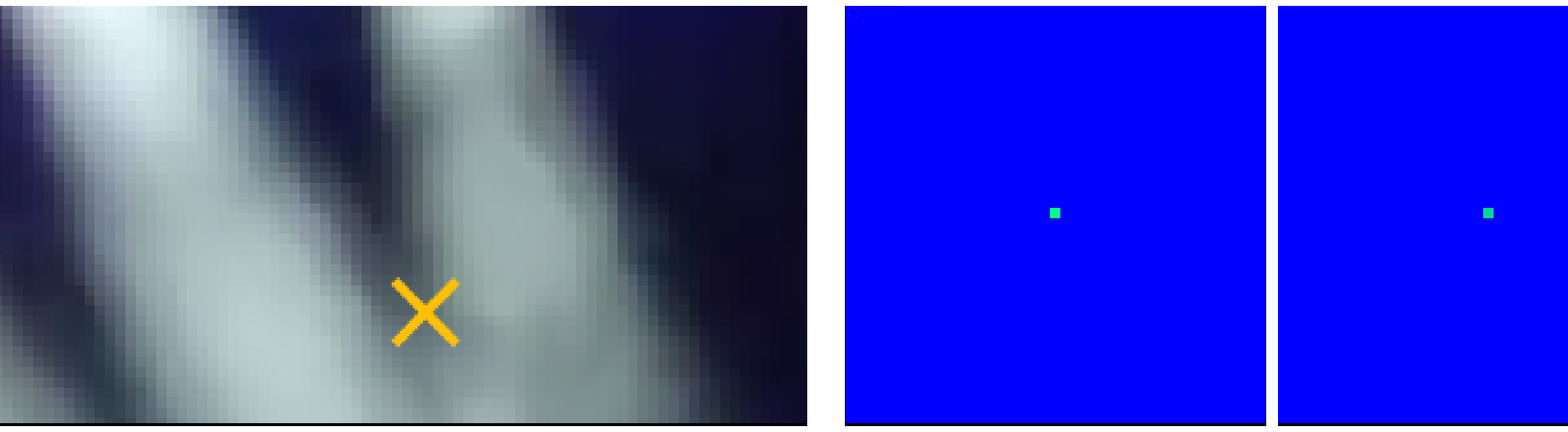}
    \includegraphics[width=\columnwidth]{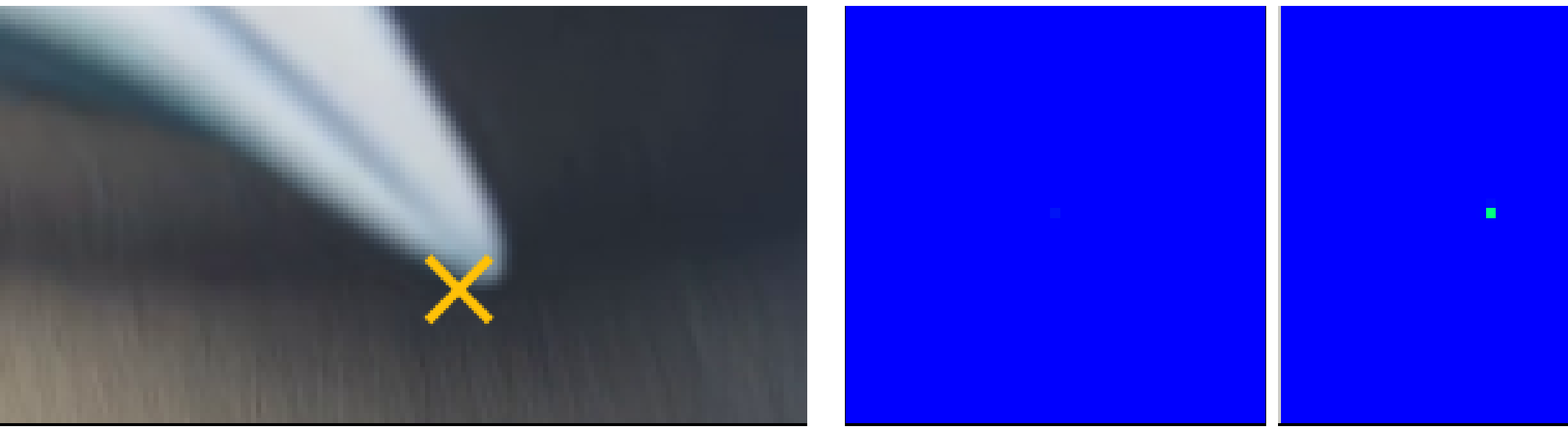}
    \begin{tabularx}{\columnwidth}{p{2.6cm} @{\hspace{0.05cm}} p{2.75cm} @{\hspace{0.05cm}} p{2.75cm}}
            \centering \footnotesize Synthesized frame
        &
            \centering \footnotesize AdaConv
        &
            \centering \footnotesize Ours - $\mathcal{L}_F$
        \\
    \end{tabularx}\vspace{-0.6cm}
    \caption{Comparison of the estimated kernels.}\vspace{-0.6cm}
	\label{fig:kernel}
\end{figure}

\noindent\textbf{Kernels.} We examine how the kernels estimated by our $\mathcal{L}_F$ method compare to those from AdaConv. We show some representative kernels in Figure~\ref{fig:kernel}. Note that we convolve each pair of 1D kernels from our method to produce its equivalent 2D kernel for comparison. As our kernels are larger than those from AdaConv, we cropped the boundary values off for better visualization as they are all zeros.

In the butterfly example, we show the kernels for a pixel on the leg. AdaConv only takes color from the corresponding pixel in the second input frame. While our method takes color mainly from the same pixel in the second input, it also takes color from the corresponding pixel in the first input frame. Since the color of that pixel remains the same in the two input frames, both methods produce proper results. Notice how both methods capture the motion encoded as the offset of the non-zero kernel values to the kernel center.

The second example shows kernels for a pixel in the lit area where the brightness changes between two input frames. Both methods output the same kernels that, due to the lack of motion, only have non-zero values in the center. Therefore, the output color is estimated as the average color of the corresponding pixels in the input frames.

The last example shows a pixel in an occluded area due to the leaf moving up. This area is only visible in the second input frame and both methods produce kernels that correctly choose to only sample from the second frame. They thus produce good results and are able to handle occlusion appropriately, unlike methods that explicitly have to establish a correspondence between pixels of the input frames.

\begin{table*}\centering
    \setlength{\tabcolsep}{0.0cm}
    
    \newcommand{\middSec}[1]{\scriptsize #1}
    \newcommand{\middAll}[1]{\scalebox{0.76}[1.0]{$\uline{ #1 }$}}
    \newcommand{\middALL}[1]{\scalebox{0.76}[1.0]{$\uline{\bm{ #1 }}$}}
    \newcommand{\middOth}[1]{\scalebox{0.76}[1.0]{$ #1 $}}
    \newcommand{\middOTH}[1]{\scalebox{0.76}[1.0]{$\bm{ #1 }$}}
    \scriptsize
    \begin{tabularx}{\textwidth}{@{\hspace{0.1cm}} X @{\hspace{0.0cm}} c @{\hspace{0.1cm}} c @{\hspace{0.1cm}} c @{\hspace{0.0cm}} l @{\hspace{0.225cm}} c @{\hspace{0.1cm}} c @{\hspace{0.1cm}} c @{\hspace{0.0cm}} l @{\hspace{0.225cm}} c @{\hspace{0.1cm}} c @{\hspace{0.1cm}} c @{\hspace{0.0cm}} l @{\hspace{0.225cm}} c @{\hspace{0.1cm}} c @{\hspace{0.1cm}} c @{\hspace{0.0cm}} l @{\hspace{0.225cm}} c @{\hspace{0.1cm}} c @{\hspace{0.1cm}} c @{\hspace{0.0cm}} l @{\hspace{0.225cm}} c @{\hspace{0.1cm}} c @{\hspace{0.1cm}} c @{\hspace{0.0cm}} l @{\hspace{0.225cm}} c @{\hspace{0.1cm}} c @{\hspace{0.1cm}} c @{\hspace{0.0cm}} l @{\hspace{0.225cm}} c @{\hspace{0.1cm}} c @{\hspace{0.1cm}} c @{\hspace{0.0cm}} l @{\hspace{0.225cm}} c @{\hspace{0.225cm}} c @{\hspace{0.1cm}} c @{\hspace{0.1cm}} c @{\hspace{0.0cm}} r @{\hspace{0.1cm}}}
        \toprule
            & \multicolumn{3}{c}{Mequon} && \multicolumn{3}{c}{Schefflera} && \multicolumn{3}{c}{Urban} && \multicolumn{3}{c}{Teddy} && \multicolumn{3}{c}{Backyard} && \multicolumn{3}{c}{Basketball} && \multicolumn{3}{c}{Dumptruck} && \multicolumn{3}{c}{Evergreen} &&& \multicolumn{3}{c}{\sc Average} &
        \\ \cmidrule{2-4} \cmidrule{6-8} \cmidrule{10-12} \cmidrule{14-16} \cmidrule{18-20} \cmidrule{22-24} \cmidrule{26-28} \cmidrule{30-32} \cmidrule{35-37}
            & \middSec{all} & \middSec{disc.} & \middSec{unt.} && \middSec{all} & \middSec{disc.} & \middSec{unt.} && \middSec{all} & \middSec{disc.} & \middSec{unt.} && \middSec{all} & \middSec{disc.} & \middSec{unt.} && \middSec{all} & \middSec{disc.} & \middSec{unt.} && \middSec{all} & \middSec{disc.} & \middSec{unt.} && \middSec{all} & \middSec{disc.} & \middSec{unt.} && \middSec{all} & \middSec{disc.} & \middSec{unt.} &&& \middSec{all} & \middSec{disc.} & \middSec{unt.} &
        \\ \midrule
            Ours - $\mathcal{L}_1$ & \middALL{2.52} & \middOTH{4.83} & \middOTH{1.11} && \middAll{3.56} & \middOTH{5.04} & \middOth{1.90} && \middAll{4.17} & \middOTH{4.15} & \middOth{2.86} && \middAll{5.41} & \middOTH{6.81} & \middOth{3.88} && \middAll{10.2} & \middOth{12.8} & \middOth{3.37} && \middAll{5.47} & \middOth{10.4} & \middOTH{2.21} && \middAll{6.88} & \middOth{15.6} & \middOth{1.72} && \middALL{6.63} & \middOTH{10.3} & \middOTH{1.62} &&\multicolumn{1}{|c}{}& \middALL{5.61} & \middOTH{8.74} & \middOth{2.33} &
        \\
            Ours - $\mathcal{L}_F$ & \middAll{2.60} & \middOth{5.00} & \middOth{1.19} && \middAll{3.87} & \middOth{5.50} & \middOth{2.07} && \middAll{4.38} & \middOth{4.29} & \middOth{2.73} && \middAll{5.78} & \middOth{7.16} & \middOth{3.94} && \middALL{10.1} & \middOTH{12.7} & \middOth{3.39} && \middAll{5.98} & \middOth{11.4} & \middOth{2.42} && \middALL{6.85} & \middOTH{15.5} & \middOth{1.78} && \middAll{6.90} & \middOth{10.8} & \middOth{1.65} &&\multicolumn{1}{|c}{}& \middAll{5.81} & \middOth{9.04} & \middOth{2.40} &
        \\
            MDP-Flow2 & \middAll{2.89} & \middOth{5.38} & \middOth{1.19} && \middALL{3.47} & \middOth{5.07} & \middOTH{1.26} && \middAll{3.66} & \middOth{6.10} & \middOth{2.48} && \middALL{5.20} & \middOth{7.48} & \middOTH{3.14} && \middAll{10.2} & \middOth{12.8} & \middOth{3.61} && \middAll{6.13} & \middOth{11.8} & \middOth{2.31} && \middAll{7.36} & \middOth{16.8} & \middOTH{1.49} && \middAll{7.75} & \middOth{12.1} & \middOth{1.69} &&\multicolumn{1}{|c}{}& \middAll{5.83} & \middOth{9.69} & \middOth{2.15} &
        \\
            DeepFlow2 & \middAll{2.99} & \middOth{5.65} & \middOth{1.22} && \middAll{3.88} & \middOth{5.79} & \middOth{1.48} && \middALL{3.62} & \middOth{6.03} & \middOTH{1.34} && \middAll{5.38} & \middOth{7.44} & \middOth{3.22} && \middAll{11.0} & \middOth{13.8} & \middOth{3.67} && \middAll{5.83} & \middOth{11.2} & \middOth{2.25} && \middAll{7.60} & \middOth{17.4} & \middOth{1.50} && \middAll{7.82} & \middOth{12.2} & \middOth{1.77} &&\multicolumn{1}{|c}{}& \middAll{6.02} & \middOth{9.94} & \middOTH{2.06} &
        \\
            AdaConv & \middAll{3.57} & \middOth{6.88} & \middOth{1.41} && \middAll{4.34} & \middOth{5.67} & \middOth{2.52} && \middAll{5.00} & \middOth{5.86} & \middOth{2.98} && \middAll{6.91} & \middOth{8.89} & \middOth{4.89} && \middAll{10.2} & \middOth{12.8} & \middOTH{3.21} && \middALL{5.33} & \middOTH{10.1} & \middOth{2.27} && \middAll{7.30} & \middOth{16.6} & \middOth{1.92} && \middAll{6.94} & \middOth{10.8} & \middOth{1.67} &&\multicolumn{1}{|c}{}& \middAll{6.20} & \middOth{9.70} & \middOth{2.61} &
        \\ \bottomrule
    \end{tabularx}\vspace{-0.2cm}
    \caption{Evaluation on the Middlebury benchmark. \emph{disc.}: regions with discontinuous motion. \emph{unt.}: textureless regions.}\vspace{-0.5cm}
    \label{tbl:middlebury}
\end{table*}

\begin{table}\centering
    \setlength{\tabcolsep}{0.0cm}

    \scriptsize
    \begin{tabularx}{\columnwidth}{@{\hspace{0.1cm}} X @{\hspace{0.0cm}} c @{\hspace{0.125cm}} c @{\hspace{0.125cm}} c @{\hspace{0.125cm}} c @{\hspace{0.0cm}} l @{\hspace{0.35cm}} c @{\hspace{0.125cm}} c @{\hspace{0.125cm}} c @{\hspace{0.125cm}} c @{\hspace{0.0cm}} r @{\hspace{0.1cm}}}
        \toprule
            & \multicolumn{4}{c}{Cross-validation} && \multicolumn{4}{c}{Video: See You Again} &
        \\
            & \multicolumn{4}{c}{\tiny $250,000$ samples at $150 \times 150$} && \multicolumn{4}{c}{\tiny $2,801$ samples at $960 \times 540$} &
        \\ \cmidrule{2-5} \cmidrule{7-10}
            & MAE & RMSE & PSNR & SSIM && MAE & RMSE & PSNR & SSIM &
        \\ \midrule
            Ours - $\mathcal{L}_1$ & $\bm{3.66}$ & $\bm{7.37}$ & $\bm{32.92}$ & $\bm{0.941}$ && $\bm{2.03}$ & $\bm{4.28}$ & $\bm{41.31}$ & $\bm{0.968}$ &
        \\
            Ours - $\mathcal{L}_F$ & $4.01$ & $7.84$ & $32.37$ & $0.934$ && $2.11$ & $4.40$ & $40.88$ & $0.965$ &
        \\
            MDP-Flow2 & $3.72$ & $7.40$ & $32.47$ & $0.940$ && $2.21$ & $5.01$ & $40.50$ & $0.961$ &
        \\
            DeepFlow2 & $3.89$ & $7.82$ & $32.16$ & $0.935$ && $2.09$ & $4.83$ & $40.52$ & $0.965$ &
        \\
            Meyer~\etal & $10.45$ & $17.16$ & $26.05$ & $0.705$ && $2.60$ & $5.36$ & $38.17$ & $0.944$ &
        \\
            AdaConv & $5.34$ & $10.14$ & $30.16$ & $0.885$ && $2.14$ & $4.44$ & $40.06$ & $0.967$ &
        \\ \bottomrule
    \end{tabularx}\vspace{-0.2cm}
    \caption{More extensive quantitative evaluation.}\vspace{-0.5cm}
    \label{tbl:quantitative}
\end{table}

\subsection{Quantitative evaluation}

We quantitatively evaluate our method on the interpolation set of the Middlebury optical flow benchmark~\cite{Baker_OTHER_2011}. Note that we did not fine-tune our models in any way. The results are shown in Table~\ref{tbl:middlebury}. Our $\mathcal{L}_1$ model and our $\mathcal{L}_F$ model perform particularly well in the regions with discontinuous motion. In terms of overall average, our $\mathcal{L}_1$ model achieves state-of-the-art results. Notice that our $\mathcal{L}_F$ model performs inferior to our $\mathcal{L}_1$ model in this quantitative evaluation due to its loss function that optimizes for perceptual quality.

For a more extensive quantitative evaluation, we performed a cross-validation and additionally assessed the interpolation capabilities of the different methods on a popular video. The results are shown in Table~\ref{tbl:quantitative}. For the former, we performed a 10-fold cross-validation on our training dataset for both of our methods and let the other methods directly interpolate the $250,000$ samples that each have a resolution of $150 \times 150$ pixels. Please note that this experiment is mainly to evaluate how our method can generalize. We did not adjust the parameters of the other methods or fine-tune them, which might limit their performance in this cross-validation experiment and we included them as baselines. For the latter, we obtained the video ``See You Again'' from Wiz Khalifa which currently is the most viewed video on YouTube. We processed this video at a size of $960 \times 540$ since this resolution is the largest that all methods and their reference implementations support. We withheld every other frame and used the remaining frames to interpolate the withheld ones. In this way, every method interpolated $2,801$ frames. Across these two additional experiments, our $\mathcal{L}_1$ model performs best regardless of the incorporated error metric. Like in the evaluation on the Middlebury benchmark, our $\mathcal{L}_F$ model quantitatively performs inferior to our $\mathcal{L}_1$ model due to the nature of the different loss functions that they were optimized with.

\subsection{User study}

\begin{figure*}\centering
    \hspace*{-0.2cm}\includegraphics[]{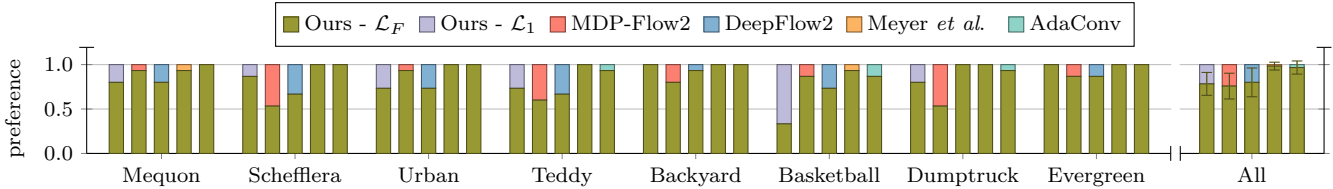}\vspace{-0.3cm}
	\caption{User study result. The error bars denote the standard deviation.}\vspace{-0.5cm}
	\label{fig:study}
\end{figure*}

\begin{figure}\centering
    \setlength{\tabcolsep}{0.0cm}
    \setlength{\itemwidth}{4.15cm}

    \begin{tabularx}{\textwidth}{c @{\hspace{0.05cm}} c}
            \includegraphics[width=\itemwidth]{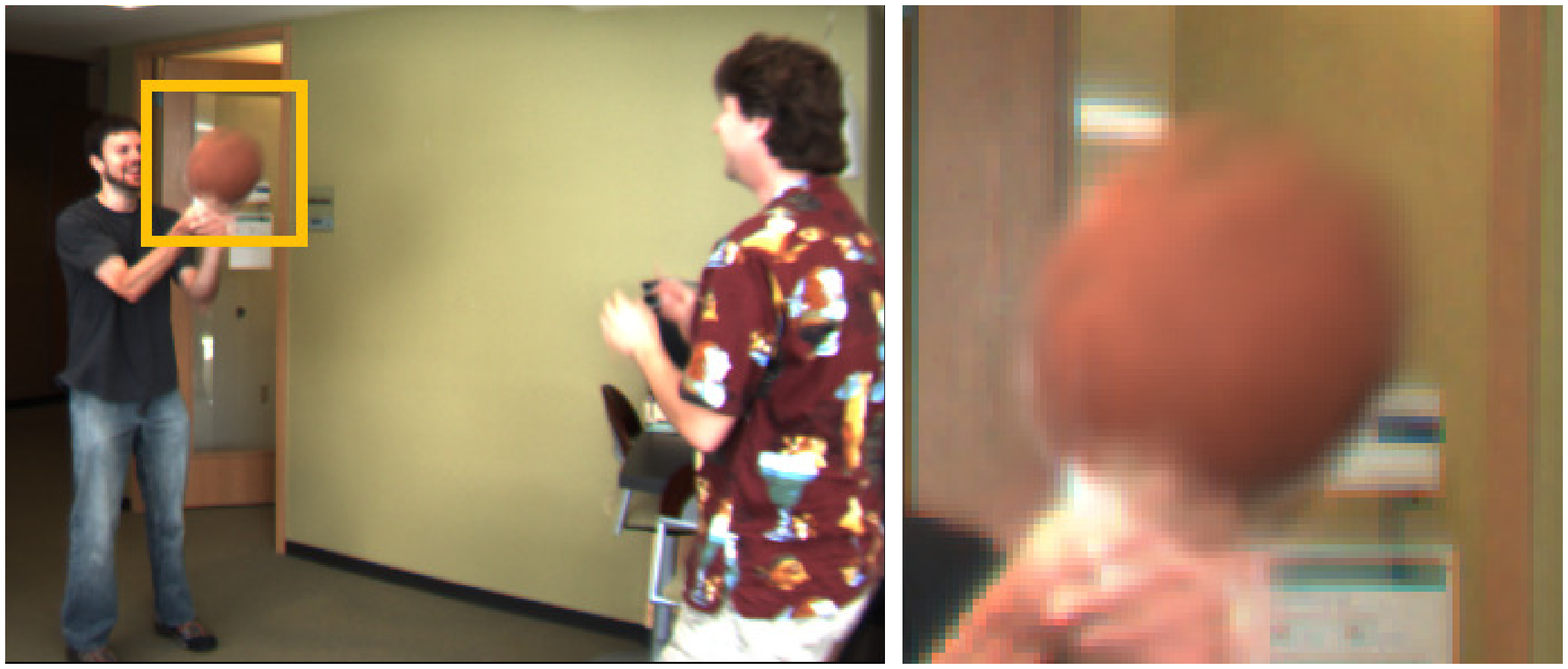}
        &
            \includegraphics[width=\itemwidth]{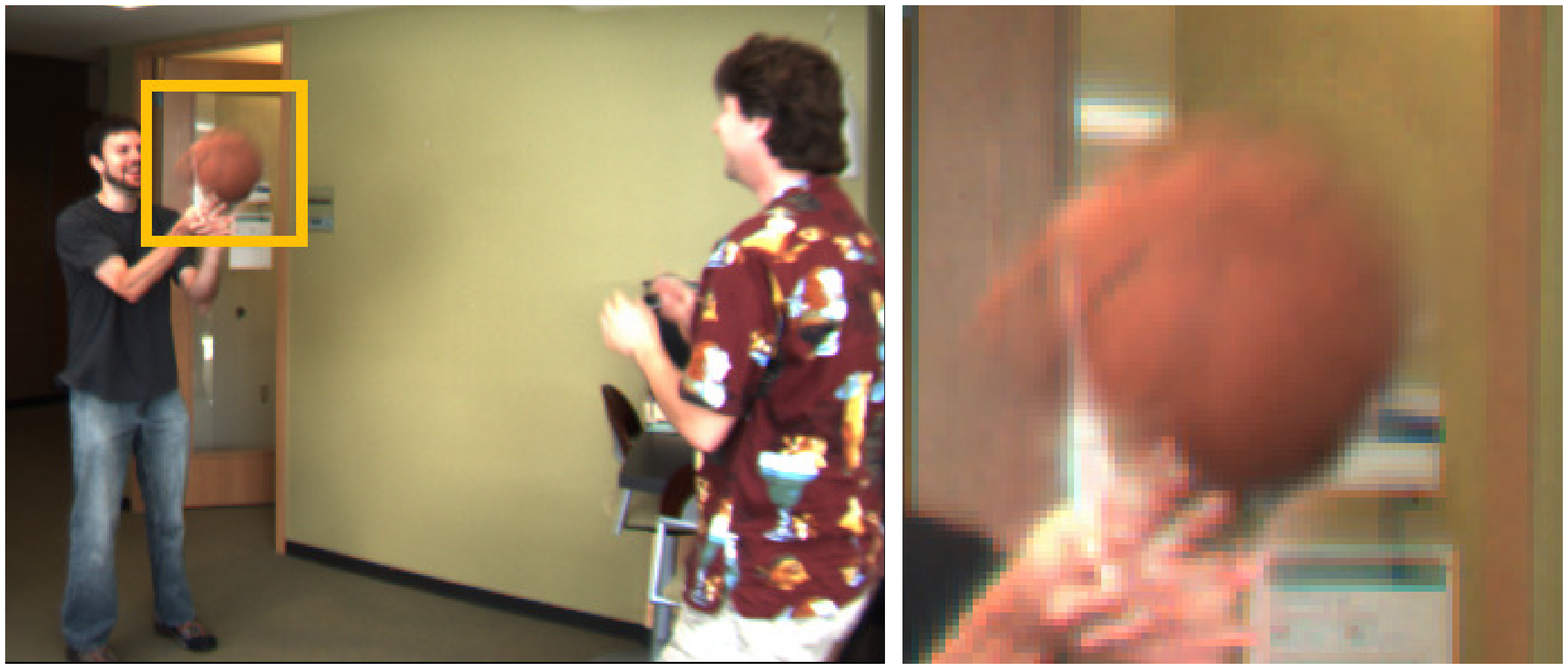}
        \vspace{-0.1cm} \\
            \footnotesize Ours - $\mathcal{L}_1$
        &
            \footnotesize Ours - $\mathcal{L}_F$
        \\
    \end{tabularx}\vspace{-0.3cm}
    \caption{Example where users prefer our $\mathcal{L}_1$ result.}\vspace{-0.3cm}
    \label{fig:basketball}
\end{figure}

We conducted a user study to further compare the visual quality of the frame interpolation results from our $\mathcal{L}_F$ method with our $\mathcal{L}_1$ method as well as the other four methods. We recruited 15 participants, who are graduate or undergraduate students in Computer Science and Statistics. This study used all 8 examples of the Middlebury testing set. On each example, our $\mathcal{L}_F$ result was compared to the other 5 results one by one. In this way, each participant compared 40 pairs of results. We developed a web-based system for the study. In each trial, the website only shows one result and supports participants to switch back and forth between two results using the arrow keys on the keyboard, allowing them to easily examine the difference between the results. The participants were asked to select the better result for each trial. The temporal order as well as the order in which the two results appear were randomized.

Figure~\ref{fig:study} shows the result of this study. For each hypothesis that users prefer the frames interpolated by our $\mathcal{L}_F$ method over those produced by one of the baselines, we get a p-value $< 0.01$ and can thus confirm them. Note that the participants preferred our $\mathcal{L}_1$ result over our $\mathcal{L}_F$ result on the Basketball example, shown in Figure~\ref{fig:basketball}. We attribute this to the introduced discontinuity to the basketball.

\subsection{Comparison with AdaConv} 

Our method builds upon AdaConv~\cite{Niklaus_CVPR_2017} by estimating 1D kernels instead of 2D kernels and developing a dedicated encoder-decoder neural network to estimate the kernels for all the pixels in a frame at once. This provides a few advantages. First, our method is over 20 times faster than AdaConv when interpolating a 1080p video. Second, as shown in the previous quantitative comparisons (Table~\ref{tbl:middlebury} and Table~\ref{tbl:quantitative}), our method produces numerically better results. Third, our methods, especially $\mathcal{L}_F$, often generates visually more appealing results than AdaConv as shown in Figure~\ref{fig:examples},~\ref{fig:gucan}, and in our study. We attribute these advantages to the separable convolution. First, it allows us to synthesize the full frame at once and to use perceptual loss that has recently been shown effective in producing visually pleasing results~\cite{Dosovitskiy_NIPS_2016, Johnson_ECCV_2016, Ledig_CORR_2016, Sajjadi_CORR_2016, Zhu_ECCV_2016}. Second, 1D kernels require significantly fewer parameters, which enforces a useful constraint to obtain good kernels. Third, our method is able to use a larger kernel than AdaConv and can thus handle larger motion. As shown in Figure~\ref{fig:direct}, AdaConv cannot capture the motion of the cars and generates blurry results.

\subsection{Discussion}

\begin{figure}\centering
    \setlength{\tabcolsep}{0.0cm}
    \setlength{\itemwidth}{2.75cm}

    \begin{tabularx}{\columnwidth}{c @{\hspace{0.05cm}} c @{\hspace{0.05cm}} c}
            \includegraphics[width=\itemwidth]{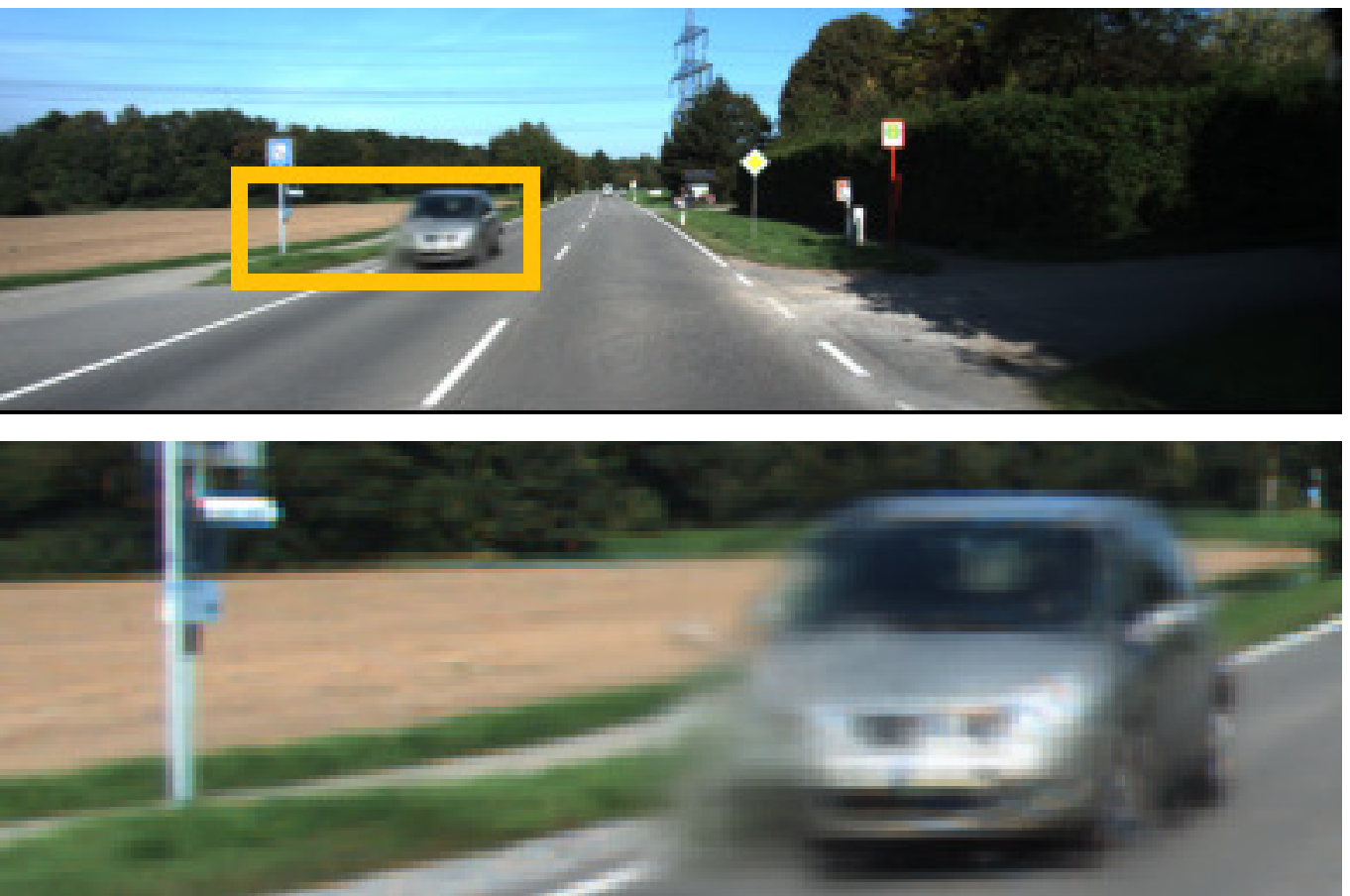}
        &
            \includegraphics[width=\itemwidth]{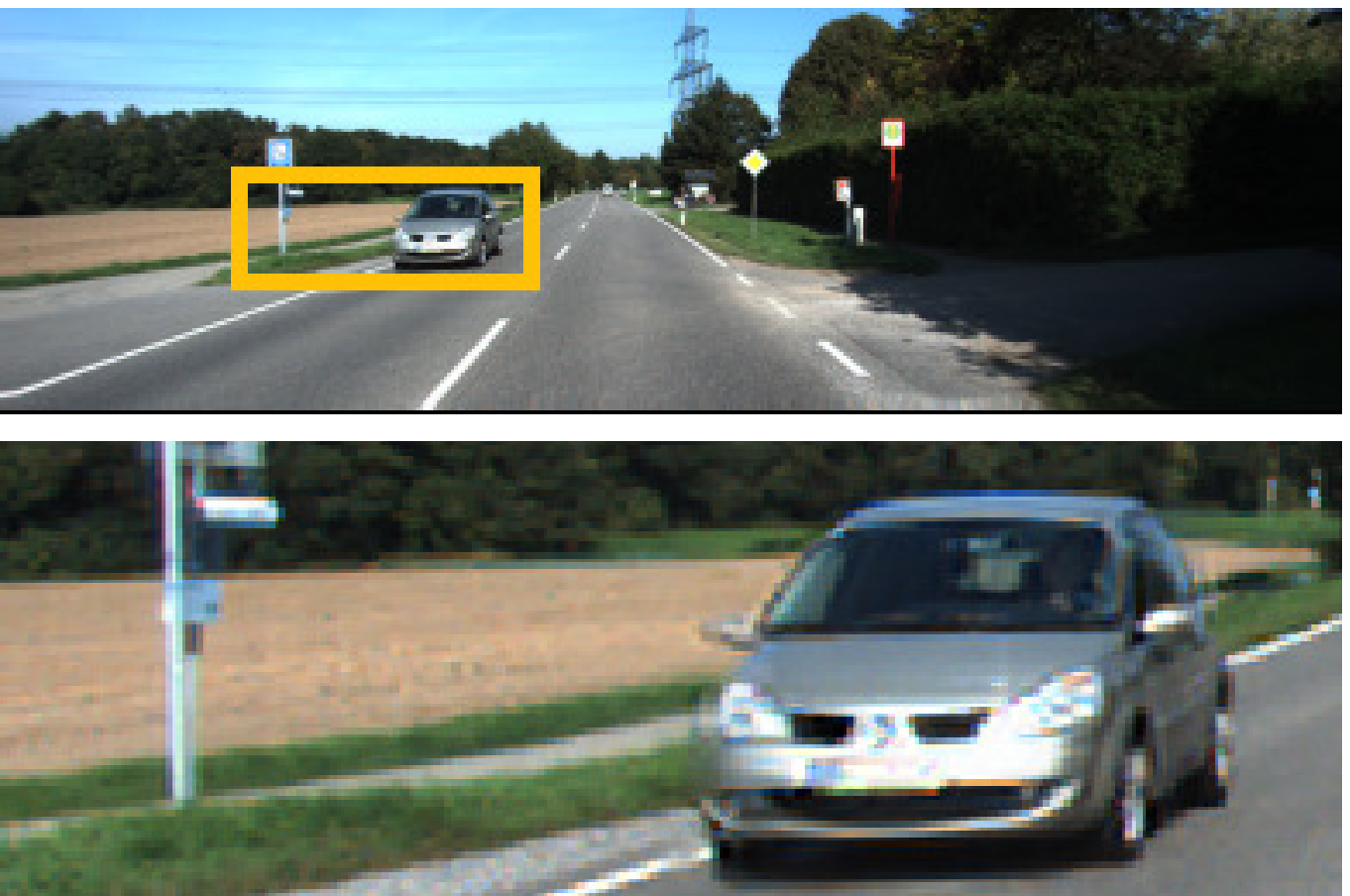}
        &
            \includegraphics[width=\itemwidth]{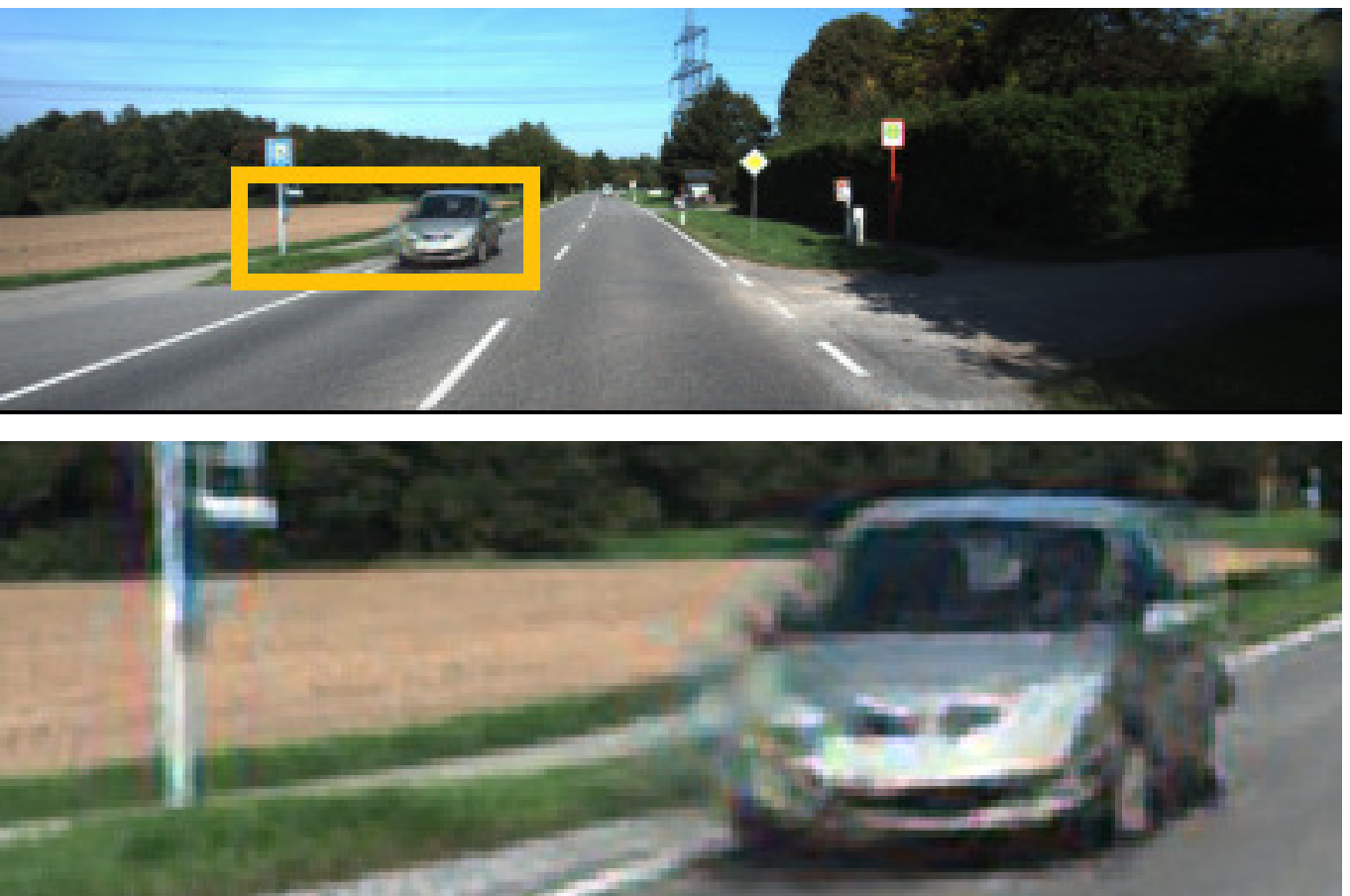}
        \vspace{-0.1cm} \\
    \end{tabularx}
    \begin{tabularx}{\columnwidth}{c @{\hspace{0.05cm}} c @{\hspace{0.05cm}} c}
            \includegraphics[width=\itemwidth]{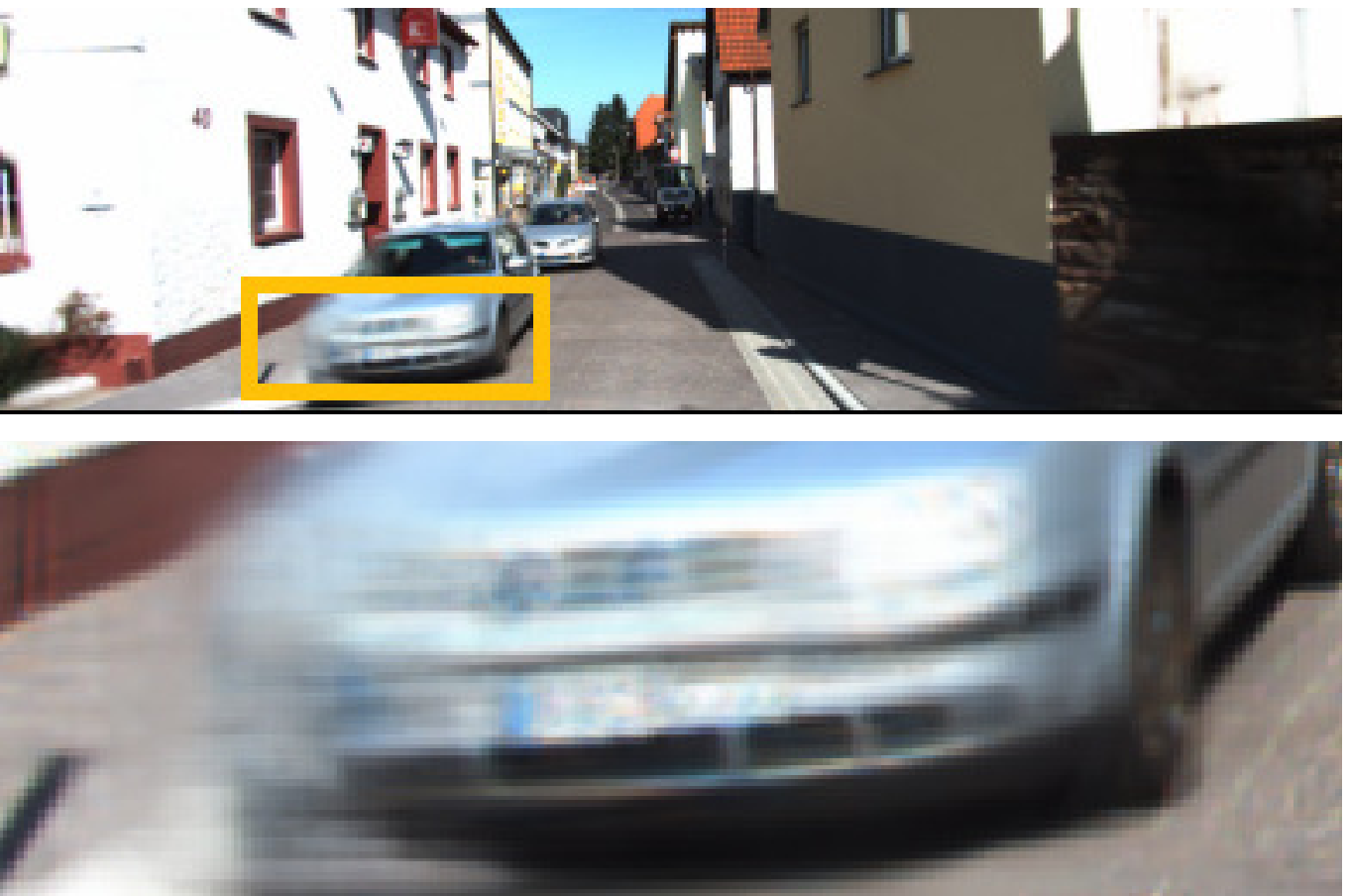}
        &
            \includegraphics[width=\itemwidth]{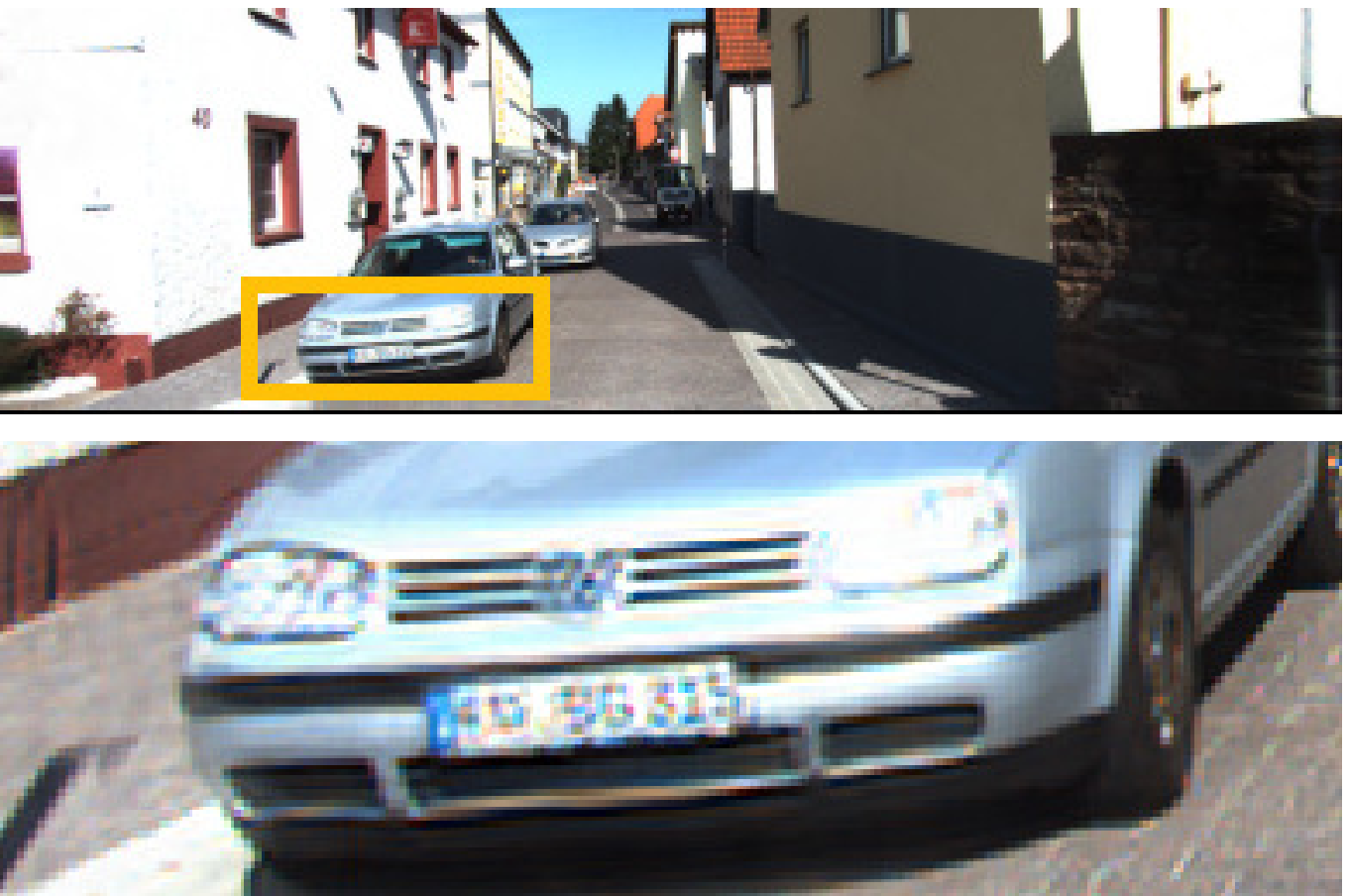}
        &
            \includegraphics[width=\itemwidth]{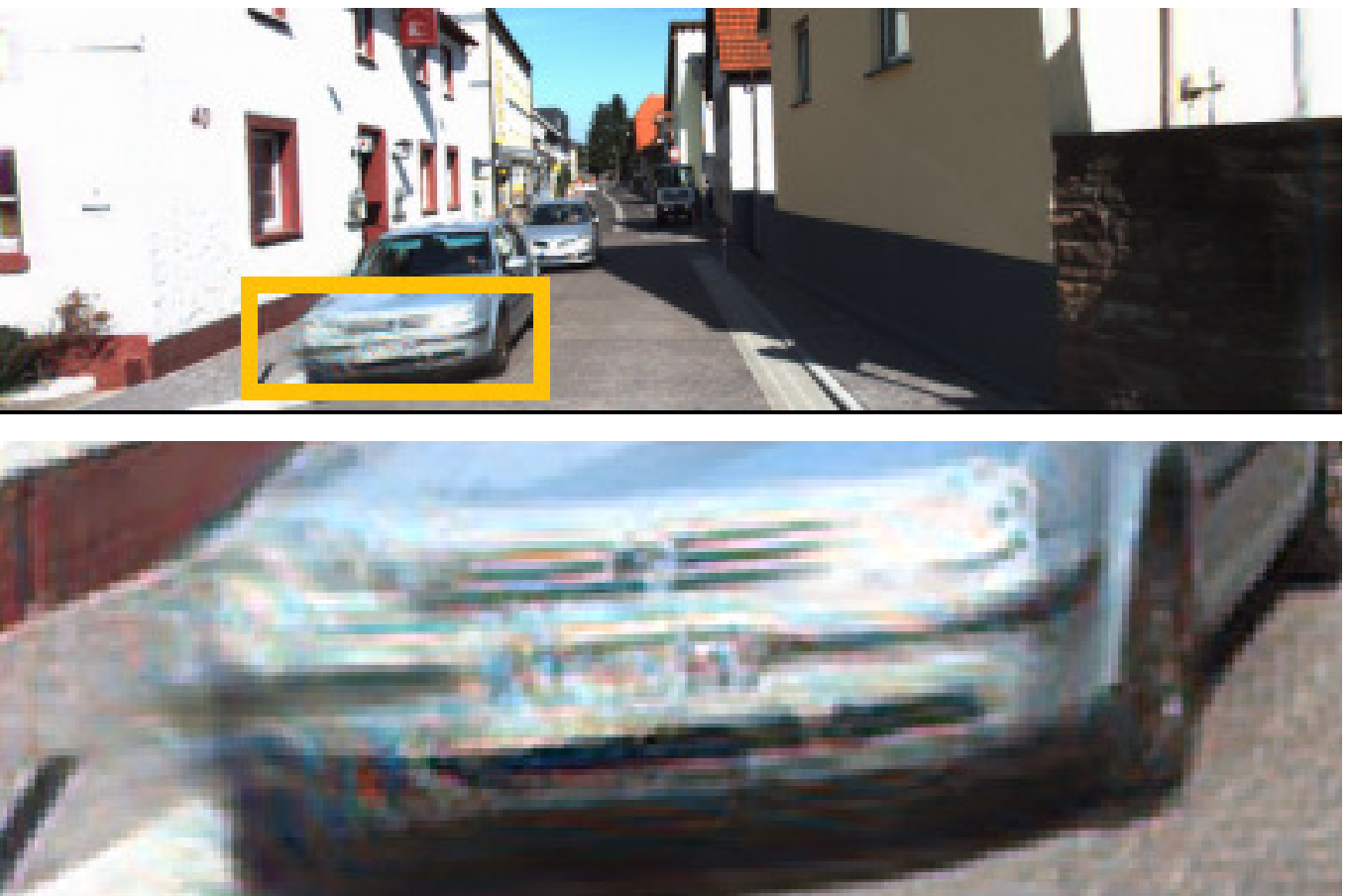}
        \vspace{-0.1cm} \\
            \footnotesize AdaConv
        &
            \footnotesize Ours - $\mathcal{L}_F$
        &
            \footnotesize Direct - $\mathcal{L}_F$
        \\
    \end{tabularx}\vspace{-0.3cm}
    \caption{Comparison with AdaConv and direct synthesis.}\vspace{-0.5cm}
    \label{fig:direct}
\end{figure}

By using different loss functions, we effectively optimized our model for different goals. While our $\mathcal{L}_1$ approach is able to provide better numerical results as reported in the quantitative evaluation in Table~\ref{tbl:middlebury} and \ref{tbl:quantitative}, our $\mathcal{L}_F$ approach achieves higher visual quality as shown in the user study where perceptual quality has been evaluated. 

\begin{figure}\centering
    \setlength{\tabcolsep}{0.0cm}
    \setlength{\itemwidth}{4.15cm}

    \begin{tabularx}{\textwidth}{c @{\hspace{0.05cm}} c}
            \includegraphics[width=\itemwidth]{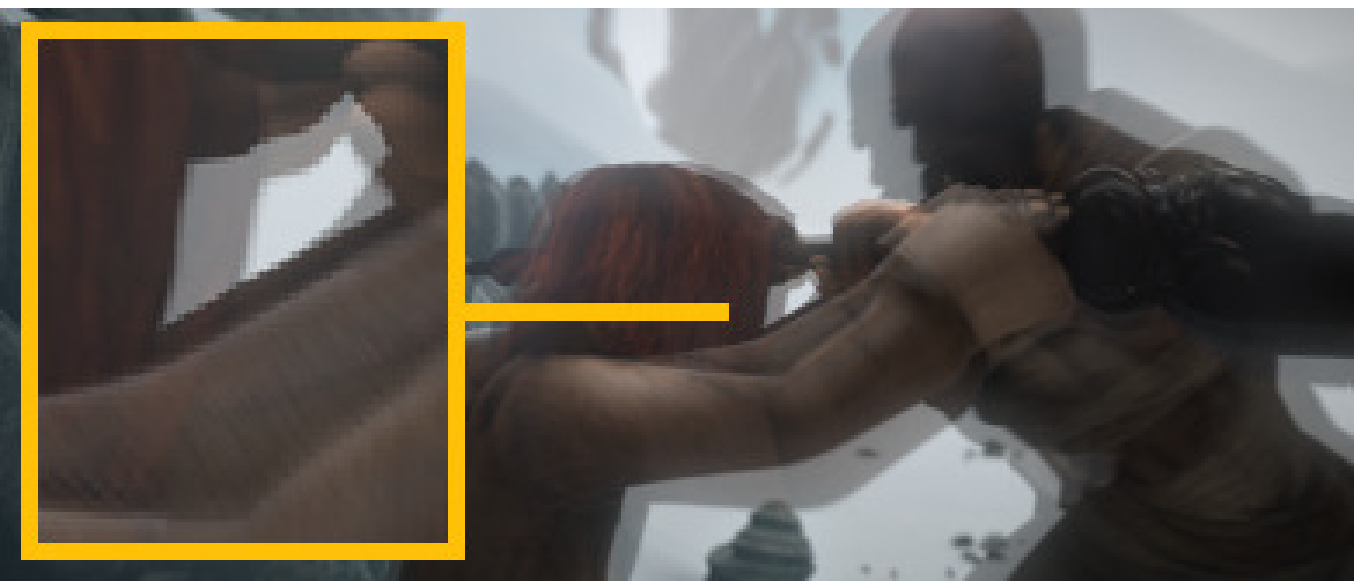}
        &
            \includegraphics[width=\itemwidth, height=1.75cm]{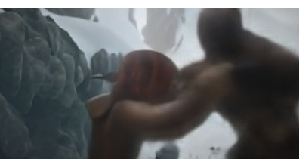}
        \vspace{-0.1cm} \\
            \footnotesize Overlayed input
        &
            \footnotesize Long~\etal
        \\
    \end{tabularx}
    \begin{tabularx}{\textwidth}{c @{\hspace{0.05cm}} c}
            \includegraphics[width=\itemwidth]{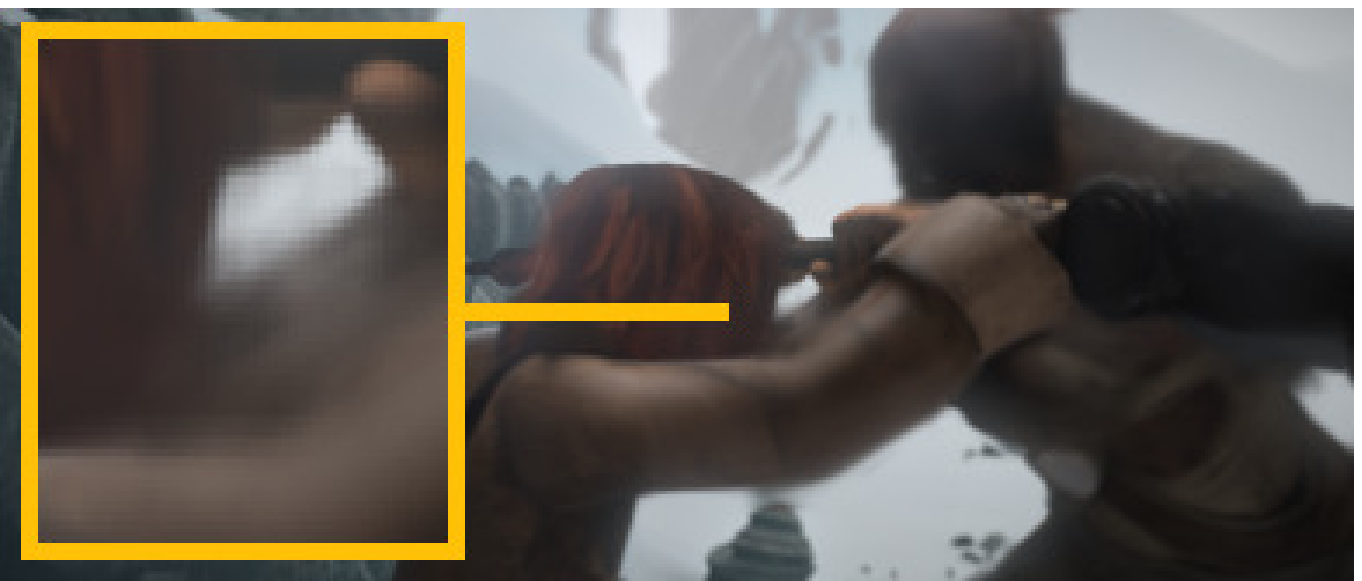}
        &
            \includegraphics[width=\itemwidth]{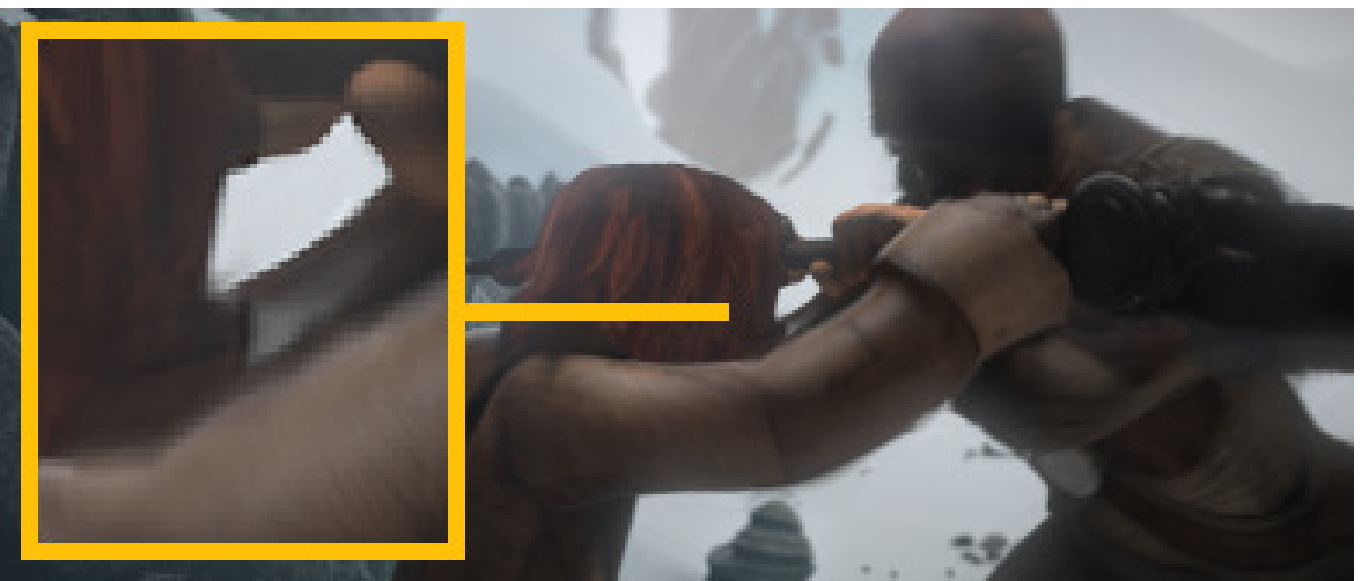}
        \vspace{-0.1cm} \\
            \footnotesize AdaConv
        &
            \footnotesize Ours - $\mathcal{L}_F$
        \\
    \end{tabularx}\vspace{-0.3cm}
    \caption{Comparison with Long~\etal~\cite{Long_ECCV_2016}.}\vspace{-0.5cm}
    \label{fig:gucan}
\end{figure}

One question that has so far been left unanswered is how interpolation via separable convolution compares to directly synthesizing frames using a neural network. Therefore, we adapted our network in order to obtain a baseline for direct synthesis. Specifically, we used one sub-network after the encoder-decoder and let it directly estimate the interpolated frame instead of the kernel coefficients. We furthermore added Batch Normalization~\cite{Sergey_ICML_2015} layers after each block of convolution layers, which improves the quality of this direct synthesis network. We trained this model in the same way we trained our $\mathcal{L}_F$ method. As shown in Figure~\ref{fig:direct}, the direct synthesis leads to blurry results. Additionally, we compare our approach with the image matching method from Long~\etal~\cite{Long_ECCV_2016} that performs direct synthesis to produce a middle frame as an intermediate result. As shown in Figure~\ref{fig:gucan}, our result is sharper. This is consistent with the findings in Zhou~\etal~\cite{Zhou_ECCV_2016} where they argue that synthesizing images from scratch is difficult.

The amount of motion that our method can handle is limited by the kernel size, which is 51 pixels in our system. While this is larger than the recent AdaConv method~\cite{Niklaus_CVPR_2017}, we plan to handle even larger motion by borrowing a multi-scale approach from optical flow research~\cite{Ranjan_CORR_2016}.

Like AdaConv, our approach currently interpolates a frame at $t = 0.5$ in the middle of the two input frames. We cannot produce a frame at an arbitrary time between the input ones. To address this, we could either recursively continue synthesizing frames at $t = 0.25$ and $t = 0.75$, or train a new model from scratch that returns frames at a different temporal offset. Both of these solutions are not ideal and are not as flexible as optical flow based interpolation. In the future, we plan to enhance our neural network to explicitly handle the temporal offset as a control variable.

\section{Conclusion}
\label{sec:concl}

This paper presents a practical solution to high-quality video frame interpolation. The presented method combines motion estimation and frame synthesis into a single convolution process by estimating spatially-adaptive separable kernels for each output pixel and convolving input frames with them to render the intermediate frame. The key to make this convolution approach practical is to use 1D kernels to approximate full 2D ones. The use of 1D kernels significantly reduces the number of kernel parameters and enables full-frame synthesis, which in turn supports the use of perceptual loss to further improve the visual quality of the interpolation results. Our experiments show that our method compares favorably to state-of-the-art interpolation results both quantitatively and qualitatively and produces high-quality frame interpolation results.

\vspace{0.05in}
\noindent\textbf{Acknowledgments.}
Figures~\ref{fig:examples} (top),~\ref{fig:kernel} (top) are used with permission from Gabor Tarnok. The remaining images in Figure~\ref{fig:examples} are used under a Creative Commons license from Alberto Antoniazzi, Ursula Mann and the city of Nuremberg. Figures~\ref{fig:teaser},~\ref{fig:architecture},~\ref{fig:checkerboard},~\ref{fig:loss},~\ref{fig:kernel} (bottom),~\ref{fig:gucan} originate from the Blender Foundation, while Figure~\ref{fig:basketball} and Figure~\ref{fig:direct} originate from the Middlebury and the Kitti benchmark respectively. We thank Nvidia for their GPU donation. This work was supported by NSF IIS-1321119.

{\small
\bibliographystyle{ieee}
\bibliography{egbib}

\begin{thebibliography}{10}\itemsep=-1pt

\bibitem{Aghajanyan_CORR_2017}
A.~Aghajanyan.
\newblock Convolution aware initialization.
\newblock {\em arXiv/1702.06295}, 2017.

\bibitem{Bailer_ICCV_2015}
C.~Bailer, B.~Taetz, and D.~Stricker.
\newblock {Flow Fields}: Dense correspondence fields for highly accurate large
  displacement optical flow estimation.
\newblock In {\em ICCV}, pages 4015--4023, 2015.

\bibitem{Baker_OTHER_2011}
S.~Baker, D.~Scharstein, J.~P. Lewis, S.~Roth, M.~J. Black, and R.~Szeliski.
\newblock A database and evaluation methodology for optical flow.
\newblock {\em International Journal of Computer Vision}, 92(1):1--31, 2011.

\bibitem{Bansal_CORR_2017}
A.~Bansal, X.~Chen, B.~Russell, A.~Gupta, and D.~Ramanan.
\newblock {PixelNet}: Representation of the pixels, by the pixels, and for the
  pixels.
\newblock {\em arXiv/1702.06506}, 2017.

\bibitem{Bishop_BOOK_2006}
C.~M. Bishop.
\newblock {\em Pattern Recognition and Machine Learning}.
\newblock Springer-Verlag New York, Inc., 2006.

\bibitem{Burger_CVPR_2012}
H.~C. Burger, C.~J. Schuler, and S.~Harmeling.
\newblock Image denoising: Can plain neural networks compete with {BM3D}?
\newblock In {\em IEEE Conference on Computer Vision and Pattern Recognition},
  pages 2392--2399, 2012.

\bibitem{Chetlur_CORR_2014}
S.~Chetlur, C.~Woolley, P.~Vandermersch, J.~Cohen, J.~Tran, B.~Catanzaro, and
  E.~Shelhamer.
\newblock {cuDNN}: Efficient primitives for deep learning.
\newblock {\em arXiv/1410.0759}, 2014.

\bibitem{Collobert_OTHER_2011}
R.~Collobert, K.~Kavukcuoglu, and C.~Farabet.
\newblock {Torch7}: A matlab-like environment for machine learning.
\newblock In {\em BigLearn, NIPS Workshop}, 2011.

\bibitem{Dong_ICCV_2015}
C.~Dong, Y.~Deng, C.~C. Loy, and X.~Tang.
\newblock Compression artifacts reduction by a deep convolutional network.
\newblock In {\em ICCV}, pages 576--584, 2015.

\bibitem{Dong_PAMI_2016}
C.~Dong, C.~C. Loy, K.~He, and X.~Tang.
\newblock Image super-resolution using deep convolutional networks.
\newblock {\em IEEE Transactions on Pattern Analysis and Machine Intelligence},
  38(2):295--307, 2016.

\bibitem{Dosovitskiy_NIPS_2016}
A.~Dosovitskiy and T.~Brox.
\newblock Generating images with perceptual similarity metrics based on deep
  networks.
\newblock In {\em Advances in Neural Information Processing Systems}, pages
  658--666, 2016.

\bibitem{Dosovitskiy_ICCV_2015}
A.~Dosovitskiy, P.~Fischer, E.~Ilg, P.~H{\"{a}}usser, C.~Hazirbas, V.~Golkov,
  P.~van~der Smagt, D.~Cremers, and T.~Brox.
\newblock {FlowNet}: Learning optical flow with convolutional networks.
\newblock In {\em ICCV}, pages 2758--2766, 2015.

\bibitem{Dosovitskiy_CVPR_2015}
A.~Dosovitskiy, J.~T. Springenberg, and T.~Brox.
\newblock Learning to generate chairs with convolutional neural networks.
\newblock In {\em IEEE Conference on Computer Vision and Pattern Recognition},
  pages 1538--1546, 2015.

\bibitem{Finn_NIPS_2016}
C.~Finn, I.~J. Goodfellow, and S.~Levine.
\newblock Unsupervised learning for physical interaction through video
  prediction.
\newblock In {\em Advances in Neural Information Processing Systems}, pages
  64--72, 2016.

\bibitem{Flynn_CVPR_2016}
J.~Flynn, I.~Neulander, J.~Philbin, and N.~Snavely.
\newblock {DeepStereo}: Learning to predict new views from the world's imagery.
\newblock In {\em IEEE Conference on Computer Vision and Pattern Recognition},
  pages 5515--5524, 2016.

\bibitem{Gadot_CVPR_2015}
D.~Gadot and L.~Wolf.
\newblock {PatchBatch}: {A} batch augmented loss for optical flow.
\newblock In {\em IEEE Conference on Computer Vision and Pattern Recognition},
  pages 4236--4245, 2016.

\bibitem{Gatys_CVPR_2016}
L.~A. Gatys, A.~S. Ecker, and M.~Bethge.
\newblock Image style transfer using convolutional neural networks.
\newblock In {\em IEEE Conference on Computer Vision and Pattern Recognition},
  pages 2414--2423, 2016.

\bibitem{Goroshin_NIPS_2015}
R.~Goroshin, M.~Mathieu, and Y.~LeCun.
\newblock Learning to linearize under uncertainty.
\newblock In {\em Advances in Neural Information Processing Systems}, pages
  1234--1242, 2015.

\bibitem{Guney_ACCV_2016}
F.~G{\"{u}}ney and A.~Geiger.
\newblock Deep discrete flow.
\newblock In {\em Asian Conference on Computer Vision}, volume 10114, pages
  207--224, 2016.

\bibitem{Sergey_ICML_2015}
S.~Ioffe and C.~Szegedy.
\newblock Batch normalization: Accelerating deep network training by reducing
  internal covariate shift.
\newblock In {\em International Conference on Machine Learning}, volume~37,
  pages 448--456, 2015.

\bibitem{Jia_NIPS_2016}
X.~Jia, B.~D. Brabandere, T.~Tuytelaars, and L.~V. Gool.
\newblock Dynamic filter networks.
\newblock In {\em Advances in Neural Information Processing Systems}, pages
  667--675, 2016.

\bibitem{Johnson_ECCV_2016}
J.~Johnson, A.~Alahi, and L.~Fei{-}Fei.
\newblock Perceptual losses for real-time style transfer and super-resolution.
\newblock In {\em European Conference on Computer Vision}, volume 9906, pages
  694--711, 2016.

\bibitem{Kalantari_TOG_2016}
N.~K. Kalantari, T.~Wang, and R.~Ramamoorthi.
\newblock Learning-based view synthesis for light field cameras.
\newblock {\em {ACM} Trans. Graph.}, 35(6):193:1--193:10, 2016.

\bibitem{Keskar_CORR_2016}
N.~S. Keskar, D.~Mudigere, J.~Nocedal, M.~Smelyanskiy, and P.~T.~P. Tang.
\newblock On large-batch training for deep learning: Generalization gap and
  sharp minima.
\newblock {\em arXiv/1609.04836}, 2016.

\bibitem{Kingma_CORR_2014}
D.~P. Kingma and J.~Ba.
\newblock Adam: {A} method for stochastic optimization.
\newblock {\em arXiv:1412.6980}, 2014.

\bibitem{Kulkarni_NIPS_2015}
T.~D. Kulkarni, W.~F. Whitney, P.~Kohli, and J.~B. Tenenbaum.
\newblock Deep convolutional inverse graphics network.
\newblock In {\em Advances in Neural Information Processing Systems}, pages
  2539--2547, 2015.

\bibitem{Ledig_CORR_2016}
C.~Ledig, L.~Theis, F.~Huszar, J.~Caballero, A.~P. Aitken, A.~Tejani, J.~Totz,
  Z.~Wang, and W.~Shi.
\newblock Photo-realistic single image super-resolution using a generative
  adversarial network.
\newblock {\em arXiv/1609.04802}, 2016.

\bibitem{Li_ECCV_2016}
C.~Li and M.~Wand.
\newblock Precomputed real-time texture synthesis with markovian generative
  adversarial networks.
\newblock In {\em European Conference on Computer Vision}, volume 9907, pages
  702--716, 2016.

\bibitem{Liu_CORR_2017}
Z.~Liu, R.~Yeh, X.~Tang, Y.~Liu, and A.~Agarwala.
\newblock Video frame synthesis using deep voxel flow.
\newblock {\em arXiv/1702.02463}, 2017.

\bibitem{Long_ECCV_2016}
G.~Long, L.~Kneip, J.~M. Alvarez, H.~Li, X.~Zhang, and Q.~Yu.
\newblock Learning image matching by simply watching video.
\newblock In {\em European Conference on Computer Vision}, volume 9910, pages
  434--450, 2016.

\bibitem{Long_CVPR_2015}
J.~Long, E.~Shelhamer, and T.~Darrell.
\newblock Fully convolutional networks for semantic segmentation.
\newblock In {\em IEEE Conference on Computer Vision and Pattern Recognition},
  pages 3431--3440, 2015.

\bibitem{Luo_NIPS_2016}
W.~Luo, Y.~Li, R.~Urtasun, and R.~S. Zemel.
\newblock Understanding the effective receptive field in deep convolutional
  neural networks.
\newblock In {\em Advances in Neural Information Processing Systems}, pages
  4898--4906, 2016.

\bibitem{Mahajan_TOG_2009}
D.~Mahajan, F.~Huang, W.~Matusik, R.~Ramamoorthi, and P.~N. Belhumeur.
\newblock Moving gradients: A path-based method for plausible image
  interpolation.
\newblock {\em {ACM} Trans. Graph.}, 28(3):42:1--42:11, 2009.

\bibitem{Mathieu_ICLR_2016}
M.~Mathieu, C.~Couprie, and Y.~LeCun.
\newblock Deep multi-scale video prediction beyond mean square error.
\newblock In {\em International Conference on Learning Representations}, 2016.

\bibitem{Meyer_CVPR_2015}
S.~Meyer, O.~Wang, H.~Zimmer, M.~Grosse, and A.~Sorkine{-}Hornung.
\newblock Phase-based frame interpolation for video.
\newblock In {\em IEEE Conference on Computer Vision and Pattern Recognition},
  pages 1410--1418, 2015.

\bibitem{Niklaus_CVPR_2017}
S.~Niklaus, L.~Mai, and F.~Liu.
\newblock Video frame interpolation via adaptive convolution.
\newblock In {\em IEEE Conference on Computer Vision and Pattern Recognition},
  July 2017.

\bibitem{Odena_OTHER_2016}
A.~Odena, V.~Dumoulin, and C.~Olah.
\newblock Deconvolution and checkerboard artifacts.
\newblock {\em Distill}, 2016.
\newblock http://distill.pub/2016/deconv-checkerboard.

\bibitem{Ranjan_CORR_2016}
A.~Ranjan and M.~J. Black.
\newblock Optical flow estimation using a spatial pyramid network.
\newblock {\em arXiv/1611.00850}, 2016.

\bibitem{Ranzato_CORR_2014}
M.~Ranzato, A.~Szlam, J.~Bruna, M.~Mathieu, R.~Collobert, and S.~Chopra.
\newblock Video (language) modeling: a baseline for generative models of
  natural videos.
\newblock {\em arXiv/1412.6604}, 2014.

\bibitem{Ridgeway_CORR_2015}
K.~Ridgeway, J.~Snell, B.~Roads, R.~S. Zemel, and M.~C. Mozer.
\newblock Learning to generate images with perceptual similarity metrics.
\newblock {\em arXiv/1511.06409}, 2015.

\bibitem{Rigamonti_CVPR_2013}
R.~Rigamonti, A.~Sironi, V.~Lepetit, and P.~Fua.
\newblock Learning separable filters.
\newblock In {\em IEEE Conference on Computer Vision and Pattern Recognition},
  pages 2754--2761, 2013.

\bibitem{Sajjadi_CORR_2016}
M.~S.~M. Sajjadi, B.~Sch{\"{o}}lkopf, and M.~Hirsch.
\newblock {EnhanceNet}: Single image super-resolution through automated texture
  synthesis.
\newblock {\em arXiv/1612.07919}, 2016.

\bibitem{Shi_CVPR_2016}
W.~Shi, J.~Caballero, F.~Huszar, J.~Totz, A.~P. Aitken, R.~Bishop, D.~Rueckert,
  and Z.~Wang.
\newblock Real-time single image and video super-resolution using an efficient
  sub-pixel convolutional neural network.
\newblock In {\em IEEE Conference on Computer Vision and Pattern Recognition},
  pages 1874--1883, 2016.

\bibitem{Simonyan_CORR_2014}
K.~Simonyan and A.~Zisserman.
\newblock Very deep convolutional networks for large-scale image recognition.
\newblock {\em arXiv/1409.1556}, 2014.

\bibitem{Srivastava_ICML_2015}
N.~Srivastava, E.~Mansimov, and R.~Salakhutdinov.
\newblock Unsupervised learning of video representations using {LSTMs}.
\newblock In {\em International Conference on Machine Learning}, volume~37,
  pages 843--852, 2015.

\bibitem{Jiansun_CVPR_2015}
J.~Sun, W.~Cao, Z.~Xu, and J.~Ponce.
\newblock Learning a convolutional neural network for non-uniform motion blur
  removal.
\newblock In {\em IEEE Conference on Computer Vision and Pattern Recognition},
  pages 769--777, 2015.

\bibitem{Svoboda_CORR_2016}
P.~Svoboda, M.~Hradis, D.~Barina, and P.~Zemc{\'{\i}}k.
\newblock Compression artifacts removal using convolutional neural networks.
\newblock {\em arXiv/1605.00366}, 2016.

\bibitem{Tao_OTHER_2012}
M.~W. Tao, J.~Bai, P.~Kohli, and S.~Paris.
\newblock {SimpleFlow}: {A} non-iterative, sublinear optical flow algorithm.
\newblock {\em Computer Graphics Forum}, 31(2):345--353, 2012.

\bibitem{Tatarchenko_ECCV_2016}
M.~Tatarchenko, A.~Dosovitskiy, and T.~Brox.
\newblock Multi-view {3D} models from single images with a convolutional
  network.
\newblock In {\em European Conference on Computer Vision}, volume 9911, pages
  322--337, 2016.

\bibitem{Teney_CORR_2016}
D.~Teney and M.~Hebert.
\newblock Learning to extract motion from videos in convolutional neural
  networks.
\newblock {\em arXiv/1601.07532}, 2016.

\bibitem{Tran_CVPR_2015}
D.~Tran, L.~D. Bourdev, R.~Fergus, L.~Torresani, and M.~Paluri.
\newblock Deep {End2End} {Voxel2Voxel} prediction.
\newblock In {\em {CVPR} Workshops}, pages 402--409, 2016.

\bibitem{Weinzaepfel_ICCV_2013}
P.~Weinzaepfel, J.~Revaud, Z.~Harchaoui, and C.~Schmid.
\newblock {DeepFlow}: Large displacement optical flow with deep matching.
\newblock In {\em ICCV}, pages 1385--1392, 2013.

\bibitem{Werlberger_OTHER_2011}
M.~Werlberger, T.~Pock, M.~Unger, and H.~Bischof.
\newblock Optical flow guided {TV-L} 1 video interpolation and restoration.
\newblock In {\em Energy Minimization Methods in Computer Vision and Pattern
  Recognition}, volume 6819, pages 273--286, 2011.

\bibitem{Xie_ECCV_2016}
J.~Xie, R.~B. Girshick, and A.~Farhadi.
\newblock {Deep3D}: Fully automatic {2D}-to-{3D} video conversion with deep
  convolutional neural networks.
\newblock In {\em European Conference on Computer Vision}, volume 9908, pages
  842--857, 2016.

\bibitem{Xie_NIPS_2012}
J.~Xie, L.~Xu, and E.~Chen.
\newblock Image denoising and inpainting with deep neural networks.
\newblock In {\em Advances in Neural Information Processing Systems}, pages
  350--358, 2012.

\bibitem{Xu_PAMI_2012}
L.~Xu, J.~Jia, and Y.~Matsushita.
\newblock Motion detail preserving optical flow estimation.
\newblock {\em IEEE Transactions on Pattern Analysis and Machine Intelligence},
  34(9):1744--1757, 2012.

\bibitem{Xu_NIPS_2014}
L.~Xu, J.~S.~J. Ren, C.~Liu, and J.~Jia.
\newblock Deep convolutional neural network for image deconvolution.
\newblock In {\em Advances in Neural Information Processing Systems}, pages
  1790--1798, 2014.

\bibitem{Xue_NIPS_2016}
T.~Xue, J.~Wu, K.~L. Bouman, and B.~Freeman.
\newblock Visual dynamics: Probabilistic future frame synthesis via cross
  convolutional networks.
\newblock In {\em Advances in Neural Information Processing Systems}, pages
  91--99, 2016.

\bibitem{Yang_NIPS_2015}
J.~Yang, S.~E. Reed, M.~Yang, and H.~Lee.
\newblock Weakly-supervised disentangling with recurrent transformations for
  {3D} view synthesis.
\newblock In {\em Advances in Neural Information Processing Systems}, pages
  1099--1107, 2015.

\bibitem{Yu_OTHER_2013}
Z.~Yu, H.~Li, Z.~Wang, Z.~Hu, and C.~W. Chen.
\newblock Multi-level video frame interpolation: Exploiting the interaction
  among different levels.
\newblock {\em {IEEE} Trans. Circuits Syst. Video Techn.}, 23(7):1235--1248,
  2013.

\bibitem{Zeiler_ICCV_2011}
M.~D. Zeiler, G.~W. Taylor, and R.~Fergus.
\newblock Adaptive deconvolutional networks for mid and high level feature
  learning.
\newblock In {\em ICCV}, pages 2018--2025, 2011.

\bibitem{Zhang_ECCV_2016}
R.~Zhang, P.~Isola, and A.~A. Efros.
\newblock Colorful image colorization.
\newblock In {\em European Conference on Computer Vision}, volume 9907, pages
  649--666, 2016.

\bibitem{Zhao_CORR_2016}
H.~Zhao, J.~Shi, X.~Qi, X.~Wang, and J.~Jia.
\newblock Pyramid scene parsing network.
\newblock {\em arXiv/1612.01105}, 2016.

\bibitem{Zhou_ECCV_2016}
T.~Zhou, S.~Tulsiani, W.~Sun, J.~Malik, and A.~A. Efros.
\newblock View synthesis by appearance flow.
\newblock In {\em European Conference on Computer Vision}, volume 9908, pages
  286--301, 2016.

\bibitem{Zhu_ECCV_2016}
J.~Zhu, P.~Kr{\"{a}}henb{\"{u}}hl, E.~Shechtman, and A.~A. Efros.
\newblock Generative visual manipulation on the natural image manifold.
\newblock In {\em European Conference on Computer Vision}, volume 9909, pages
  597--613, 2016.

\bibitem{Zitnick_TOG_2004}
C.~L. Zitnick, S.~B. Kang, M.~Uyttendaele, S.~A.~J. Winder, and R.~Szeliski.
\newblock High-quality video view interpolation using a layered representation.
\newblock {\em {ACM} Trans. Graph.}, 23(3):600--608, 2004.

\end{thebibliography}
}

\end{document}